\documentclass{article} 
\usepackage{iclr2026_conference,times}

\usepackage{amsmath,amsfonts,bm}

\def\eqref#1{equation~\ref{#1}}

\def\1{\bm{1}}

\DeclareMathAlphabet{\mathsfit}{\encodingdefault}{\sfdefault}{m}{sl}
\SetMathAlphabet{\mathsfit}{bold}{\encodingdefault}{\sfdefault}{bx}{n}

\usepackage[hyperfootnotes=false]{hyperref}
\usepackage[utf8]{inputenc} 
\usepackage[T1]{fontenc}   
\usepackage{booktabs}       
\usepackage{amsfonts}      
\usepackage{nicefrac}       
\usepackage{microtype}    
\usepackage{xcolor}        
\usepackage{wrapfig}
\usepackage{multirow}
\usepackage{epsfig}
\usepackage{amsmath}
\usepackage{amssymb}
\usepackage{subfigure}
\usepackage{marvosym}
\usepackage{soul,color}
\usepackage{enumitem}
\usepackage{threeparttable}
\usepackage{algorithm}
\usepackage{algpseudocode}
\usepackage{setspace}
\usepackage{amsthm}
\usepackage{dutchcal}
\usepackage{makecell}
\usepackage{pifont}
\usepackage{xspace}
\usepackage{colortbl}

\usepackage{cleveref}

\usepackage{adjustbox}

\usepackage{titletoc}

\definecolor{MyDarkBlue}{rgb}{0,0.5,1}
\definecolor{MyDarkGreen}{rgb}{0.02,0.6,0.02}
\definecolor{MyDarkRed}{rgb}{0.8,0.02,0.02}
\definecolor{MyDarkOrange}{rgb}{0.40,0.2,0.02}
\definecolor{MyYellow}{rgb}{1,0.55,0}
\definecolor{MyPurple}{RGB}{111,0,255}
\definecolor{MyRed}{rgb}{1.0,0.0,0.0}
\definecolor{MyGold}{rgb}{0.75,0.6,0.12}
\definecolor{MyDarkgray}{rgb}{0.66, 0.66, 0.66}
\definecolor{default}{RGB}{0,0,0}

\newcommand{\boldres}[1]{{\textbf{\textcolor{red}{#1}}}}
\newcommand{\secondres}[1]{{\underline{\textcolor{blue}{#1}}}}

\title{HTMformer: Hybrid Time and Multivariate\\Transformer for Time Series Forecasting}

\author{Tan Wang$^{1}$ 
\quad
Yunwei Dong $^{1}$
\quad
Qi Wang$^{2,3,4}$
\\
$^1$  Northwestern Polytechnical University \\
$^2$ MoE Key Lab of Artificial Intelligence, AI Institute, Shanghai Jiao Tong University\\
$^3$ Ningbo Institute of Digital Twin, Eastern Institute of Technology, Ningbo\\
$^4$ Ningbo Key Laboratory of Spatial Intelligence and Digital Derivative, Ningbo\\
{\tt 330754522@mail.nwpu.edu.cn, yunweidong@nwpu.edu.cn} \\
{\tt tazhang@must.edu.mo, qiwang067@sjtu.edu.cn}
}

\iclrfinalcopy 
\begin{document}

\maketitle

\begin{abstract}
Existing predictors primarily focus on modeling the temporal dimension, often achieving superior performance even with channel-independent designs.
This is because, due to the inherent limitations of time series data, existing multivariate feature extraction strategies tend to introduce substantial noise and brings additional computational overhead.
However, these predictors neglect the inherent multivariate correlations within time series, inevitably leading to an inability to accurately identify complex patterns.
To address this issue, we propose a novel multivariate extraction strategy, \textbf{Hybrid Temporal and Multivariate Embedding (HTME)}. 
HTME separates the multivariate feature space for denoising and pattern extraction, and maps the features to the temporal feature space, yielding multidimensional embeddings that convey richer and more meaningful sequence representations. 
This strategy enables predictors focusing on the temporal dimension fully consider multivariate features with minimal additional computational cost, consistently improving their performance. We construct HTME-versions for eight predictors across three different predictive architectures. In all experiments, the resulting HTMPredictors consistently achieve state-of-the-art performance.

\end{abstract}
\section{Introduction}
\begin{wrapfigure}{r}{0.3\textwidth}
\label{fig:radar}
  \begin{center}
  \vspace{-17pt}
    \includegraphics[width=0.3\textwidth]{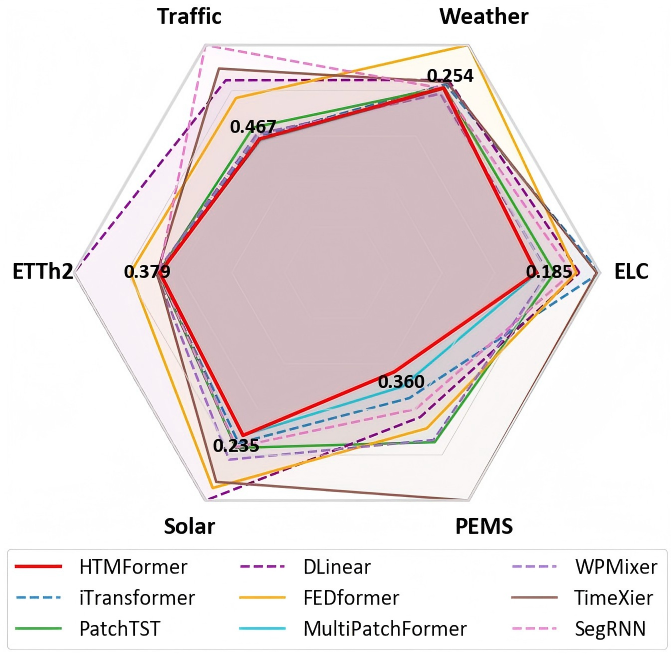}
  \end{center}
  \vspace{-13pt}
  \caption{\small{Performance (MSE) of HTMformer. Smaller image area yields better performance.}}
  \vspace{-14pt}
\end{wrapfigure}

Long-term time series forecasting holds significant importance in various fields such as finance and economics~\citep{Andersen2006}, climate science~\citep{Mudelsee2019}, healthcare~\citep{Zeger2006}, geophysics~\citep{Gubbins2004}, industrial monitoring~\citep{Truong2022}, \textit{etc.} 
The key to time series forecasting lies in capturing temporal dependencies and multivariate correlations, while also meeting the demands of real-time prediction~\citep{Lan2022IntraInter,Yu2024Revitalizing}.

Due to the inherent constraints of time series, the extraction of multivariate features inevitably introduces significant noise\citep{liu2024itransformer,han2023capacity,li2023revisiting}.
So existing prediction models, such as PatchTST and DLinear, focusing solely on the temporal dimension have superior performance. 
We point that this strategy of neglecting multivariate correlations inevitably prevents the model from fully capturing complex time-series patterns.
The absence of key patterns inevitably prevents the model from effectively fitting diverse and complex time series, which severely limits its accuracy and robustness.
Current hybrid stragetys, such as Crossformer, employing a flawed approach to extracting multivariate features. These models introduce extra, complex modules into their backbone networks, which significantly increases computational overhead. Furthermore, severe noise interference  due to a shared feature space, leading to a degradation in accuracy (see \underline{Appendix~\ref{app:A}}).
Therefore, it is necessary to redesign a strategy for extracting multivariate correlations, enabling these models to reconsider them.


To mitigate this issue, this paper proposes a strategy that hybrid temporal and multivariate features in the embedding (HTME) layer , thereby yielding semantically rich embedding representations to overcome the inherent limitations (see \underline{Appendix~\ref{app:C}}).

HTME employs parallel and specialized feature extraction modules to separately capture temporal and multivariate features from the raw input, thereby minimizing interference between them. 
More importantly, the two feature spaces in HTME are disjoint, which confines the noise exclusively to the multivariate feature space.
Then, HTME projects the multivariate features into the temporal feature space. HTME is designed to leverage multivariate features to appropriately supplement the temporal feature space, yielding semantically richer embedding representations.
This enables subsequent modules to sufficiently extract both temporal patterns and mutilvariate correlations by focusing on just a single dimension without additional moudels. 
Rather than mixing or concatenating the feature spaces, HTME suppresses noise within the multivariate space while preserves the integrity and consistency of the temporal feature space.
HTME consistently enhances the performance of various predictors focusing on temporal dimension with only negligible computational overhead.

\begin{figure}[t]  
    \centering 
    \includegraphics[width=0.85\textwidth,height=0.1\textheight]{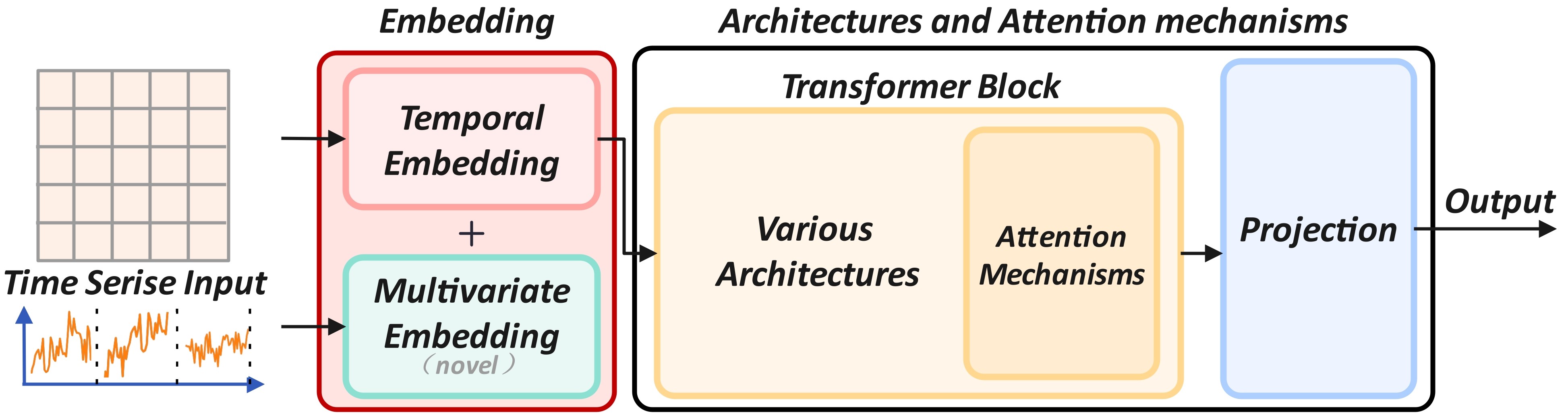}  
    \vspace{-5.5pt}
    \caption{Previous works have focused on redesigning Transformer architectures or attention mechanisms. We propose a novel embedding strategy.
    }  
    \vspace{-11pt}
    \label{fig:introduced}  
\end{figure}

Leveraging this design, we construct a novel Transformer-based framework(HTMformers), as illustrated in Figure~\ref{fig:introduced}. HTMformers consistently enhance the modeling capabilities of various attention mechanisms. By applying the HTME strategy, even simple models can achieve performance comparable to complex models. HTMformer using Full-Attention as a presentative of HTMformers achieves state-of-the-art performance on real-world benchmarks, as shown in Figure~\ref{fig:radar}. 

We further applied the HTME strategy to other forecasting models with different architectures, and it consistently improved their performance.This demonstrates the validity and strong generalizability of the HTME strategy.

The contributions of this work are as follows:
\begin{itemize}
  \item We propose the Hybrid Temporal and Multivariate Embedding (HTME) strategy. It enables predictors focusing on the temporal dimension to fully account for multivariate features, while introducing only minimal computational overhead.
  \item We introduce a novel forecasting framework, HTMformers, which consistently improves the performance of attention mechanisms in forecasting. The representation HTMformer consistently achieves state-of-the-art performance across various benchmarks.  
    \item We selected seven widely-used forecasting models and developed their HTME-augmented versions. Extensive experiments demonstrate that HTME consistently enhances the ability of diverse architectures to model complex time series, particularly for Transformer framework.

\end{itemize}

\section{related Work}
\paragraph{Temporal Feature Extraction.} Previous time series forecasters, especially Transformer-based, such as FEDformer~\citep{zhou2022fedformer} and MultiPatchFormer~\citep{naghashi2025multipatchformer}, tend to overprioritize temporal feature extraction. 
This design often incurs significant computational overhead and can lead to model overfitting.
In contrast, HTME performs a deep embedding of temporal features at the embedding layer and augments them with multivariate features. This process generates semantically richer embedding representations, enabling various models to capture finer-grained features and achieve superior performance.

\paragraph{Multivariate Correlation.}  
GCN~\citep{kipf2017gcn,yu2018stgcn,guo2019astgcn} extracts correlations among variables by leveraging graph adjacency matrices, while iTransformer~\citep{liu2024itransformer} extracts multivariate correlations using Attention. 
Their success highlights the importance of multivariate correlations.
However, this approach of explicitly modeling multivariate correlations incurs significant computational overhead. 
In contrast, the HTME strategy employs a lightweight module to independently extract multivariate features without introducing additional complex modules.

\paragraph{Channel Dependence strategy.}  
Recent studies~\citep{han2023capacity,li2023revisiting} have found that Channel Dependence model makes it prone to capturing spurious correlations from noise and overemphasizing coincidental patterns in the training data. 
This is because existing models, such as Crossformer, make the multivariate features and temporal features share a single space.
HTME independently extracts multivariate features and projects them to the temporal feature space, allowing subsequent modules to remain focused on the temporal dimension, which significantly suppresses noise.

\section{HTMformer}
This paper focuses on multivariate time series forecasting tasks, which can be formally defined as follows:
Given a historical observation matrix $\mathbf{X} \in \mathbb{R}^{L \times C}$, where $L$ denotes the input sequence length and $C$ represents the number of variables, the goal is to predict the future values $\mathbf{Y} \in \mathbb{R}^{H \times C}$, with $H$ indicating the prediction horizon.

\subsection{HTME Extractor}
To fully capture informative representations from the raw time series, the HTME extractor incorporates two novel components: a temporal feature extractor that can more effectively extract temporal patterns and a lightweight multivariate feature extractor.This parallel design which enable the independent modeling reduces interference between features. Finally, we employ a weighted summation for feature fusion, enabling multivariate features to complement temporal features, thereby yielding a richer representation. The lightweight design ensures minimal additional computational overhead. 

Within the two branches, we adopting a bottom-up fusion followed by a top-down decomposition strategy. This strategy encourages the model to focus on specific dimension, thereby avoiding interference between dimensions, while guiding the model to emphasize import patterns, resulting in more effective feature embeddings.

\begin{figure}[t]  
    \centering  
    \includegraphics[width=0.85\textwidth,height=0.12\textheight]{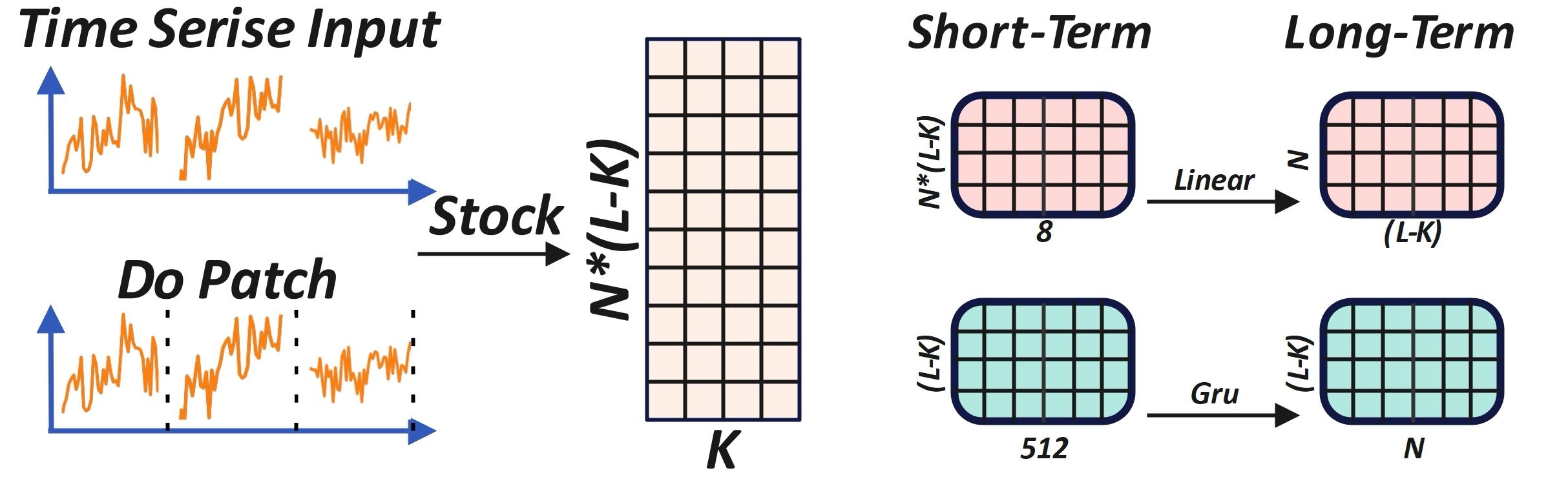}  
    \caption{HTME consists of two independent and parallel feature extraction modules. It employs a strategy of first merging the outputs and then hierarchically decomposing the resulting representation.}
    \label{fig:htme}  
\end{figure}

The HTME extractor takes raw multivariate time series $X_{\text{in}} \in \mathbb{R}^{L \times N}$ as input and outputs embedding representations that fuse informative features, as illustrated in Figure~\ref{fig:htme}.
\subsubsection{Temporal Feature Extractor}
The key to temporal modeling is to extract both short-term and long-term dependencies. It is worth noting that short-term correlations are predominant, while the influence of long-term correlations should not be neglected. The strategy of first merging and then progressively decomposing features effectively guides the model to focus step-by-step on the most important patterns.

First,  we adopt a patching strategy to segment short-term temporal representations from the original sequence, and iteratively merge all channels to obtain a short-step representation $T^{N \times (L-K), K}$ .It is worth noting that the process merely partitions the time series without introducing trainable parameters at this stage.
\begin{align}
P_1, P_2, P_3, \dots, P_{L-K} &= X_{1:1+K}, X_{2+K}, X_{3+K}, \dots, X_{L-K:L} \\
T^{N \times (L-K), K} &= \text{Stack}(P_1, P_2, P_3, \dots, P_{L-K})
\end{align}
Short-term dependencies are particularly important. this module first prioritizes the deep extraction of short-term temporal features. Specifically, we apply multiple convolutional filters to capture short-term temporal patterns at multiple scales. We then fuse the convolutional outputs via a linear layer to obtain a feature representation enriched with short-term dependencies.
\begin{align}
    E^{N \times (L-K), 8} &= \text{Linear}(\text{Conv}(T^{N \times (L-K), 1, K}))
\end{align}
Finally, we unfold the long temporal dimension and employ a second linear layer to capture the long-term dependencies within the sequence, yielding a deep embedding that encapsulates both short-term and long-term temporal dependencies.
\begin{align}
    D_{\text{out}} &= \text{Linear}(\text{Flatten}(E^{N \times (L-K), 8}))
\end{align}
Compared with existing embedding strategies, our method achieves a more fine-grained modeling of temporal contextual dependencies using only one convolutional layer and two linear layers.
\subsubsection{Multivariate Feature Extractor}
We adopt the same merge‑decompose strategy as described above. We segment and stack the time series to obtain $T^{N \times (L-K), K}$, following exactly the same procedure as in the aforementioned module.

Subsequently, we merge the short‑term temporal dimension with the multivariate dimension, and employ a linear layer to process the multi‑step variables, thereby obtaining multivariate features $N^{ (L-K), 128}$ that account for short‑term temporal lags.
\begin{align}
    N^{ (L-K), 128} &= \text{Linear}(T^{(L-K), K \times N})
\end{align}
Similar to Eq. (4), we employ multiple convolutional filters to extract multivariate features at different scales, resulting in a more fine-grained multi-scale multivariate feature block. As this module is substantial in size, we therefore apply a specialized GRU network to aggregate the multi-scale features and obtain the integrated representation $B^N$.
\begin{align}
    B^N &= \text{GRU}(\text{Conv}(N^{ (L-K), 128}))
\end{align}
Considering that the correlations among multivariate variables may differ across time stamps, we apply different convolutional kernels to project the integrated representation $B^N$ onto the temporal dimension, thereby obtaining the final representation.
\begin{align}
    V_{\text{out}} &= \text{Conv}(B^N)
\end{align}
here, $D_{\text{out}}, V_{\text{out}} \in \mathbb{R}^{N \times D}$. We define a learnable fusion weight $\alpha$ to adaptively balance the contributions of the two modules to the downstream model, thereby enhancing the scalability of HTME for adaptation to datasets with diverse characteristics.
The final output is the sum of the two modules:
\begin{align}
Y_{\text{in}} &= \alpha D_{\text{out}} + (1 - \alpha) V_{\text{out}},\label{eq:11}
\end{align}
where $Y_{\text{in}} \in \mathbb{R}^{N \times D}$ denotes the embedded representation that fuses temporal and multivariate features. 
HTME is fed into a vanilla Transformer encoder to generate the predictive representations.

\subsection{Structure Overview}
\begin{figure}[t]  
    \centering  
    \includegraphics[width=1\textwidth]{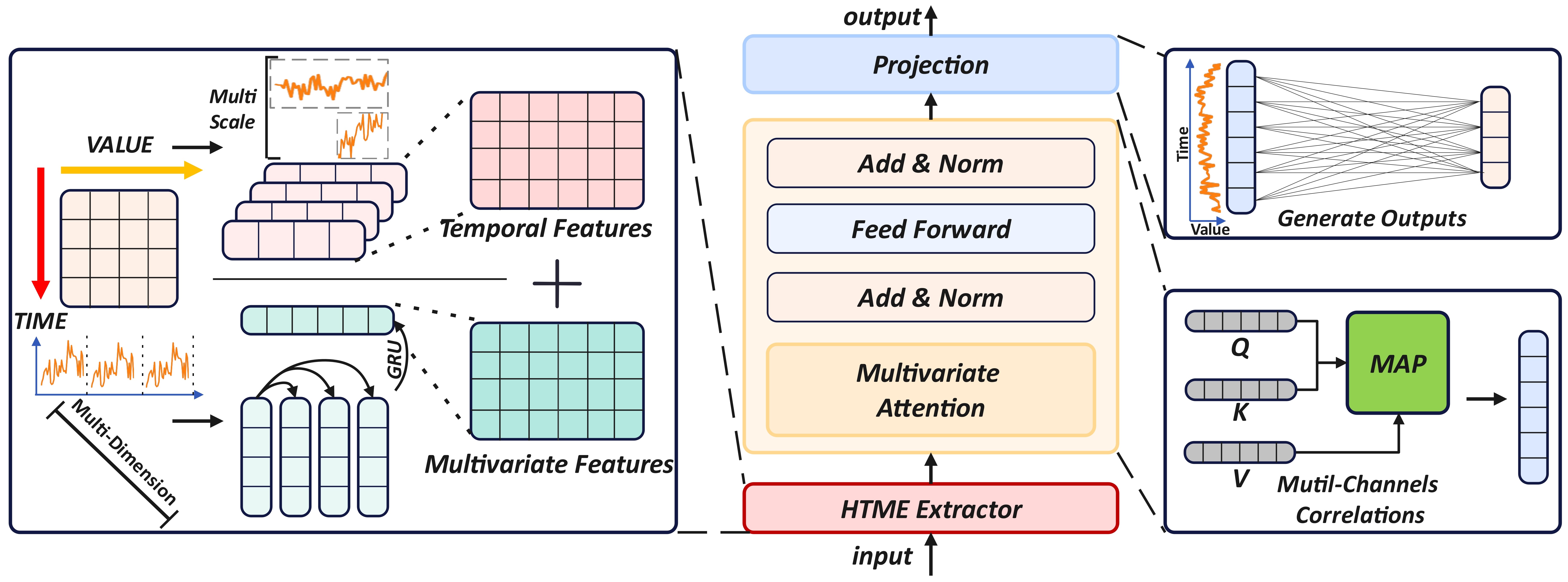}  
    \caption{Architecture of HTMformer. The HTME directly replacing the iTransformer embeddings. It also employs inverted input to enable the attention mechanism to capture channels correlations.}
    \label{fig:figure2}  
\end{figure}

We construct HTMformer based on iTransformer, a state-of-the-art and generic Transformer architecture. It keeps the native components of the original Transformer intact. This implies that other components within HTMformer can be flexibly interchanged with their respective variants.
As shown in Figure~\ref{fig:figure2}, the model includes an HTME extractor, a vanilla Transformer encoder layer, and a projection layer. 
RevIN~\citep{kim2022reversible} is a normalization technique, which can help models learn and generalize better. Each data batch is first normalized with RevIN before being fed into the model.

In HTMformer, showned in Figure~\ref{fig:figure2} (\textbf{Middle}), the overall formulation for predicting the future sequence $Y \in \mathbb{R}^{H \times C}$ from the historical sequence $X \in \mathbb{R}^{L \times C}$ is as follows: 
\begin{equation}
\begin{aligned}
X_{\text{in}} &= \text{CAT}(X,T),\qquad Y_{\text{in}} = \text{HTMEE}(X_{\text{in}}), \\
Y_{\text{out}} &= \text{Encode}(Y_{\text{in}}),\qquad Y = \text{Project}(Y_{\text{out}}).
\end{aligned}
\label{eq:1}
\end{equation}
Following RevIN normalization, timestamps $T$ are appended as an additional variable to the sequence $X \in \mathbb{R}^{L \times C}$ before being fed into the model. 
$X_{\text{in}} \in \mathbb{R}^{L \times N}$, where $N$ denotes the multivariate dimension augmented with the timestamp.
$Y_{\text{in}} \in \mathbb{R}^{N \times D}$ denotes the hybrid temporal and multivariate embedding representations after the HTME extractor, where the embedding dimension of each variable is $D$.
$Y_{\text{out}} \in \mathbb{R}^{N \times D}$ represents the predictive features produced by the Transformer encoder.
Finally, a linear model serves as the projection layer to generate the final output $Y \in \mathbb{R}^{H \times C}$.

\subsection{Encoder and Projection}
The  Transformer encoder layer is composed of a self-attention module and a position-wise feed-forward network.
We adopt an inverted input design, which enables the self-attention mechanism to directly model global channel correlations.
This inverted structure avoids forced alignment of variables within timestamps and facilitates the concatenation of temporal and multivariate dimensions. 
Consequently, it enables the attention mechanism to learn more appropriate sequence representations. 

The attention mechanism projects $Y_{\text{in}} \in \mathbb{R}^{N \times D}$ to generate query matrix $Q_k$, key matrix $K_k$, and value matrix $V_k$, where $d_k$ denotes the projection dimension for each attention head. 
Then, the scaled dot-product is used to derive the attention output $O_k \in \mathbb{R}^{N \times d_k}$:
\begin{align}
\left( {O}_k \right)^T &= \text{Softmax} \left( \frac{Q_k K_k^T}{\sqrt{d_k}} \right) V_k.
\end{align}
The self-attention block additionally consists of layer normalization and a position-wise feed-forward network. 
Finally, a linear projection head serves as the decoder, mapping the learned representations to the target dimensionality and yielding the final predictions.
iTransformer~\citep{liu2024itransformer} observes that this Transformer‑based predictor architecture is more efficient for time series forecasting tasks.

\subsection{Complexity Analysis}
HTME consists of two parallel extraction modules. The time and space complexities of the Feature Extraction Modules are both O(LN). For each dimension, the computational complexity is linear, which guarantees the high efficiency and scalability of HTME. The HTME module does not introduce significant additional computational overhead. The primary computational cost of HTMProdectors stems from the underlying base predictor they are built upon.


\section{Experiments}
To validate the effectiveness and transferability of the HTME strategy, we integrated it into diverse forecasting backbones and evaluated its impact on predictive performance.
Moreover, we conducted a comprehensive evaluation of the proposed HTMformer on diverse forecasting tasks to further validated the advantages of the HTME strategy. 
We also evaluated the computational overhead of HTME, showing that it introduces only modest additional cost.
The detailed experimental configurations and implementation details can be found in \underline{Appendix~\ref{app:E}}.
\subsection{Experiment Settings}
All models are evaluated under identical settings, following the Time-Series-Library~\citep{wang2024tssurvey}, to ensure a fair comparison and reliable conclusions..
We evaluated the performance of HTMformer on eight widely used datasets, including Weather, Traffic, Electricity, ETTh2 adopted in Autoformer~\citep{wu2021autoformer}, Solar-Energy proposed in LSTNet~\citep{lai2018modeling}, and three PEMS datasets (PEMS03, PEMS04, PEMS08) investigated in SCINet~\citep{liu2022scinet}. 
These datasets span diverse domains and sampling frequencies, providing a comprehensive experimental scenario for model evaluation.
We employ eight carefully selected state-of-the-art models from three different architectures.

\subsection{Performance Promotion with HTME}
To assess the effectiveness and scalability of HTME, the vanilla self-attention mechanism~\citep{vaswani2017attention} and its variants are integrtated into the HTMformer framework, as shown in Table~\ref{tab:pre_performance}. 
Specifically, HTMformers achieve consistent improvements across different variants, with average performance gains of \emph{35.8\%}, \emph{34.6\%}, \emph{43.6\%}, \emph{31.9\%},and \emph{33.1\%} for Transformer, Reformer~\citep{kitaev2020reformer}, Informer~\citep{zhou2021informer}, Flowformer~\citep{ma2023flowformer}, and Flashformer~\citep{dao2022flashattention}, respectively. 

\begin{table}[t]
\vspace{-10pt}
\centering
\small
\caption{Performance improvementst by HTME. Full results can be found in \underline{Appendix~\ref{app:F 1}.}}
\vspace{3pt}
\label{tab:pre_performance}
\resizebox{0.95\columnwidth}{!}{
\begin{threeparttable}
\begin{small}
\renewcommand{\multirowsetup}{\centering}
\setlength{\tabcolsep}{2.7pt}
\begin{tabular}{cc|cc|cc|cc|cc|cc|cc|cc|cc|cc}
\toprule
\multicolumn{1}{c}{Models} &\multicolumn{1}{c}{ }  & \multicolumn{2}{c}{Transformer} & \multicolumn{2}{c}{Reformer} & \multicolumn{2}{c}{Informer} & \multicolumn{2}{c}{Flowformer} & \multicolumn{2}{c}{Flashformer} & \multicolumn{2}{c}{iTransformer} & \multicolumn{2}{c}{DLinear} & \multicolumn{2}{c}{NLinear} & \multicolumn{2}{c}{SegRNN}\\
\midrule
Metric &  & MSE & MAE & MSE & MAE & MSE & MAE & MSE & MAE & MSE & MAE & MSE & MAE & MSE & MAE & MSE & MAE & MSE & MAE\\
\midrule
\multirow{3}{*}{ECL} 
& Original   & \scalebox{0.8}{0.271} & \scalebox{0.8}{0.343} & \scalebox{0.8}{0.342} & \scalebox{0.8}{0.417} & \scalebox{0.8}{0.379} & \scalebox{0.8}{0.442} & \scalebox{0.8}{0.272} & \scalebox{0.8}{0.370} & \scalebox{0.8}{0.268} & \scalebox{0.8}{0.364} & \scalebox{0.8}{0.244} & \scalebox{0.8}{0.290} & \scalebox{0.8}{0.225} & \scalebox{0.8}{0.318} & \scalebox{0.8}{0.217} & \scalebox{0.8}{0.295} & \scalebox{0.8}{0.194} & \scalebox{0.8}{0.288} \\
& +HTME      & \scalebox{0.8}{0.185} & \scalebox{0.8}{0.272} & \scalebox{0.8}{0.201} & \scalebox{0.8}{0.292} & \scalebox{0.8}{0.202} & \scalebox{0.8}{0.292} & \scalebox{0.8}{0.198} & \scalebox{0.8}{0.289} & \scalebox{0.8}{0.194} & \scalebox{0.8}{0.287} & \scalebox{0.8}{0.194} & \scalebox{0.8}{0.288} & \scalebox{0.8}{0.219} & \scalebox{0.8}{0.314} & \scalebox{0.8}{0.208} & \scalebox{0.8}{0.291}& \scalebox{0.8}{0.190} & \scalebox{0.8}{0.283} \\
& Promotion  & \scalebox{0.8}{35.3\%} & \scalebox{0.8}{20.6\%} & \scalebox{0.8}{41.2\%} & \scalebox{0.8}{29.9\%} & \scalebox{0.8}{46.7\%} & \scalebox{0.8}{33.9\%} & \scalebox{0.8}{27.2\%} & \scalebox{0.8}{19.4\%} & \scalebox{0.8}{27.6\%} & \scalebox{0.8}{21.1\%} & \scalebox{0.8}{20.4\%}& \scalebox{0.8}{0.6\%}& \scalebox{0.8}{2.6\%}& \scalebox{0.8}{1.2\%}& \scalebox{0.8}{4.1\%}& \scalebox{0.8}{1.3\%}& \scalebox{0.8}{2.0\%} & \scalebox{0.8}{1.7\%}\\
\midrule
\multirow{3}{*}{Weather} 
& Original   & \scalebox{0.8}{0.651} & \scalebox{0.8}{0.572} & \scalebox{0.8}{0.454} & \scalebox{0.8}{0.458} & \scalebox{0.8}{0.644} & \scalebox{0.8}{0.557} & \scalebox{0.8}{0.631} & \scalebox{0.8}{0.559} & \scalebox{0.8}{0.633} & \scalebox{0.8}{0.559} & \scalebox{0.8}{0.259} & \scalebox{0.8}{0.279} & \scalebox{0.8}{0.269} & \scalebox{0.8}{0.290} & \scalebox{0.8}{0.274} & \scalebox{0.8}{0.299}& \scalebox{0.8}{0.254} & \scalebox{0.8}{0.301}   \\
& +HTME      & \scalebox{0.8}{0.254} & \scalebox{0.8}{0.278} & \scalebox{0.8}{0.257} & \scalebox{0.8}{0.276} & \scalebox{0.8}{0.254} & \scalebox{0.8}{0.280} & \scalebox{0.8}{0.254} & \scalebox{0.8}{0.279} & \scalebox{0.8}{0.253} & \scalebox{0.8}{0.280} & \scalebox{0.8}{0.252} & \scalebox{0.8}{0.267} & \scalebox{0.8}{0.261} & \scalebox{0.8}{0.286} & \scalebox{0.8}{0.261} & \scalebox{0.8}{0.290}& \scalebox{0.8}{0.249} & \scalebox{0.8}{0.296}  \\
& Promotion  & \scalebox{0.8}{60.8\%} & \scalebox{0.8}{51.3\%} & \scalebox{0.8}{45.1\%} & \scalebox{0.8}{39.9\%} & \scalebox{0.8}{60.5\%} & \scalebox{0.8}{49.7\%} & \scalebox{0.8}{57.9\%} & \scalebox{0.8}{50.0\%} & \scalebox{0.8}{60.0\%} & \scalebox{0.8}{49.9\%}& \scalebox{0.8}{2.7\%}& \scalebox{0.8}{4.0\%}& \scalebox{0.8}{2.9\%}& \scalebox{0.8}{1.3\%}& \scalebox{0.8}{4.7\%} & \scalebox{0.8}{3.0\%}& \scalebox{0.8}{1.9\%}& \scalebox{0.8}{1.6\%}\\
\midrule
\multirow{3}{*}{Traffic} 
& Original   & \scalebox{0.8}{0.666} & \scalebox{0.8}{0.366} & \scalebox{0.8}{0.706} & \scalebox{0.8}{0.389} & \scalebox{0.8}{0.825} & \scalebox{0.8}{0.465} & \scalebox{0.8}{0.655} & \scalebox{0.8}{0.359} & \scalebox{0.8}{0.661} & \scalebox{0.8}{0.362} & \scalebox{0.8}{0.474} & \scalebox{0.8}{0.317} & \scalebox{0.8}{0.672} & \scalebox{0.8}{0.418} & \scalebox{0.8}{0.632} & \scalebox{0.8}{0.391}& \scalebox{0.8}{0.761} & \scalebox{0.8}{0.397}   \\
& +HTME      & \scalebox{0.8}{0.467} & \scalebox{0.8}{0.312} & \scalebox{0.8}{0.475} & \scalebox{0.8}{0.314} & \scalebox{0.8}{0.488} & \scalebox{0.8}{0.323} & \scalebox{0.8}{0.481} & \scalebox{0.8}{0.321} & \scalebox{0.8}{0.474} & \scalebox{0.8}{0.319} & \scalebox{0.8}{0.473} & \scalebox{0.8}{0.321} & \scalebox{0.8}{0.609} & \scalebox{0.8}{0.383} & \scalebox{0.8}{0.611} & \scalebox{0.8}{0.384}& \scalebox{0.8}{0.750} & \scalebox{0.8}{0.392} \\
& Promotion  & \scalebox{0.8}{29.8\%} & \scalebox{0.8}{17.3\%} & \scalebox{0.8}{32.7\%} & \scalebox{0.8}{19.2\%} & \scalebox{0.8}{40.8\%} & \scalebox{0.8}{30.5\%} & \scalebox{0.8}{26.5\%} & \scalebox{0.8}{10.5\%} & \scalebox{0.8}{28.2\%} & \scalebox{0.8}{11.8\%} & \scalebox{0.8}{0.2\%}& \scalebox{0.8}{0.0\%} & \scalebox{0.8}{9.3\%}& \scalebox{0.8}{8.1\%}& \scalebox{0.8}{3.3\%}& \scalebox{0.8}{1.7\%}& \scalebox{0.8}{1.4\%}& \scalebox{0.8}{1.3\%} \\
\midrule
Promotion Avg &  & \multicolumn{2}{c|}{\scalebox{0.8}{35.8\%}} & \multicolumn{2}{c|}{\scalebox{0.8}{34.6\%}} & \multicolumn{2}{c|}{\scalebox{0.8}{43.6\%}} & \multicolumn{2}{c|}{\scalebox{0.8}{31.9\%}} & \multicolumn{2}{c|}{\scalebox{0.8}{33.1\%}}& \multicolumn{2}{c|}{\scalebox{0.8}{4.6\%}}& \multicolumn{2}{c|}{\scalebox{0.8}{4.2\%}}& \multicolumn{2}{c|}{\scalebox{0.8}{3.0\%}}& \multicolumn{2}{c}{\scalebox{0.8}{1.7\%}}\\
\bottomrule
\end{tabular}
\end{small}
\end{threeparttable}
}
\end{table}

To demonstrate the strong scalability of HTME strategy, we apply it to diverse architectures, including the Transformer-based iTransformer, the linear-based DLinear, NLinear and the RNN-based SegRNN. For these models, we supplement the original input with multivariate features. The results show that HTME consistently improves the forecasting performance across all these models.

The HTME strategy consistently boosts the performance of a wide range of forecasting models. More importantly, the multivariate representations produced by HTME can be seamlessly integrated with other embedding schemes that emphasize temporal characteristics. These further enhance the effectiveness and scalability of HTME.
\subsection{Long-term Forecasting}
Long-term forecasting is critical for real-world applications such as meteorology, traffic management, and energy systems. 
Table~\ref{tab:longterm_avg} presents the experimental forecasting results.
Within each row, the lowest averaged MSE and MAE values across four prediction horizons are highlighted in \boldres{red}, and the second-lowest values are underscored in \secondres{blue}. 
HTMformer achieves the best average performance on most datasets, with particularly strong gains on high-dimensional benchmarks such as Traffic.
We further provide qualitative visualizations of the predicted trajectories (see \underline{Appendix~\ref{app:J}}). 
Overall, HTMformer delivers significant improvements in long-term forecasting, highlighting its practical value.
\begin{table}[ht]
  \caption{Average performance for long-term forecasting over prediction horizons $H \in \{96, 192, 336, 720\}$.
  MPFormer is MultiPatchFormer. Full results are listed in \underline{Appendix~\ref{app:F 2}}. }
  \centering
  \resizebox{1\columnwidth}{!}{
  \begin{threeparttable}
  \begin{small}
  \renewcommand{\multirowsetup}{\centering}
  \setlength{\tabcolsep}{1.45pt}
  \label{tab:longterm_avg}
  \begin{tabular}{c|cc|cc|cc|cc|cc|cc|cc|cc|cc}
    \toprule
    \multicolumn{1}{c}{\multirow{2}{*}{Models}} & 
    \multicolumn{2}{c}{\rotatebox{0}{\scalebox{0.8}{\textbf{HTMformer}}}} &
    \multicolumn{2}{c}{\rotatebox{0}{\scalebox{0.8}{iTransformer}}} &
    \multicolumn{2}{c}{\rotatebox{0}{\scalebox{0.8}{PatchTST}}} &
    \multicolumn{2}{c}{\rotatebox{0}{\scalebox{0.8}{DLinear}}} &
    \multicolumn{2}{c}{\rotatebox{0}{\scalebox{0.8}{FEDformer}}} &
    \multicolumn{2}{c}{\rotatebox{0}{\scalebox{0.8}{MPFormer}}} &
    \multicolumn{2}{c}{\rotatebox{0}{\scalebox{0.8}{WPMixer}}} & 
    \multicolumn{2}{c}{\rotatebox{0}{\scalebox{0.8}{TimeMixer}}} &
    \multicolumn{2}{c}{\rotatebox{0}{\scalebox{0.8}{SegRNN}}} \\ 
    
    \multicolumn{1}{c}{} & 
    \multicolumn{2}{c}{\scalebox{0.8}{(\textbf{Ours})}} &
    \multicolumn{2}{c}{\scalebox{0.8}{(\citeyear{liu2024itransformer})}} &
    \multicolumn{2}{c}{\scalebox{0.8}{\citeyear{nie2023time}}} &
    \multicolumn{2}{c}{\scalebox{0.8}{\citeyear{zeng2023dlinear}}}&
    \multicolumn{2}{c}{\scalebox{0.8}{\citeyear{zhou2022fedformer}}}&
    \multicolumn{2}{c}{\scalebox{0.8}{\citeyear{naghashi2025multipatchformer}}}&
    \multicolumn{2}{c}{\scalebox{0.8}{\citeyear{murad2024wp}}}&
    \multicolumn{2}{c}{\scalebox{0.8}{\citeyear{wang2023timemixer}}}&
    \multicolumn{2}{c}{\scalebox{0.8}{\citeyear{lin2023segrnn}}} \\ 
    \midrule
    \multicolumn{1}{c|}{Metric} & MSE & MAE & MSE & MAE & MSE & MAE& MSE & MAE& MSE & MAE & MSE & MAE & MSE & MAE& MSE & MAE& MSE & MAE\\
    \midrule
    \scalebox{0.95}{ECL}
    & \boldres{\scalebox{0.8}{0.185}} & \boldres{\scalebox{0.8}{0.272}}
    & \scalebox{0.8}{0.244} & \scalebox{0.8}{0.290}
    & \scalebox{0.8}{0.202} & \scalebox{0.8}{0.286}
    & \scalebox{0.8}{0.225} & \scalebox{0.8}{0.318}
    & \scalebox{0.8}{0.222} & \scalebox{0.8}{0.333}
    & \secondres{\scalebox{0.8}{0.186}} & \secondres{\scalebox{0.8}{0.273}}
    & \scalebox{0.8}{0.196} & \scalebox{0.8}{0.281}
    & \scalebox{0.8}{0.241} & \scalebox{0.8}{0.328}
    & \scalebox{0.8}{0.218} & \scalebox{0.8}{0.302} \\
    \midrule
    \scalebox{0.95}{Weather} &
    {\scalebox{0.8}{0.254}} & {\scalebox{0.8}{0.277}} &
    {\scalebox{0.8}{0.259}} & {\scalebox{0.8}{0.279}} &
    {\scalebox{0.8}{0.255}} & {\scalebox{0.8}{0.278}} &
    {\scalebox{0.8}{0.265}} & {\scalebox{0.8}{0.316}} &
    {\scalebox{0.8}{0.313}} & {\scalebox{0.8}{0.362}} &
    \secondres{\scalebox{0.8}{0.252}} & \secondres{\scalebox{0.8}{0.274}} &
    \boldres{\scalebox{0.8}{0.246}} & \boldres{\scalebox{0.8}{0.273}} &
    {\scalebox{0.8}{0.262}} & {\scalebox{0.8}{0.286}} &
    {\scalebox{0.8}{0.253}} & {\scalebox{0.8}{0.3}} \\
    \midrule 
    \scalebox{0.95}{Traffic} & 
    \secondres{\scalebox{0.8}{0.467}} & \secondres{\scalebox{0.8}{0.312}} &
    {\scalebox{0.8}{0.474}} & {\scalebox{0.8}{0.317}} &
    {\scalebox{0.8}{0.507}} & {\scalebox{0.8}{0.324}} &
    {\scalebox{0.8}{0.672}} & {\scalebox{0.8}{0.418}} &
    {\scalebox{0.8}{0.610}} & {\scalebox{0.8}{0.378}} &
    \boldres{\scalebox{0.8}{0.462}} & \boldres{\scalebox{0.8}{0.304}} &
    {\scalebox{0.8}{0.484}} & {\scalebox{0.8}{0.338}} &
    {\scalebox{0.8}{0.713}} & {\scalebox{0.8}{0.445}} &
    {\scalebox{0.8}{0.795}} & {\scalebox{0.8}{0.408}} \\
    \midrule
    \scalebox{0.95}{ETTh2} & 
    \boldres{\scalebox{0.8}{0.379}} & \boldres{\scalebox{0.8}{0.399}} &
    {\scalebox{0.8}{0.383}} & {\scalebox{0.8}{0.406}} &
    {\scalebox{0.8}{0.382}} & \secondres{\scalebox{0.8}{0.404}} &
    {\scalebox{0.8}{0.563}} & {\scalebox{0.8}{0.519}} &
    {\scalebox{0.8}{0.442}} & {\scalebox{0.8}{0.454}} &
    \secondres{\scalebox{0.8}{0.381}} & {\scalebox{0.8}{0.406}} &
    {\scalebox{0.8}{0.387}} & {\scalebox{0.8}{0.410}} &
    {\scalebox{0.8}{0.385}} & {\scalebox{0.8}{0.409}} &
    \secondres{\scalebox{0.8}{0.381}} & {\scalebox{0.8}{0.414}} \\
    \midrule
    \scalebox{0.95}{Solar} &     
    \boldres{\scalebox{0.8}{0.235}} & \boldres{\scalebox{0.8}{0.265}} &
    \secondres{\scalebox{0.8}{0.246}} & {\scalebox{0.8}{0.278}} &
    {\scalebox{0.8}{0.253}} & {\scalebox{0.8}{0.289}} &
    {\scalebox{0.8}{0.329}} & {\scalebox{0.8}{0.400}} &
    {\scalebox{0.8}{0.311}} & {\scalebox{0.8}{0.392}} &
    \boldres{\scalebox{0.8}{0.235}} & \secondres{\scalebox{0.8}{0.267}} &
    {\scalebox{0.8}{0.27}} & {\scalebox{0.8}{0.303}} &
    {\scalebox{0.8}{0.302}} & {\scalebox{0.8}{0.323}} &
    {\scalebox{0.8}{0.252}} & {\scalebox{0.8}{0.304}} \\
    \midrule 
    \scalebox{0.95}{PEMS03} &     
    \boldres{\scalebox{0.8}{0.289}} & \boldres{\scalebox{0.8}{0.369}} &
    {\scalebox{0.8}{0.360}} & {\scalebox{0.8}{0.421}} &
    {\scalebox{0.8}{0.502}} & {\scalebox{0.8}{0.51}} &
    {\scalebox{0.8}{0.442}} & {\scalebox{0.8}{0.498}} &
    {\scalebox{0.8}{0.467}} & {\scalebox{0.8}{0.503}} &
    \secondres{\scalebox{0.8}{0.336}} & \secondres{\scalebox{0.8}{0.413}} &
    {\scalebox{0.8}{0.511}} & {\scalebox{0.8}{0.497}} &
    {\scalebox{0.8}{0.726}} & {\scalebox{0.8}{0.614}} &
    {\scalebox{0.8}{0.413}} & {\scalebox{0.8}{0.456}} \\
    \midrule 
    \scalebox{0.95}{PEMS04} &     
    \boldres{\scalebox{0.8}{0.284}} & \boldres{\scalebox{0.8}{0.376}} &
    {\scalebox{0.8}{0.406}} & {\scalebox{0.8}{0.455}} &
    {\scalebox{0.8}{0.624}} & {\scalebox{0.8}{0.575}} &
    {\scalebox{0.8}{0.441}} & {\scalebox{0.8}{0.494}} &
    {\scalebox{0.8}{0.471}} & {\scalebox{0.8}{0.507}} &
    \secondres{\scalebox{0.8}{0.381}} & \secondres{\scalebox{0.8}{0.441}} &
    {\scalebox{0.8}{0.586}} & {\scalebox{0.8}{0.548}} &
    {\scalebox{0.8}{0.800}} & {\scalebox{0.8}{0.652}} &
    {\scalebox{0.8}{0.452}} & {\scalebox{0.8}{0.485}} \\
    \midrule 
    \scalebox{0.95}{PEMS08} &     
    \boldres{\scalebox{0.8}{0.508}} & \boldres{\scalebox{0.8}{0.448}} &
    {\scalebox{0.8}{0.598}} & {\scalebox{0.8}{0.498}} &
    {\scalebox{0.8}{0.719}} & {\scalebox{0.8}{0.573}} &
    {\scalebox{0.8}{0.69}} & {\scalebox{0.8}{0.556}} &
    {\scalebox{0.8}{0.758}} & {\scalebox{0.8}{0.61}} &
    \secondres{\scalebox{0.8}{0.521}} & \secondres{\scalebox{0.8}{0.462}} &
    {\scalebox{0.8}{0.724}} & {\scalebox{0.8}{0.557}} &
    {\scalebox{0.8}{0.956}} & {\scalebox{0.8}{0.664}} &
    {\scalebox{0.8}{0.621}} & {\scalebox{0.8}{0.511}} \\
    \midrule 
    \scalebox{0.95}{Count} &     
    \boldres{\scalebox{0.8}{6}} & \boldres{\scalebox{0.8}{6}} &
    {\scalebox{0.8}{0}} & {\scalebox{0.8}{0}} &
    {\scalebox{0.8}{0}} & {\scalebox{0.8}{0}} &
    {\scalebox{0.8}{0}} & {\scalebox{0.8}{0}} &
    {\scalebox{0.8}{0}} & {\scalebox{0.8}{0}} &
    \secondres{\scalebox{0.8}{3}} & \secondres{\scalebox{0.8}{1}} &
    {\scalebox{0.8}{1}} & \secondres{\scalebox{0.8}{1}} &
    {\scalebox{0.8}{0}} & {\scalebox{0.8}{0}} &
    {\scalebox{0.8}{0}} & {\scalebox{0.8}{0}} \\
    
    \bottomrule 
  \end{tabular}
    \end{small}
  \end{threeparttable}
}

\end{table}

\subsection{Short-term Forecasting}
Short-term time series forecasting tasks is also prevalent in many real-world applications. Table~\ref{tab:shortterm_avg} presents the experimental results on the PEMS (PEMS03, PEMS04, and PEMS08) datasets.
The proposed HTMformer demonstrates BEST performance across all three PEMS datasets, outperforming the second-best model, MultiPatchFormer, with a \emph{21.7\%} reduction in MSE and a \emph{12.0\%} reduction in MAE.
We attribute this performance gain to the exploitation of rich multivariate representations in HTME. 
Since short-term forecasting, constrained by the limited prediction length, does not entail intricate long temporal dependencies, it renders inter-variable correlations more salient.
These results further demonstrate HTMformer’s strong and robust performance, complementing its long-term results.
\begin{table}[htbp]
  \vspace{-5pt}
  \caption{Short-term forecasting results, averaged across four prediction lengths $H \in \{12, 24, 48, 96\}$ with a fixed lookback window of $L = 96$. Full results are available in \underline{Appendix~\ref{app:F 3}}.}
  \vspace{5.5pt}
  \centering
  \label{tab:shortterm_avg}
  \resizebox{1\columnwidth}{!}{
  \begin{threeparttable}
  \begin{small}
  \renewcommand{\multirowsetup}{\centering}
  \setlength{\tabcolsep}{1.45pt}
  \begin{tabular}{c|cc|cc|cc|cc|cc|cc|cc|cc}
    \toprule
    \multicolumn{1}{c}{\multirow{2}{*}{Models}} & 
    \multicolumn{2}{c}{\rotatebox{0}{\scalebox{0.8}{\textbf{HTMformer}}}} &
    \multicolumn{2}{c}{\rotatebox{0}{\scalebox{0.8}{iTransformer}}} &
    \multicolumn{2}{c}{\rotatebox{0}{\scalebox{0.8}{PatchTST}}} &
    \multicolumn{2}{c}{\rotatebox{0}{\scalebox{0.8}{DLinear}}} &
    \multicolumn{2}{c}{\rotatebox{0}{\scalebox{0.8}{FEDformer}}} &
    \multicolumn{2}{c}{\rotatebox{0}{\scalebox{0.8}{MPFormer}}} &
    \multicolumn{2}{c}{\rotatebox{0}{\scalebox{0.8}{WPMixer}}} & 
    \multicolumn{2}{c}{\rotatebox{0}{\scalebox{0.8}{TimeMixer}}} \\

    \multicolumn{1}{c}{} & 
    \multicolumn{2}{c}{\scalebox{0.8}{(\textbf{Ours})}} &
    \multicolumn{2}{c}{\scalebox{0.8}{(\citeyear{liu2024itransformer})}} &
    \multicolumn{2}{c}{\scalebox{0.8}{\citeyear{nie2023time}}} &
    \multicolumn{2}{c}{\scalebox{0.8}{\citeyear{zeng2023dlinear}}}&
    \multicolumn{2}{c}{\scalebox{0.8}{\citeyear{zhou2022fedformer}}}&
    \multicolumn{2}{c}{\scalebox{0.8}{\citeyear{naghashi2025multipatchformer}}}&
    \multicolumn{2}{c}{\scalebox{0.8}{\citeyear{murad2024wp}}}&
    \multicolumn{2}{c}{\scalebox{0.8}{\citeyear{wang2023timemixer}}}\\
    \midrule
    \multicolumn{1}{c|}{Metric} & MSE & MAE & MSE & MAE& MSE & MAE& MSE & MAE & MSE & MAE & MSE & MAE& MSE & MAE& MSE & MAE\\
    \midrule
    \scalebox{0.95}{PEMS03} &     
    \boldres{\scalebox{0.8}{0.138}} & \boldres{\scalebox{0.8}{0.245}} &
    {\scalebox{0.8}{0.181}} & {\scalebox{0.8}{0.283}} &
    {\scalebox{0.8}{0.267}} & {\scalebox{0.8}{0.350}} &
    {\scalebox{0.8}{0.278}} & {\scalebox{0.8}{0.377}} &
    {\scalebox{0.8}{0.206}} & {\scalebox{0.8}{0.325}} &
    \secondres{\scalebox{0.8}{0.172}} & \secondres{\scalebox{0.8}{0.278}} &
    {\scalebox{0.8}{0.262}} & {\scalebox{0.8}{0.336}} &
    {\scalebox{0.8}{0.348}} & {\scalebox{0.8}{0.391}} \\
    \midrule 
    \scalebox{0.95}{PEMS04} &     
    \boldres{\scalebox{0.8}{0.138}} & \boldres{\scalebox{0.8}{0.249}} &
    {\scalebox{0.8}{0.216}} & {\scalebox{0.8}{0.308}} &
    {\scalebox{0.8}{0.323}} & {\scalebox{0.8}{0.385}} &
    {\scalebox{0.8}{0.295}} & {\scalebox{0.8}{0.388}} &
    \secondres{\scalebox{0.8}{0.206}} & {\scalebox{0.8}{0.328}} &
    {\scalebox{0.8}{0.207}} & \secondres{\scalebox{0.8}{0.305}} &
    {\scalebox{0.8}{0.304}} & {\scalebox{0.8}{0.369}} &
    {\scalebox{0.8}{0.387}} & {\scalebox{0.8}{0.426}} \\
    \midrule 
    \scalebox{0.95}{PEMS08} &     
    \boldres{\scalebox{0.8}{0.187}} & \boldres{\scalebox{0.8}{0.278}} &
    {\scalebox{0.8}{0.238}} & {\scalebox{0.8}{0.31}} &
    {\scalebox{0.8}{0.297}} & {\scalebox{0.8}{0.361}} &
    {\scalebox{0.8}{0.377}} & {\scalebox{0.8}{0.414}} &
    {\scalebox{0.8}{0.286}} & {\scalebox{0.8}{0.359}} &
    \secondres{\scalebox{0.8}{0.213}} & \secondres{\scalebox{0.8}{0.296}} &
    {\scalebox{0.8}{0.314}} & {\scalebox{0.8}{0.360}} &
    {\scalebox{0.8}{0.392}} & {\scalebox{0.8}{0.408}} \\
    \midrule 
    \scalebox{0.95}{Count} &     
    \boldres{\scalebox{0.8}{3}} & \boldres{\scalebox{0.8}{3}} &
    {\scalebox{0.8}{0}} & {\scalebox{0.8}{0}} &
    {\scalebox{0.8}{0}} & {\scalebox{0.8}{0}} &
    {\scalebox{0.8}{0}} & {\scalebox{0.8}{0}} &
    {\scalebox{0.8}{0}} & {\scalebox{0.8}{0}} &
    {\scalebox{0.8}{0}} & {\scalebox{0.8}{0}} &
    {\scalebox{0.8}{0}} & {\scalebox{0.8}{0}} &
    {\scalebox{0.8}{0}} & {\scalebox{0.8}{0}} \\
    \bottomrule 
  \end{tabular}
    \end{small}
  \end{threeparttable}
}
\end{table}


\section{Model Analyses}
\subsection{Ablation Studies}
To evaluate the contribution of each component in HTME, we conduct ablation studies (see \underline{Appendix~\ref{app:G}}). 
Specifically, we examined three different variants of HTMProductors:
1) \textbf{Productor}, the baseline forecasting model.
2) \textbf{ProductorV1}, which utilizes only the multivariate feature extraction module of the HTME extractor. 
3) \textbf{ProductorV2}, which utilizes the multivariate feature extraction module of the HTME extractor to supplement the embedding representations.
We further introduce two variants: \textbf{HTMformer}, builds upon iTransformer by incorporating the complete HTME extractor; \textbf{HTMformerV1}, replaces the original embedding layer in iTransformer with the temporal feature extractor from HTME.
We adopt iTransformer~\citep{liu2024itransformer} as the state-of-the-art baseline.  
The results are summarized in Table~\ref{tab:ablation1}.

\begin{table}[t]
  \vspace{-5,5pt}
  \caption{Results (Average) of different variants under prediction lengths $H \in \{96, 192, 336, 720\}$. The input lookback $L$ is set to 96. }
  \vspace{5.5pt}
  \centering
  \label{tab:ablation1}
  \resizebox{0.9\columnwidth}{!}{
  \begin{threeparttable}
  \begin{small}
  \renewcommand{\multirowsetup}{\centering}
  \setlength{\tabcolsep}{2.4pt}
  \begin{tabular}{c|cc|cc|cc|cc|cc|cc}
    \toprule
    \multicolumn{1}{c}{\multirow{1}{*}{Models}} & 
    \multicolumn{2}{c}{\rotatebox{0}{\scalebox{1}{HTMformer}}} &
    \multicolumn{2}{c}{\rotatebox{0}{\scalebox{1}{HTMformer-A}}} &
    \multicolumn{2}{c}{\rotatebox{0}{\scalebox{1}{iTransV2}}} &
    \multicolumn{2}{c}{\rotatebox{0}{\scalebox{1}{iTransformer}}} &
    \multicolumn{2}{c}{\rotatebox{0}{\scalebox{1}{iTransV1}}} &
    \multicolumn{2}{c}{\rotatebox{0}{\scalebox{1}{MPFormer}}}\\
    \midrule
    \multicolumn{1}{c|}{Metric} & MSE & MAE & MSE & MAE & MSE & MAE& MSE & MAE& MSE & MAE& MSE & MAE\\
    \midrule
    \scalebox{1}{ECL}  &
    \scalebox{0.8}{0.185}& \textcolor{red}{\scalebox{0.8}{0.272}}& \textcolor{blue}{\scalebox{0.8}{0.184}}& \textcolor{blue}{\scalebox{0.8}{0.273}}& \scalebox{0.8}{0.275}& \scalebox{0.8}{0.357}& \scalebox{0.8}{0.244}& \scalebox{0.8}{0.290}& \textcolor{red}{\scalebox{0.8}{0.194}}& \textcolor{red}{\scalebox{0.8}{0.288}}& \scalebox{0.8}{0.186}& \scalebox{0.8}{0.273} \\
    \midrule 
    \scalebox{1}{Weather}  &
    \scalebox{0.8}{0.254}& \textcolor{red}{\scalebox{0.8}{0.277}}& \textcolor{blue}{\scalebox{0.8}{0.253}}& \textcolor{blue}{\scalebox{0.8}{0.278}}& \scalebox{0.8}{0.262}& \scalebox{0.8}{0.290}& \scalebox{0.8}{0.259}& \scalebox{0.8}{0.279}& \textcolor{red}{\scalebox{0.8}{0.252}}& \textcolor{red}{\scalebox{0.8}{0.267}}& \scalebox{0.8}{0.252}& \scalebox{0.8}{0.274} \\
    \midrule 
    \scalebox{1}{Traffic}  &
    \textcolor{red}{\scalebox{0.8}{0.467}}& \textcolor{red}{\scalebox{0.8}{0.312}}& \scalebox{0.8}{0.477}& \scalebox{0.8}{0.318}& \scalebox{0.8}{0.748}& \scalebox{0.8}{0.440}& \scalebox{0.8}{0.474}& \scalebox{0.8}{0.317}& \textcolor{red}{\scalebox{0.8}{0.473}}& \scalebox{0.8}{0.321}& \scalebox{0.8}{0.462}& \scalebox{0.8}{0.304} \\
    \midrule
    \scalebox{1}{ETTh2}  &
    \textcolor{red}{\scalebox{0.8}{0.379}}& \textcolor{red}{\scalebox{0.8}{0.399}}& \textcolor{blue}{\scalebox{0.8}{0.381}}& \textcolor{blue}{\scalebox{0.8}{0.405}}& \scalebox{0.8}{0.444}& \scalebox{0.8}{0.444}& \scalebox{0.8}{0.383}& \scalebox{0.8}{0.406}& \scalebox{0.8}{0.384}& \textcolor{red}{\scalebox{0.8}{0.406}}& \scalebox{0.8}{0.381}& \scalebox{0.8}{0.406} \\
    \midrule 
    \scalebox{1}{Solar}  &
    \textcolor{red}{\scalebox{0.8}{0.235}} & \textcolor{red}{\scalebox{0.8}{0.265}} & \textcolor{blue}{\scalebox{0.8}{0.243}} & \textcolor{blue}{\scalebox{0.8}{0.278}} & \scalebox{0.8}{0.288} & \scalebox{0.8}{0.311} & \scalebox{0.8}{0.246} & \scalebox{0.8}{0.278} & \textcolor{red}{\scalebox{0.8}{0.244}} & \scalebox{0.8}{0.280}& \scalebox{0.8}{0.235}& \scalebox{0.8}{0.267} \\
    \midrule 
    \scalebox{1}{PEMS03}  &
    \textcolor{red}{\scalebox{0.8}{0.289}} & \textcolor{red}{\scalebox{0.8}{0.369}} & \textcolor{blue}{\scalebox{0.8}{0.340}} & \textcolor{blue}{\scalebox{0.8}{0.410}} & \scalebox{0.8}{0.351} & \scalebox{0.8}{0.403} & \scalebox{0.8}{0.360} & \scalebox{0.8}{0.421} & \textcolor{red}{\scalebox{0.8}{0.294}} & \textcolor{red}{\scalebox{0.8}{0.375}}& \scalebox{0.8}{0.336}& \scalebox{0.8}{0.413} \\
    \midrule 
    \scalebox{1}{PEMS04}  &
    \textcolor{red}{\scalebox{0.8}{0.284}} & \textcolor{red}{\scalebox{0.8}{0.376}} & \textcolor{blue}{\scalebox{0.8}{0.391}} & \textcolor{blue}{\scalebox{0.8}{0.431}} & \scalebox{0.8}{0.315} & \scalebox{0.8}{0.402} & \scalebox{0.8}{0.406} & \scalebox{0.8}{0.455} & \textcolor{red}{\scalebox{0.8}{0.300}} & \textcolor{red}{\scalebox{0.8}{0.386}}& \scalebox{0.8}{0.381}& \scalebox{0.8}{0.441} \\
    \midrule 
    \scalebox{1}{PEMS08}  &     
    \textcolor{red}{\scalebox{0.8}{0.508}} & \textcolor{red}{\scalebox{0.8}{0.448}} & \textcolor{blue}{\scalebox{0.8}{0.594}} & \textcolor{blue}{\scalebox{0.8}{0.492}} & \scalebox{0.8}{0.558} & \scalebox{0.8}{0.465} & \scalebox{0.8}{0.598} & \scalebox{0.8}{0.498} & \textcolor{red}{\scalebox{0.8}{0.485}} & \textcolor{red}{\scalebox{0.8}{0.434}}& \scalebox{0.8}{0.521}& \scalebox{0.8}{0.462} \\
    \bottomrule 
  \end{tabular}
    \end{small}
  \end{threeparttable}
}
\end{table}

\paragraph{Effect of the temporal extractor.} The experimental results show that \textbf{HTMformer-A} and \textbf{iTransformer} outperform \textbf{iTransformerV1} in most cases. This suggests that the primary features of time series are embedded within the temporal dimension. Furthermore, \textbf{HTMformer-A} demonstrates superior performance over \textbf{iTransformer} across the majority of  which showwd in \textcolor{blue}{blue}. This indicates that deeply mining temporal patterns can further enhance prediction performance. These findings underscore the dominant role of temporal features in time series forecasting. 
\paragraph{Effect of the hybrid strategy.}
On the PEMS dataset, which exhibits strong multivariate correlations, \text{PredictorV2} achieved superior performance. This indicates that multivariate correlation is a significant pattern in prediction. However, for most datasets, multivariate feature extraction can introduce substantial noise.

\paragraph{Effect of the hybrid strategy.} We further conduct comparative evaluations between \textbf{Predictors} and \textbf{PredictorV1}. 
\textcolor{red}{Red} numbers indicate cases where the versions augmented with multivariate integration in the embedding layer outperform their original counterparts. 
Across most datasets, \textbf{PredictorV1} consistently achieved superior performance. 
This provides strong evidence for the effectiveness of our proposed strategy for incorporating multivariate features into embedding representations.
The multivariate features consistently enriched the temporal feature embeddings.

\subsection{Model Efficiency}
To assess the computational efficiency of HTME, we compare the floating-point operations (FLOPs) and the number of parameters of iTransformer, DLinear, RLinear, and SegRNN against their HTME versions. Owing to its carefully designed architecture, the computational overhead introduced by the HTME multivariate feature extraction module is nearly constant: for complex Transformer-based models it adds negligible extra cost, and for simple models the overall overhead remains modest.

\begin{figure}[t]  
    \centering  
    \includegraphics[width=1\textwidth]{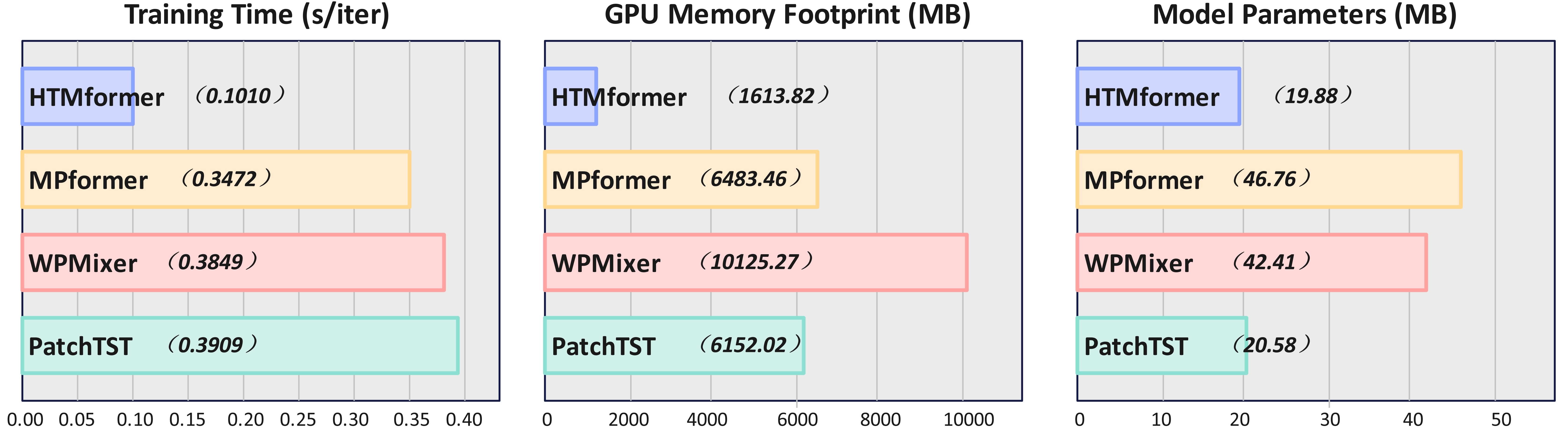}  
    \vspace{-11pt}
    \caption{We compare HTMformer with three state-of-the-art models on eight benchmark datasets. The input sequence length is uniformly set to 96, and the prediction sequence length is 192.
    }  
    \vspace{-11pt}
    \label{fig:figure3} 
\end{figure}

We perform a comparative analysis of HTMformer against MultiPatchFormer~\citep{naghashi2025multipatchformer}, and PatchTST~\citep{nie2023time}. 
As shown in Figure~\ref{fig:figure3}, taking the state-of-the-art MultiPatchFormer as the reference baseline, HTMformer reduces training runtime to roughly one third while using only 20\%–45\% of the GPU memory.
Moreover, HTMformer's model parameters are just half those of MultiPatchFormer.
Importantly, these gains are achieved without sacrificing predictive accuracy, as HTMformer consistently attains superior or comparable MSE and MAE across eight benchmark datasets. 
The HTME strategy enables simple backbones to achieve performance comparable to much more complex architectures by introducing only a lightweight module~(see \underline{Appendix~\ref{app:H}}).

\subsection{Analysis of HTME}
To better understand the feature space induced by HTME, we visualize the parameters defined in Equation~(6), which maps multivariate feature embeddings onto the temporal feature space.
The visualization shows that temporal features are consistently assigned substantially larger weights than multivariate features.
The vast majority of features in time series are contained within the temporal dimension~\citep{hyndman2021forecasting}. 
The multivariate dimension also contains a number of features~\citep{Pravilovic2013SpatialForecasting}. However, it does not warrant the use of an additional complex module for its modeling.
In the visualization, temporal features exhibit a line-shaped distribution, indicating a consistent temporal correlation within the series. In contrast, multivariate features appear as scattered point-like clusters, emphasizing a few salient variables.
With this design, HTME augments temporal embeddings with multivariate signals: the embeddings fed into the main backbone are still treated as temporal features, but they are now enriched with additional semantic information, thereby enhancing the backbone’s modeling capacity.
This approach avoids introducing extra modules into the backbone network and ensures the consistency of the feature space~(see \underline{Appendix~\ref{app:feature}}).

\section{Conclusion and Limitations}
In this work, we propose the HTME strategy.
The HTME module jointly encodes temporal dynamics and inter-variate dependencies present in the raw sequences, yielding embedding representations that are richer and more informative than those produced by prior designs. This allows the predictor to avoid introducing additional modules for modeling multivariate correlations. Consequently, HTME consistently enhances the modeling capabilities of various models, enabling lightweight models to achieve performance comparable to that of complex ones.


However, time series data exhibit complex correlations, and simply adding multivariate features to the temporal dimension through the embedding layer cannot fully model such a complex pattern.
Jointly modeling spatiotemporal dependencies in time series remains an open problem that requires further exploration.

\section{AI Usage Statement}
AI tools were used only for minor language polishing to standardize expressions.
No AI assistance was employed for idea generation, data analysis, experiment design, or content creation. 
All intellectual contributions, results, and conclusions are the authors' own.
\section{Reproducibility Statement}
All relevant implementation details, such as dataset specifications, main model hyperparameters, and experimental setups, are provided in the Appendix. 
The source code will be released publicly once the manuscript is accepted.
\section{Ethics Statement}
This study only focuses on the domain of time series forecasting.
All datasets employed are widely recognized, real-world, and publicly available. 
Therefore, there is no potential ethical risk.
\section{Acknowledgements}
\bibliography{iclr2026_conference}
\bibliographystyle{iclr2026_conference}

\newpage

\appendix
The appendix is divided into several sections, each giving extra information and details.

\startcontents[sections]
\printcontents[sections]{}{1}{\setcounter{tocdepth}{2}}
\vskip 0.2in
\hrule
\vspace{22pt}
\section{Discussion on Multivariate Feature Extraction}\label{app:A}
Recent studies~\citep{han2023capacity,li2023revisiting} have found that Channel Dependence ideally gains from higher capacity, while Channel Independence can significantly enhance performance due to sample scarcity. 
As most current forecasting benchmarks are not sufficiently large, the complex design of the Channel Dependence model makes it prone to capturing spurious correlations from noise and overemphasizing coincidental patterns in the training data.
In time series forecasting, an increasing number of models employ Channel Independence~\citep{nie2023time} or exclusively focus on modeling the temporal dimension. 
These models achieved good performance. 
However, they inevitably overlook the multivariate correlations, resulting in the loss of some information. 
We believe that multivariate correlations should not be ignored. 
Past hybrid models explicitly model multivariate correlations by incorporating various extra and complex modules within their backbone networks.
This strategy mixes temporal and multivariate features within the same space. However, constrained by the sample size, the multivariate extraction process can introduce significant noise and spurious patterns. Moreover, the additional multivariate features dilute the temporal features that should be dominant, leading to a disorganized feature space.
Moreover, the additional and complex module for multivariate feature extraction introduces significant computational overhead.
\section{Experiments of Different Embedding Methods}\label{app:C}
\begin{figure}[ht] 
    \centering  
    \includegraphics[width=1.0\textwidth, height=0.5\textheight]{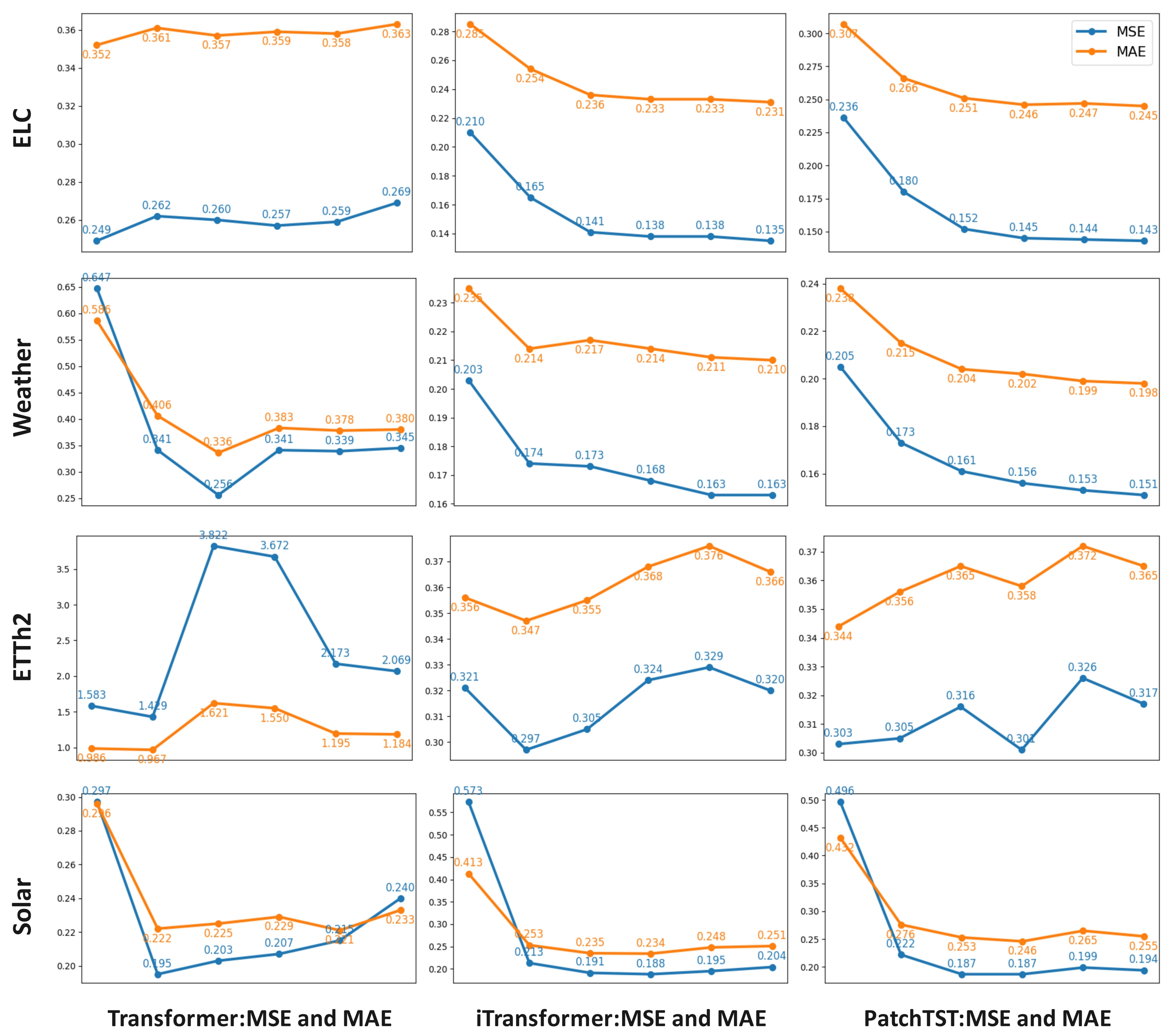}  
    \caption{On the ECL, Weather, ETTh2, and Solar, we compared three models with input lengths of {48, 96, 144, 192, 240, 288} and an output length of 96. The evaluation metrics were MSE and MAE.}  
    \label{fig:myplot2}  
\end{figure}

To demonstrate that Transformer-based forecasters have limited ability to understand sequences and the necessity of deeply mining sequence features in the embedding layer, we compared the prediction performance of Transformer~\citep{vaswani2017attention}, iTransformer~\citep{liu2024itransformer}, and PatchTST~\citep{nie2023time} with different embedding approaches on four datasets: ECL, Weather, ETTh2, and Solar.
The input sequence length is set to 48, 96, 144, 192, 240, and 288, with the prediction length being 96. The experimental results are shown in Figure~\ref{fig:myplot2}.

Experimental results reveal that employing more sophisticated embedding methods to capture complex spatiotemporal dependencies can effectively enhance the performance of Transformers.
These findings suggest that Transformer-based models exhibit limitations in proficiency in capturing effective features.
This limitation arises because the embedding layer fails to effectively represent the complex characteristics of time series. Constructing a sophisticated embedding method to extract semantically richer representations from the raw sequence can significantly enhance the performance of Transformers.

We also noticed that increasing the input sequence length initially leads to a significant decline in prediction loss. 
However, as the sequence length continues to grow, this benefit markedly lessens, and in some cases, prediction loss may even increase. 
Simply increasing the lookback window does not consistently capture more informative patterns and will also incur additional computational overhead.
This is because time series forecasting focuses on capturing short-term dependencies~\citep{kim2025comprehensivesurveydeeplearning} and the temporal dimension inherently offers limited contextual information.
Therefore, we argue that it is highly beneficial to consider multivariate correlations during the embedding stage. Doing so can greatly enrich the information content of the embedded representations and, in turn, enhance the modeling capability of subsequent modules.

Motivated by these findings, we propose a novel strategy that moderately extracts features along the multivariate dimension to enhance informative content, thereby enabling the efficient acquisition of high-quality embeddings.
\section{Detail of HTME}\label{app:HTME}
To demonstrate the correctness of the HTME extractor, we provide a detailed theoretical explanation for each of its steps.
The HTME temporal feature extraction module consists of four steps: Patching, Stacking, short-term feature extraction, and long-term feature extraction.
The purpose of the Patching step is to segment the input into short-term temporal sequences. The Stacking step is inspired by the concept of channel independence, which has been proven to be more effective (see Appendix A).
Time series data inherently exhibit strong short-term correlations and weaker long-term dependencies (see Appendix C). Motivated by this, we first employ convolutional layers to extract multi-scale representations from the short-term sequences, followed by a linear layer to capture local features. Subsequently, another linear layer is used to model the long-term dependencies among these features. This hierarchical process yields a more informative temporal feature representation.

The key to multivariate feature extraction is to reduce overfitting and mitigate noise (see Appendix A).
For the multivariate feature extraction module, we adopt the exact same logic and methodology. It is worth noting that we cannot simply treat the variables as being independent of the temporal dimension.
We concatenate the multivariate data over short time windows and then use a linear layer to extract features, to account for the effect of time delays. A change occurring at one node may exert influence on other nodes at the subsequent time step.
This extraction strategy, which fuses short timesteps, enables the model to deeply consider the complex multivariate correlations within time series, thereby effectively suppressing over-attention to coincidental patterns.
\begin{align*}
    N^{ (L-K), 128} &= \text{Linear}(T^{(L-K), K \times N})
\end{align*}
Subsequently, we employ a limited number of convolutional layers to extract multivariate features at different scales. Fewer convolutional kernels allows the model to focus on significant patterns, thereby substantially mitigating overfitting.
Then, a specialized linear layer that GRU network integrated multi-scale features to further extract effective features.
\begin{align*}
    B^N &= \text{GRU}(\text{Conv}(N^{ (L-K), 128}))
\end{align*}
Finally, small convolutional kernels are used to fine-tune the multivariate features along the temporal dimension, which prevents the model from learning fixed patterns.
\begin{align*}
    V_{\text{out}} &= \text{Conv}(B^N)
\end{align*}

Finally, we employ a parameter to adjust the weights of the two spaces. This parameter is initialized with a large value to guide subsequent modules to focus primarily on temporal features. During training, HTME adaptively considers the influence of multivariate features to generate high-quality embedding representations.
\begin{align*}
Y_{\text{in}} &= \alpha D_{\text{out}} + (1 - \alpha) V_{\text{out}},\label{eq:11}
\end{align*}
\section{Detailed Experiment Settings}\label{app:E}
\subsection{Dataset Descriptions}\label{app:E 1}

We perform extensive experiments on eight real-world datasets to evaluate the effectiveness of the proposed HTME strategy. These datasets cover diverse domains and temporal resolutions, including:
\begin{itemize}
  \item \textbf{ECL.} Contains hourly electricity consumption data of 321 clients from 2012 to 2014.

  \item \textbf{Weather.} Includes 21 meteorological variables collected every 10 minutes in 2020 from the meteorological station at the Max Planck Institute for Biogeochemistry.

  \item \textbf{Traffic.} Consists of hourly road occupancy rates collected from 862 sensors on highways in the San Francisco Bay Area between January 2015 and December 2016.

  \item \textbf{ETTh.} Comprises hourly electricity transformer temperature readings collected over two years from two counties in China, with seven recorded variables.

  \item \textbf{Solar-Energy.} Records 10-minute solar power production records of 137 photovoltaic (PV) plants in Alabama from the year 2006.

  \item \textbf{PEMS.} Includes PEMS03, PEMS04, and PEMS08 datasets, each containing public transportation network data collected at 5-minute intervals in California.
\end{itemize}

To ensure the consistency of our experimental approach and make sure there are no data leakage issues, we split the ETT and PEMS datasets into the training, validation, and test sets, with a 6:2:2 ratio and other datasets with a 7:1:2 ratio, following the protocols recommended in recent literature~\citep{wu2021autoformer,zeng2023dlinear}. 
In all forecasting experiments, the lookback window was fixed at 96 time steps, with prediction horizons of {96, 192, 336, 720} for the long-term experiments and {12, 24, 48, 96} for the short-term experiments. 
The details of all datasets are provided in Table~\ref{tab:dataset2}.
\begin{table}[ht]
  \caption{Detailed dataset descriptions. Dim denotes the number of variables in each dataset. Dataset Size indicates the number of time points in splits respectively. Prediction Length is the number of future time points to forecast. Frequency denotes the sampling interval of the time points.} 
  \label{tab:dataset2}  
  \vspace{5.5pt}
  \centering
  \resizebox{1.0\columnwidth}{!}{
  \begin{threeparttable}
  \begin{small}
  \renewcommand{\multirowsetup}{\centering}
  \setlength{\tabcolsep}{3.8pt}
  \begin{tabular}{l|c|c|c|c|c|c}
    \toprule
    Dataset & Dim & Prediction Length & Dataset Size & Frequency & Forecastability$\ast$ & \scalebox{0.8}{Information} \\
    \toprule
    ETTh2 & 7 & \scalebox{0.8}{\{96, 192, 336, 720\}} & (8545, 2881, 2881) & 15 min & 0.45 & \scalebox{0.8}{Temperature} \\
    \cmidrule{1-7}
    Electricity & 321 & \scalebox{0.8}{\{96, 192, 336, 720\}} & (18317, 2633, 5261) & Hourly & 0.77 & \scalebox{0.8}{Electricity} \\
    \cmidrule{1-7}
    Traffic & 862 & \scalebox{0.8}{\{96, 192, 336, 720\}} & (12185, 1757, 3509) & Hourly & 0.68 & \scalebox{0.8}{Transportation} \\
    \cmidrule{1-7}
    Weather & 21 & \scalebox{0.8}{\{96, 192, 336, 720\}} & (36792, 5271, 10540) & 10 min & 0.75 & \scalebox{0.8}{Weather} \\
    \cmidrule{1-7}
    Solar-Energy & 137 & \scalebox{0.8}{\{96, 192, 336, 720\}} & (36601, 5161, 10417) & 10 min & 0.33 & \scalebox{0.8}{Solar Energy} \\ 
    \midrule
    PEMS03 & 358 & \scalebox{0.8}{\{12, 24, 48, 96, 192, 336, 720\}} & (15617, 5135, 5135) & 5 min & 0.65 & \scalebox{0.8}{Transportation} \\
    \cmidrule{1-7}
    PEMS04 & 307 & \scalebox{0.8}{\{12, 24, 48, 96, 192, 336, 720\}} & (10172, 3375, 3375) & 5 min & 0.45 & \scalebox{0.8}{Transportation} \\
    \cmidrule{1-7}
    PEMS08 & 170 & \scalebox{0.8}{\{12, 24, 48, 96, 192, 336, 720\}} & (10690, 3548, 265) & 5 min & 0.52 & \scalebox{0.8}{Transportation} \\ 
    \bottomrule
  \end{tabular}
  \begin{tablenotes}
    \item $\ast$ The forecastability is calculated by one minus the entropy of Fourier decomposition of time series \citep{goerg2013forecastable}. A larger value indicates better predictability.
  \end{tablenotes}
  \end{small}
  \end{threeparttable}
  }
\end{table}

\subsection{Baseline Details}\label{app:E 2}
We employ four state-of-the-art Transformer-based models, such as iTransformer~\citep{liu2024itransformer}, PatchTST~\citep{nie2023time}, FEDformer~\citep{zhou2022fedformer}, and MultiPatchFormer~\citep{naghashi2025multipatchformer}, two well-acknowledged Linear models, Dlinear~\citep{zeng2023dlinear}, Nlinear, one prominent RNN-based model SegRNN~\citep{lin2023segrnn} and  WPMixer~\citep{murad2024wp}. 
We comprehensively evaluate the HTMformer framework using four mainstream attention mechanisms Transformer~\citep{vaswani2017attention}, Reformer~\citep{kitaev2020reformer}, Informer~\citep{zhou2021informer}, Flowformer~\citep{ma2023flowformer}, and Flashformer~\citep{dao2022flashattention}.
We utilized the official repositories of the Time-Series-Library~\citep{wang2024tssurvey} directly for reproduction, and adopted the same comparison settings.
The Time-Series-Library consolidates and integrates implementations of prior time-series forecasting models, making it convenient for us to conduct experiments.
Among these models, \textbf{MultiPatchFormer} is a highly advanced architecture and yields the best overall forecasting accuracy. 
\textbf{iTransformer} and HTMformers share the same architecture. HTMformers replaces the embedding layer of iTransformer.
Note that SegRNN is incompatible with short-term forecasting, so no comparison is conducted.

\subsection{Experiment Details}\label{app:E 3}

All experiments were implemented in PyTorch~\citep{paszke2019pytorch} and conducted on a dedicated NVIDIA Quadro RTX 6000 GPU. 
We trained the model for 10 epochs using the Adam optimizer~\citep{kingma2015adam}, with an early stopping mechanism configured at a patience value of 3.
The key parameter configurations are listed in Table~\ref{tab:params2}.  
We verified that, for these baselines, all other hyperparameters were taken from their official repositories within the Time-Series-Library~\citep{wang2024tssurvey}, which consolidates implementations of prior time-series forecasting models.
This ensures consistency with the established fair comparison protocol. 
Unless otherwise stated, all models adopt the aforementioned configurations across all experiments.
The only modification was to the output sequence lengths.

\begin{itemize}
    \item \textbf{dropout}: Dropout rate.
    \item \textbf{batch size}: Batch size used for training.
    \item \textbf{l\_rate}: The learning rate used in the optimization process.
    \item \textbf{d\_ff}: Dimension of Fully FFN
    \item \textbf{d\_model}: Dimension of model
    \item \textbf{e\_layers}: Number of Transformer encoder layers.
    \item \textbf{n\_heads}: Num of heads    
    \item \textbf{seq\_len}: Inputs sequence length
    \item \textbf{label\_len}: Start token length
    \item \textbf{loss}: Loss function used for training
\end{itemize}
\begin{table}[ht]
\centering
\Large
\vspace{-5.5pt}
\caption{Experiments of all models on all datasets follow the following experimental settings to ensure the accuracy of the experiments.} 
\vspace{5.5pt}
\label{tab:params2}  
\adjustbox{max width=\textwidth}{
  \begin{tabular}{cccccccccc}
    \toprule
    dropout & batch size & l\_rate & d\_ff & d\_model & e\_layers & n\_heads & seq\_lenl & label\_len & loss\\
    \midrule
    0.1    & 32      & 0.0001       & 1024        & 512     & 2     & 8    & 96    & 48  & L2\\
    \bottomrule
  \end{tabular}
}
\end{table}
\section{Full Results}\label{app:F}
\subsection{Performance Promotion with HTME}\label{app:F 1}
To assess the effectiveness and scalability of the HTMformers for enhancing the Attention's capacity to learn temporal sequence representations, we integrated various attention variants: Transformer~\citep{vaswani2017attention}, Reformer~\citep{kitaev2020reformer}, Informer~\citep{zhou2021informer}, Flowformer~\citep{ma2023flowformer}, and Flashformer~\citep{dao2022flashattention}. 
We apply the HTME strategy to models with other architectures, including the Transformer-based iTransformer, the linear-based DLinear, NLinear and the RNN-based SegRNN.  
The length of historical data sequences was set to 96 time steps, with the forecast horizon varying among 96, 192, 336, and 720 time steps. 
The full predictive outcomes are detailed in Table~\ref{tab:model_performance}.
On three benchmark datasets (ECL, Traffic, and Weather), the HTME versions yield significant improvements over all four attentions across all proposed prediction horizons. 
Across three mainstream forecasting architectures, the HTME strategy consistently improves performance by margins ranging from 1.7\% to 4.6\%.
These results highlight the effectiveness and robustness of the HTME strategy.
\vspace{-3pt}
\begin{table}[t]
\centering
\caption{Full results of original Transformers and Transformers with HTME.}
\vspace{5.5pt}
\resizebox{1\columnwidth}{!}{
\begin{threeparttable}
\begin{small}
\begin{tabular}{c|c|c|cc|cc|cc|cc|cc|cc|cc|cc|cc}
\toprule
\multicolumn{3}{c}{\multirow{1}{*}{Models}}  & \multicolumn{2}{c}{Transformer } & \multicolumn{2}{c}{Reformer } & \multicolumn{2}{c}{Informer } & \multicolumn{2}{c}{Flowformer } & \multicolumn{2}{c}{Flashformer }& \multicolumn{2}{c}{iTransformer }& \multicolumn{2}{c}{DLinear }& \multicolumn{2}{c}{Nlinear }& \multicolumn{2}{c}{SegRNN }\\
\midrule
\textbf{Dataset} & \textbf{Type} & \textbf{Step} & MSE & MAE & MSE & MAE & MSE & MAE & MSE & MAE & MSE & MAE & MSE & MAE & MSE & MAE & MSE & MAE& MSE & MAE\\
\midrule
\multirow{12}{*}{\textbf{Electricity}} 
& \multirow{5}{*}{Original} 
& 96  & 0.264 & 0.363 & 0.304 & 0.391 & 0.330 & 0.415 & 0.272 & 0.371 & 0.263 & 0.361 & 0.163 & 0.252 &0.210 &0.301&0.199 &0.277&0.168 &0.264\\
& & 192 & 0.275 & 0.375 & 0.350 & 0.426 & 0.356 & 0.438 & 0.269 & 0.371 & 0.270 & 0.369 & 0.175 & 0.263&0.210 &0.304&0.199 &0.280&0.183 &0.279\\
& & 336 & 0.262 & 0.362 & 0.354 & 0.429 & 0.376 & 0.454 & 0.264 & 0.364 & 0.263 & 0.360 &0.282 & 0.299&0.223 &0.319&0.214 &0.295&0.192 &0.289\\
& & 720 & 0.286 & 0.373 & 0.361 & 0.422 & 0.454 & 0.464 & 0.284 & 0.374 & 0.279 & 0.368 & 0.357 & 0.348&0.257 &0.349&0.256 &0.328&0.231 &0.321\\
\cmidrule(lr){3-21}
& & Avg & 0.271 & 0.363 & 0.342 & 0.417 & 0.379 & 0.442 & 0.272 & 0.370 & 0.268 & 0.364 & 0.244 &0.290 &0.225 &0.318&0.217 &0.295&0.194 &0.288\\
\cmidrule(lr){2-21}
& \multirow{5}{*}{+HTME} 
& 96  & 0.157 & 0.247 & 0.169 & 0.266 & 0.170 & 0.266 & 0.170 & 0.266 & 0.163 & 0.262 & 0.157 & 0.247 &0.204 &0.299&0.189 &0.273&0.164 &0.258\\
& & 192 & 0.171 & 0.260 & 0.190 & 0.283 & 0.187 & 0.279 & 0.188 & 0.280 & 0.183 & 0.277 & 0.171 & 0.260&0.211 &0.307&0.192 &0.273&0.175 &0.269\\
& & 336 & 0.192 & 0.277 & 0.200 & 0.293 & 0.207 & 0.296 & 0.202 & 0.294 & 0.199 & 0.292 & 0.192 & 0.277&0.222 &0.318&0.206 &0.290&0.192 &0.285\\
& & 720 & 0.220 & 0.304 & 0.248 & 0.329 & 0.247 & 0.329 & 0.235 & 0.319 & 0.229 & 0.316 & 0.220 & 0.304&0.24 &0.331&0.246 &0.329&0.230 &0.320\\
\cmidrule(lr){3-21}
& & Avg & 0.185 & 0.272 & 0.201 & 0.292 & 0.202 & 0.292 & 0.198 & 0.289 & 0.194 & 0.287 & 0.185 & 0.272 &0.219 &0.314&0.208 &0.291&0.190 &0.283\\
\midrule
\multirow{12}{*}{\textbf{Weather}} 
& \multirow{5}{*}{Original} 
& 96  & 0.372 & 0.417 & 0.311 & 0.355 & 0.420 & 0.452 & 0.283 & 0.364 & 0.326 & 0.395 & 0.176 & 0.216&0.193 &0.233&0.196 &0.237&0.168 &0.23\\
& & 192 & 0.632 & 0.580 & 0.453 & 0.464 & 0.423 & 0.455 & 0.548 & 0.541 & 0.587 & 0.544 & 0.223 & 0.255&0.236 &0.268&0.241 &0.279&0.215 &0.277\\
& & 336 & 0.699 & 0.606 & 0.476 & 0.474 & 0.588 & 0.539 & 0.614 & 0.548 & 0.714 & 0.614 & 0.280 & 0.298&0.288 &0.304&0.293 &0.317&0.274 &0.319\\
& & 720 & 0.903 & 0.688 & 0.579 & 0.541 & 1.146 & 0.784 & 1.079 & 0.783 & 0.906 & 0.684 & 0.353 & 0.346&0.359 &0.349&0.366 &0.361&0.359 &0.378\\
\cmidrule(lr){3-21}
& & Avg & 0.651 & 0.572 & 0.454 & 0.458 & 0.644 & 0.557 & 0.631 & 0.559 & 0.633 & 0.559 & 0.259 & 0.279&0.269 &0.289&0.274 &0.299&0.254 &0.301\\
\cmidrule(lr){2-21}
& \multirow{5}{*}{+HTME} 
& 96  & 0.163 & 0.209 & 0.164 & 0.209 & 0.170 & 0.213 & 0.167 & 0.212 & 0.168 & 0.215 & 0.163 & 0.209 &0.177 &0.223&0.168 &0.230&0.161 &0.226\\
& & 192 & 0.222 & 0.261 & 0.214 & 0.255 & 0.220 & 0.260 & 0.216 & 0.255 & 0.218 & 0.258 & 0.222 & 0.261&0.227 &0.266&0.224 &0.275&0.211 &0.273\\
& & 336 & 0.272 & 0.290 & 0.270 & 0.288 & 0.274 & 0.298 & 0.279 & 0.302 & 0.275 & 0.298 & 0.272 & 0.290&0.281 &0.303&0.295 &0.313&0.269 &0.315\\
& & 720 & 0.359 & 0.352 & 0.351 & 0.348 & 0.354 & 0.351 & 0.354 & 0.350 & 0.352 & 0.349 & 0.359 & 0.352&0.358 &0.352&0.359 &0.343&0.355 &0.371\\
\cmidrule(lr){3-21}
& & Avg & 0.254 & 0.277 & 0.249 & 0.275 & 0.254 & 0.280 & 0.254 & 0.279 & 0.253 & 0.380 & 0.254 & 0.277&0.261 &0.286&0.261 &0.290&0.249 &0.296\\
\midrule
\multirow{12}{*}{\textbf{Traffic}} 
& \multirow{5}{*}{Original} 
& 96  & 0.658 & 0.364 & 0.715 & 0.399 & 0.731 & 0.413 & 0.646 & 0.359 & 0.637 & 0.353 & 0.442 & 0.302&0.696 &0.428&0.654 &0.396&0.723 &0.376\\
& & 192 & 0.655 & 0.359 & 0.705 & 0.390 & 0.761 & 0.431 & 0.646 & 0.355 & 0.659 & 0.364 & 0.459 & 0.308&0.646 &0.407&0.607 &0.374&0.737 &0.385\\
& & 336 & 0.657 & 0.361 & 0.703 & 0.386 & 0.840 & 0.474 & 0.657 & 0.358 & 0.653 & 0.357 & 0.479 & 0.319&0.653 &0.409&0.615 &0.376&0.766 &0.405\\
& & 720 & 0.697 & 0.380 & 0.703 & 0.383 & 0.969 & 0.545 & 0.671 & 0.364 & 0.697 & 0.375 & 0.516 & 0.342&0.694 &0.428&0.654 &0.417&0.818 &0.423\\
\cmidrule(lr){3-21}
& & Avg & 0.666 & 0.366 & 0.706 & 0.389 & 0.825 & 0.465 & 0.655 & 0.359 & 0.661 & 0.362 & 0.474 & 0.317&0.672 &0.418&0.632 &0.391&0.761 &0.397\\
\cmidrule(lr){2-21}
& \multirow{5}{*}{+HTME} 
& 96  & 0.439 & 0.300 & 0.445 & 0.302 & 0.466 & 0.314 & 0.451 & 0.309 & 0.439 & 0.300 & 0.439 & 0.300&0.626 &0.393&0.624 &0.394&0.696 &0.366\\
& & 192 & 0.450 & 0.302 & 0.463 & 0.308 & 0.468 & 0.312 & 0.466 & 0.310 & 0.460 & 0.310 & 0.450 & 0.302&0.583 &0.369&0.588 &0.372&0.717 &0.382\\
& & 336 & 0.462 & 0.310 & 0.478 & 0.313 & 0.488 & 0.322 & 0.484 & 0.320 & 0.480 & 0.321 & 0.462 & 0.310&0.591 &0.373&0.597 &0.376&0.754 &0.401\\
& & 720 & 0.517 & 0.339 & 0.514 & 0.334 & 0.530 & 0.347 & 0.524 & 0.345 & 0.516 & 0.342 & 0.517 & 0.339&0.636 &0.395&0.635 &0.395&0.803 &0.419\\
\cmidrule(lr){3-21}
& & Avg & 0.467 & 0.312 & 0.475 & 0.314 & 0.488 & 0.323 & 0.481 & 0.321 & 0.474 & 0.319 & 0.467 & 0.312&0.609 &0.383&0.611 &0.384&0.750 &0.392\\
\bottomrule
\end{tabular}

\label{tab:model_performance}
\end{small}
\end{threeparttable}
}
\end{table}

\subsection{Long-term Forecasting Results}\label{app:F 2}
The results of the multivariate long-term forecasting tasks are summarized in Table~\ref{tab:longterm2}.
Lower Mean Squared Error (MSE) and Mean Absolute Error (MAE) values indicate superior predictive accuracy. 
The best results are highlighted in \boldres{red}, and the second-best results in \secondres{blue}.
Notably, HTMformer achieves state-of-the-art performance in most evaluation scenarios, ranking first in 26 out of 40 MSE metrics and 29 out of 40 MAE metrics.
Moreover, our model consistently ranks within the top two across the majority of evaluation scenarios.
Table~\ref{tab:seed1} also reports the standard deviations of HTMformer over five runs with different random seeds, demonstrating its performance stability.

\begin{table}[htbp]
  \caption{Full results for the long-term forecasting task. We compare extensive competitive models under different prediction lengths. \emph{Avg} is averaged from all four prediction lengths.}
  \vspace{5.5pt}
  \centering
  \label{tab:longterm2}
  \resizebox{1.0\columnwidth}{!}{
  \begin{threeparttable}
  \begin{small}
  \renewcommand{\multirowsetup}{\centering}
  \setlength{\tabcolsep}{1pt}
\begin{tabular}{c|c|cc|cc|cc|cc|cc|cc|cc|cc|cc}
    \toprule
    \multicolumn{2}{c}{\multirow{2}{*}{Models}} & 
    \multicolumn{2}{c}{\rotatebox{0}{\scalebox{0.8}{\textbf{HTMformer}}}} &
    \multicolumn{2}{c}{\rotatebox{0}{\scalebox{0.8}{iTransformer}}} &
    \multicolumn{2}{c}{\rotatebox{0}{\scalebox{0.8}{PatchTST}}} &
    \multicolumn{2}{c}{\rotatebox{0}{\scalebox{0.8}{DLinear}}} &
    \multicolumn{2}{c}{\rotatebox{0}{\scalebox{0.8}{FEDformer}}} &
    \multicolumn{2}{c}{\rotatebox{0}{\scalebox{0.8}{MPFormer}}} &
    \multicolumn{2}{c}{\rotatebox{0}{\scalebox{0.8}{WPMixer}}} & 
    \multicolumn{2}{c}{\rotatebox{0}{\scalebox{0.8}{TimeMixer}}} &
    \multicolumn{2}{c}{\rotatebox{0}{\scalebox{0.8}{SegRNN}}} \\ 
    
    \multicolumn{2}{c}{} & 
    \multicolumn{2}{c}{\scalebox{0.8}{(\textbf{Ours})}} &
    \multicolumn{2}{c}{\scalebox{0.8}{(\citeyear{liu2024itransformer})}} &
    \multicolumn{2}{c}{\scalebox{0.8}{\citeyear{nie2023time}}} &
    \multicolumn{2}{c}{\scalebox{0.8}{\citeyear{zeng2023dlinear}}}&
    \multicolumn{2}{c}{\scalebox{0.8}{\citeyear{zhou2022fedformer}}}&
    \multicolumn{2}{c}{\scalebox{0.8}{\citeyear{naghashi2025multipatchformer}}}&
    \multicolumn{2}{c}{\scalebox{0.8}{\citeyear{murad2024wp}}}&
    \multicolumn{2}{c}{\scalebox{0.8}{\citeyear{wang2023timemixer}}}&
    \multicolumn{2}{c}{\scalebox{0.8}{\citeyear{lin2023segrnn}}} \\ 
    
    \cmidrule(lr){3-4} \cmidrule(lr){5-6} \cmidrule(lr){7-8} \cmidrule(lr){9-10} 
    \cmidrule(lr){11-12} \cmidrule(lr){13-14} \cmidrule(lr){15-16} \cmidrule(lr){17-18} \cmidrule(lr){19-20}
    
    \multicolumn{2}{c}{Metric} & 
    \scalebox{0.78}{MSE} & \scalebox{0.78}{MAE} & 
    \scalebox{0.78}{MSE} & \scalebox{0.78}{MAE} & 
    \scalebox{0.78}{MSE} & \scalebox{0.78}{MAE} & 
    \scalebox{0.78}{MSE} & \scalebox{0.78}{MAE} & 
    \scalebox{0.78}{MSE} & \scalebox{0.78}{MAE} & 
    \scalebox{0.78}{MSE} & \scalebox{0.78}{MAE} & 
    \scalebox{0.78}{MSE} & \scalebox{0.78}{MAE} & 
    \scalebox{0.78}{MSE} & \scalebox{0.78}{MAE} & 
    \scalebox{0.78}{MSE} & \scalebox{0.78}{MAE} \\ 
    
    \toprule

    \multirow{5}{*}{\rotatebox{90}{\scalebox{0.95}{Electricity}}} 
    & \scalebox{0.85}{96} &
    \boldres{\scalebox{0.85}{0.157}}&\boldres{\scalebox{0.85}{0.247}}&
    {\scalebox{0.85}{0.163}}&{\scalebox{0.85}{0.252}}&
    {\scalebox{0.85}{0.180}}&{\scalebox{0.85}{0.266}}&
    {\scalebox{0.85}{0.210}}&{\scalebox{0.85}{0.301}}&
    {\scalebox{0.85}{0.195}}&{\scalebox{0.85}{0.309}}&
    \secondres{\scalebox{0.78}{0.159}}&\secondres{\scalebox{0.78}{0.249}}&
    {\scalebox{0.85}{0.170}}&{\scalebox{0.85}{0.259}}&
    {\scalebox{0.85}{0.213}}&{\scalebox{0.85}{0.301}}&
    {\scalebox{0.85}{0.191}}&{\scalebox{0.85}{0.278}} \\
    
    & \scalebox{0.85}{192} &
    \boldres{\scalebox{0.85}{0.171}}&\boldres{\scalebox{0.85}{0.260}}&
    \secondres{\scalebox{0.78}{0.175}}&\secondres{\scalebox{0.78}{0.263}}&
    {\scalebox{0.85}{0.185}}&{\scalebox{0.85}{0.271}}&
    {\scalebox{0.85}{0.210}}&{\scalebox{0.85}{0.304}}&
    {\scalebox{0.85}{0.202}}&{\scalebox{0.85}{0.315}}&
    \boldres{\scalebox{0.85}{0.171}}&\boldres{\scalebox{0.85}{0.260}}&
    {\scalebox{0.85}{0.180}}&{\scalebox{0.85}{0.267}}&
    {\scalebox{0.85}{0.225}}&{\scalebox{0.85}{0.316}}&
    {\scalebox{0.85}{0.202}}&{\scalebox{0.85}{0.287}} \\
    
    & \scalebox{0.85}{336} &
    \secondres{\scalebox{0.78}{0.192}}&\boldres{\scalebox{0.78}{0.277}}&
    {\scalebox{0.85}{0.282}}&{\scalebox{0.85}{0.299}}&
    {\scalebox{0.85}{0.202}}&{\scalebox{0.85}{0.288}}&
    {\scalebox{0.85}{0.223}}&{\scalebox{0.85}{0.319}}&
    {\scalebox{0.85}{0.229}}&{\scalebox{0.85}{0.342}}&
    \boldres{\scalebox{0.85}{0.188}}&\boldres{\scalebox{0.85}{0.277}}&
    {\scalebox{0.85}{0.196}}&\secondres{{\scalebox{0.85}{0.284}}}&
    {\scalebox{0.85}{0.242}}&{\scalebox{0.85}{0.334}}&
    {\scalebox{0.85}{0.221}}&{\scalebox{0.85}{0.305}} \\
    
    & \scalebox{0.85}{720} &
    \boldres{\scalebox{0.85}{0.220}}&\boldres{\scalebox{0.85}{0.304}}&
    {\scalebox{0.85}{0.357}}&{\scalebox{0.85}{0.348}}&
    {\scalebox{0.85}{0.241}}&{\scalebox{0.85}{0.319}}&
    {\scalebox{0.85}{0.257}}&{\scalebox{0.85}{0.349}}&
    {\scalebox{0.85}{0.264}}&{\scalebox{0.85}{0.367}}&
    \secondres{\scalebox{0.78}{0.228}}&\secondres{\scalebox{0.78}{0.309}}&
    {\scalebox{0.85}{0.238}}&{\scalebox{0.85}{0.317}}&
    {\scalebox{0.85}{0.285}}&{\scalebox{0.85}{0.363}}&
    {\scalebox{0.85}{0.261}}&{\scalebox{0.85}{0.339}} \\
    
    \cmidrule(lr){2-20}
    
    & \scalebox{0.85}{Avg} &
    \boldres{\scalebox{0.8}{0.185}}&\boldres{\scalebox{0.8}{0.272}}&
    {\scalebox{0.85}{0.244}}&{\scalebox{0.85}{0.290}}&
    {\scalebox{0.85}{0.202}}&{\scalebox{0.85}{0.286}}&
    {\scalebox{0.85}{0.225}}&{\scalebox{0.85}{0.318}}&
    {\scalebox{0.85}{0.222}}&{\scalebox{0.85}{0.333}}&
    \secondres{\scalebox{0.78}{0.186}}&\secondres{\scalebox{0.78}{0.273}}&
    {\scalebox{0.85}{0.196}}&{\scalebox{0.85}{0.281}}&
    {\scalebox{0.85}{0.241}}&{\scalebox{0.85}{0.328}}&
    {\scalebox{0.85}{0.218}}&{\scalebox{0.85}{0.302}} \\

    \midrule

    \multirow{5}{*}{\rotatebox{90}{\scalebox{0.95}{Weather}}} 
& \scalebox{0.85}{96} &
\boldres{\scalebox{0.78}{0.164}}&\boldres{\scalebox{0.78}{0.209}}&
{\scalebox{0.85}{0.176}}&\secondres{{\scalebox{0.85}{0.216}}}&
{\scalebox{0.85}{0.174}}&{\scalebox{0.85}{0.217}}&
{\scalebox{0.85}{0.196}}&{\scalebox{0.85}{0.256}}&
{\scalebox{0.85}{0.224}}&{\scalebox{0.85}{0.304}}&
{\scalebox{0.85}{0.168}}&\boldres{\scalebox{0.85}{0.209}}&
\secondres{\scalebox{0.85}{0.165}}&\boldres{\scalebox{0.85}{0.209}}&
{\scalebox{0.85}{0.179}}&{\scalebox{0.85}{0.225}}&
{\scalebox{0.78}{0.167}}&{\scalebox{0.85}{0.230}} \\

& \scalebox{0.85}{192} &
{\scalebox{0.85}{0.222}}&{\scalebox{0.85}{0.261}}&
{\scalebox{0.85}{0.223}}&{\scalebox{0.85}{0.255}}&
{\scalebox{0.85}{0.220}}&{\scalebox{0.85}{0.256}}&
{\scalebox{0.85}{0.238}}&{\scalebox{0.85}{0.299}}&
{\scalebox{0.85}{0.281}}&{\scalebox{0.85}{0.348}}&
\secondres{\scalebox{0.78}{0.213}}&\boldres{\scalebox{0.85}{0.250}}&
\boldres{\scalebox{0.85}{0.210}}&\secondres{\scalebox{0.78}{0.251}}&
{\scalebox{0.85}{0.230}}&{\scalebox{0.85}{0.268}}&
{\scalebox{0.85}{0.214}}&{\scalebox{0.85}{0.275}} \\

& \scalebox{0.85}{336} &
{\scalebox{0.85}{0.272}}&\boldres{{\scalebox{0.85}{0.290}}}&
{\scalebox{0.85}{0.280}}&{\scalebox{0.85}{0.298}}&
{\scalebox{0.85}{0.276}}&{\scalebox{0.85}{0.296}}&
{\scalebox{0.85}{0.281}}&{\scalebox{0.85}{0.330}}&
{\scalebox{0.85}{0.339}}&{\scalebox{0.85}{0.381}}&
{\scalebox{0.85}{0.273}}&\scalebox{0.78}{0.293}&
\boldres{\scalebox{0.85}{0.266}}&\secondres{\scalebox{0.85}{0.291}}&
{\scalebox{0.85}{0.284}}&{\scalebox{0.85}{0.304}}&
\secondres{\scalebox{0.78}{0.271}}&{\scalebox{0.85}{0.317}} \\

& \scalebox{0.85}{720} &
{\scalebox{0.85}{0.359}}&\scalebox{0.78}{0.352}&
\secondres{\scalebox{0.78}{0.353}}&\secondres{\scalebox{0.78}{0.346}}&
\secondres{\scalebox{0.78}{0.353}}&\secondres{\scalebox{0.78}{0.346}}&
\boldres{\scalebox{0.85}{0.345}}&{\scalebox{0.85}{0.381}}&
{\scalebox{0.85}{0.408}}&{\scalebox{0.85}{0.417}}&
{\scalebox{0.85}{0.354}}&{\scalebox{0.85}{0.347}}&
\boldres{\scalebox{0.85}{0.345}}&\boldres{\scalebox{0.85}{0.344}}&
{\scalebox{0.85}{0.356}}&{\scalebox{0.85}{0.350}}&
{\scalebox{0.85}{0.360}}&{\scalebox{0.85}{0.378}} \\

\cmidrule(lr){2-20}

& \scalebox{0.85}{Avg} &
{\scalebox{0.85}{0.254}}&{\scalebox{0.85}{0.277}}&
{\scalebox{0.85}{0.259}}&{\scalebox{0.85}{0.279}}&
{\scalebox{0.85}{0.255}}&{\scalebox{0.85}{0.278}}&
{\scalebox{0.85}{0.265}}&{\scalebox{0.85}{0.316}}&
{\scalebox{0.85}{0.313}}&{\scalebox{0.85}{0.362}}&
\secondres{\scalebox{0.78}{0.252}}&\secondres{\scalebox{0.78}{0.274}}&
\boldres{\scalebox{0.8}{0.246}}&\boldres{\scalebox{0.8}{0.273}}&
{\scalebox{0.85}{0.262}}&{\scalebox{0.85}{0.286}}&
{\scalebox{0.85}{0.253}}&{\scalebox{0.85}{0.300}} \\

\midrule

    \multirow{5}{*}{\rotatebox{90}{\scalebox{0.95}{Traffic}}} 
    & \scalebox{0.85}{96} &
    \secondres{\scalebox{0.78}{0.439}}&\secondres{\scalebox{0.78}{0.300}}&
    {\scalebox{0.85}{0.442}}&{\scalebox{0.85}{0.302}}&
    {\scalebox{0.85}{0.494}}&{\scalebox{0.85}{0.313}}&
    {\scalebox{0.85}{0.696}}&{\scalebox{0.85}{0.428}}&
    {\scalebox{0.85}{0.580}}&{\scalebox{0.85}{0.362}}&
    \boldres{\scalebox{0.85}{0.433}}&\boldres{\scalebox{0.85}{0.290}}&
    {\scalebox{0.85}{0.516}}&{\scalebox{0.85}{0.336}}&
    {\scalebox{0.85}{0.688}}&{\scalebox{0.85}{0.429}}&
    {\scalebox{0.85}{0.781}}&{\scalebox{0.85}{0.401}} \\
    
    & \scalebox{0.85}{192} &
    \boldres{\scalebox{0.78}{0.450}}&\secondres{\scalebox{0.78}{0.302}}&
    \secondres{{\scalebox{0.85}{0.459}}}&{\scalebox{0.85}{0.308}}&
    {\scalebox{0.85}{0.490}}&{\scalebox{0.85}{0.330}}&
    {\scalebox{0.85}{0.646}}&{\scalebox{0.85}{0.407}}&
    {\scalebox{0.85}{0.606}}&{\scalebox{0.85}{0.379}}&
    \boldres{\scalebox{0.85}{0.450}}&\boldres{\scalebox{0.85}{0.296}}&
    {\scalebox{0.85}{0.512}}&{\scalebox{0.85}{0.331}}&
    {\scalebox{0.85}{0.687}}&{\scalebox{0.85}{0.434}}&
    {\scalebox{0.85}{0.789}}&{\scalebox{0.85}{0.405}} \\
    
    & \scalebox{0.85}{336} &
    \boldres{\scalebox{0.78}{0.462}}&\secondres{\scalebox{0.78}{0.310}}&
    {\scalebox{0.85}{0.479}}&{\scalebox{0.85}{0.319}}&
    {\scalebox{0.85}{0.502}}&{\scalebox{0.85}{0.317}}&
    {\scalebox{0.85}{0.653}}&{\scalebox{0.85}{0.409}}&
    {\scalebox{0.85}{0.613}}&{\scalebox{0.85}{0.380}}&
    \secondres{\scalebox{0.85}{0.467}}&\boldres{\scalebox{0.85}{0.306}}&
    {\scalebox{0.85}{0.353}}&{\scalebox{0.85}{0.334}}&
    {\scalebox{0.85}{0.719}}&{\scalebox{0.85}{0.451}}&
    {\scalebox{0.85}{0.797}}&{\scalebox{0.85}{0.405}} \\
    
    & \scalebox{0.85}{720} &
    {\scalebox{0.85}{0.517}}&{\scalebox{0.85}{0.339}}&
    \secondres{\scalebox{0.85}{0.516}}&{\scalebox{0.85}{0.342}}&
    {\scalebox{0.78}{0.545}}&\secondres{\scalebox{0.78}{0.337}}&
    {\scalebox{0.85}{0.694}}&{\scalebox{0.85}{0.428}}&
    {\scalebox{0.85}{0.641}}&{\scalebox{0.85}{0.393}}&
    \boldres{\scalebox{0.85}{0.500}}&\boldres{\scalebox{0.85}{0.326}}&
    {\scalebox{0.85}{0.558}}&{\scalebox{0.85}{0.353}}&
    {\scalebox{0.85}{0.761}}&{\scalebox{0.85}{0.466}}&
    {\scalebox{0.85}{0.813}}&{\scalebox{0.85}{0.421}} \\
    
    \cmidrule(lr){2-20}
    
    & \scalebox{0.85}{Avg} &
    \secondres{\scalebox{0.78}{0.467}}&\secondres{\scalebox{0.78}{0.312}}&
    {\scalebox{0.85}{0.474}}&{\scalebox{0.85}{0.317}}&
    {\scalebox{0.85}{0.507}}&{\scalebox{0.85}{0.324}}&
    {\scalebox{0.85}{0.672}}&{\scalebox{0.85}{0.418}}&
    {\scalebox{0.85}{0.610}}&{\scalebox{0.85}{0.378}}&
    \boldres{\scalebox{0.8}{0.462}}&\boldres{\scalebox{0.8}{0.304}}&
    {\scalebox{0.85}{0.484}}&{\scalebox{0.85}{0.338}}&
    {\scalebox{0.85}{0.713}}&{\scalebox{0.85}{0.445}}&
    {\scalebox{0.85}{0.795}}&{\scalebox{0.85}{0.408}} \\
    \midrule

    \multirow{5}{*}{\rotatebox{90}{\scalebox{0.95}{ETTh2}}} 
    & \scalebox{0.85}{96} &
    {\scalebox{0.85}{0.300}}&{\scalebox{0.85}{0.348}}&
    \secondres{\scalebox{0.78}{0.295}}&\secondres{\scalebox{0.78}{0.344}}&
    \boldres{\scalebox{0.85}{0.293}}&\boldres{\scalebox{0.85}{0.343}}&
    {\scalebox{0.85}{0.341}}&{\scalebox{0.85}{0.395}}&
    {\scalebox{0.85}{0.350}}&{\scalebox{0.85}{0.391}}&
    {\scalebox{0.85}{0.299}}&{\scalebox{0.85}{0.350}}&
    \boldres{\scalebox{0.85}{0.293}}&{\scalebox{0.85}{0.345}}&
    {\scalebox{0.85}{0.296}}&{\scalebox{0.85}{0.345}}&
    \secondres{\scalebox{0.78}{0.295}}&{\scalebox{0.85}{0.355}} \\
    
    & \scalebox{0.85}{192} &
    {\scalebox{0.85}{0.389}}&{\scalebox{0.85}{0.402}}&
    \secondres{\scalebox{0.78}{0.375}}&{\scalebox{0.85}{0.398}}&
    \boldres{\scalebox{0.85}{0.373}}&\boldres{\scalebox{0.85}{0.393}}&
    {\scalebox{0.85}{0.481}}&{\scalebox{0.85}{0.479}}&
    {\scalebox{0.85}{0.441}}&{\scalebox{0.85}{0.449}}&
    {\scalebox{0.85}{0.384}}&{\scalebox{0.85}{0.401}}&
    {\scalebox{0.85}{0.382}}&{\scalebox{0.85}{0.401}}&
    {\scalebox{0.85}{0.384}}&{\scalebox{0.85}{0.401}}&
    \secondres{\scalebox{0.78}{0.375}}&\secondres{\scalebox{0.78}{0.397}} \\
    
    & \scalebox{0.85}{336} &
    \boldres{\scalebox{0.85}{0.409}}&\boldres{\scalebox{0.85}{0.408}}&
    {\scalebox{0.85}{0.436}}&{\scalebox{0.85}{0.435}}&
    {\scalebox{0.85}{0.422}}&{\scalebox{0.85}{0.430}}&
    {\scalebox{0.85}{0.592}}&{\scalebox{0.85}{0.542}}&
    {\scalebox{0.85}{0.498}}&{\scalebox{0.85}{0.490}}&
    \secondres{\scalebox{0.78}{0.411}}&\secondres{\scalebox{0.78}{0.428}}&
    {\scalebox{0.85}{0.424}}&{\scalebox{0.85}{0.437}}&
    {\scalebox{0.85}{0.429}}&{\scalebox{0.85}{0.439}}&
    {\scalebox{0.85}{0.414}}&{\scalebox{0.85}{0.436}} \\
    
    & \scalebox{0.85}{720} &
    \boldres{\scalebox{0.85}{0.421}}&\boldres{\scalebox{0.85}{0.439}}&
    \secondres{\scalebox{0.78}{0.429}}&{\scalebox{0.85}{0.447}}&
    {\scalebox{0.85}{0.440}}&{\scalebox{0.85}{0.452}}&
    {\scalebox{0.85}{0.840}}&{\scalebox{0.85}{0.661}}&
    {\scalebox{0.85}{0.480}}&{\scalebox{0.85}{0.487}}&
    {\scalebox{0.85}{0.430}}&\secondres{\scalebox{0.78}{0.446}}&
    {\scalebox{0.85}{0.449}}&{\scalebox{0.85}{0.458}}&
    {\scalebox{0.85}{0.433}}&{\scalebox{0.85}{0.451}}&
    {\scalebox{0.85}{0.443}}&{\scalebox{0.85}{0.469}} \\
    
    \cmidrule(lr){2-20}
    
    & \scalebox{0.85}{Avg} &
    \boldres{\scalebox{0.8}{0.379}}&\boldres{\scalebox{0.8}{0.399}}&
    {\scalebox{0.85}{0.383}}&{\scalebox{0.85}{0.406}}&
    {\scalebox{0.85}{0.382}}&\secondres{\scalebox{0.78}{0.404}}&
    {\scalebox{0.85}{0.563}}&{\scalebox{0.85}{0.519}}&
    {\scalebox{0.85}{0.442}}&{\scalebox{0.85}{0.454}}&
    \secondres{\scalebox{0.78}{0.381}}&{\scalebox{0.85}{0.406}}&
    {\scalebox{0.85}{0.387}}&{\scalebox{0.85}{0.410}}&
    {\scalebox{0.85}{0.385}}&{\scalebox{0.85}{0.409}}&
    \secondres{\scalebox{0.78}{0.381}}&{\scalebox{0.85}{0.414}} \\
    \midrule

    \multirow{5}{*}{\rotatebox{90}{\scalebox{0.95}{Solar-Energy}}} 
    & \scalebox{0.85}{96} &
    \boldres{\scalebox{0.85}{0.199}}&\boldres{\scalebox{0.85}{0.239}}&
    {\scalebox{0.85}{0.213}}&{\scalebox{0.85}{0.253}}&
    {\scalebox{0.85}{0.214}}&{\scalebox{0.85}{0.257}}&
    {\scalebox{0.85}{0.289}}&{\scalebox{0.85}{0.377}}&
    {\scalebox{0.85}{0.279}}&{\scalebox{0.85}{0.363}}&
    \secondres{\scalebox{0.78}{0.203}}&\secondres{\scalebox{0.78}{0.240}}&
    {\scalebox{0.85}{0.232}}&{\scalebox{0.85}{0.278}}&
    {\scalebox{0.85}{0.256}}&{\scalebox{0.85}{0.292}}&
    {\scalebox{0.85}{0.232}}&{\scalebox{0.85}{0.294}} \\
    
    & \scalebox{0.85}{192} &
    \boldres{\scalebox{0.85}{0.233}}&\boldres{\scalebox{0.85}{0.263}}&
    {\scalebox{0.85}{0.242}}&{\scalebox{0.85}{0.274}}&
    {\scalebox{0.85}{0.254}}&{\scalebox{0.85}{0.296}}&
    {\scalebox{0.85}{0.319}}&{\scalebox{0.85}{0.397}}&
    {\scalebox{0.85}{0.288}}&{\scalebox{0.85}{0.378}}&
    \secondres{\scalebox{0.78}{0.237}}&\secondres{\scalebox{0.78}{0.268}}&
    {\scalebox{0.85}{0.268}}&{\scalebox{0.85}{0.303}}&
    {\scalebox{0.85}{0.286}}&{\scalebox{0.85}{0.315}}&
    {\scalebox{0.85}{0.253}}&{\scalebox{0.85}{0.306}} \\
    
    & \scalebox{0.85}{336} &
    \secondres{\scalebox{0.78}{0.252}}&\boldres{\scalebox{0.85}{0.279}}&
    {\scalebox{0.85}{0.262}}&{\scalebox{0.85}{0.290}}&
    {\scalebox{0.85}{0.284}}&{\scalebox{0.85}{0.314}}&
    {\scalebox{0.85}{0.352}}&{\scalebox{0.85}{0.415}}&
    {\scalebox{0.85}{0.316}}&{\scalebox{0.85}{0.399}}&
    \boldres{\scalebox{0.85}{0.249}}&\secondres{\scalebox{0.78}{0.280}}&
    {\scalebox{0.85}{0.291}}&{\scalebox{0.85}{0.316}}&
    {\scalebox{0.85}{0.323}}&{\scalebox{0.85}{0.337}}&
    {\scalebox{0.85}{0.264}}&{\scalebox{0.85}{0.310}} \\
    
    & \scalebox{0.85}{720} &
    \secondres{\scalebox{0.78}{0.259}}&\boldres{\scalebox{0.85}{0.280}}&
    {\scalebox{0.85}{0.270}}&{\scalebox{0.85}{0.296}}&
    {\scalebox{0.85}{0.263}}&{\scalebox{0.85}{0.291}}&
    {\scalebox{0.85}{0.356}}&{\scalebox{0.85}{0.412}}&
    {\scalebox{0.85}{0.363}}&{\scalebox{0.85}{0.430}}&
    \boldres{\scalebox{0.85}{0.254}}&\secondres{\scalebox{0.78}{0.283}}&
    {\scalebox{0.85}{0.290}}&{\scalebox{0.85}{0.315}}&
    {\scalebox{0.85}{0.346}}&{\scalebox{0.85}{0.349}}&
    {\scalebox{0.85}{0.260}}&{\scalebox{0.85}{0.306}} \\
    
    \cmidrule(lr){2-20}
    
    & \scalebox{0.85}{Avg} &
    \boldres{\scalebox{0.8}{0.235}}&\boldres{\scalebox{0.8}{0.265}}&
    \secondres{\scalebox{0.78}{0.246}}&{\scalebox{0.85}{0.278}}&
    {\scalebox{0.85}{0.253}}&{\scalebox{0.85}{0.289}}&
    {\scalebox{0.85}{0.329}}&{\scalebox{0.85}{0.400}}&
    {\scalebox{0.85}{0.311}}&{\scalebox{0.85}{0.392}}&
    \boldres{\scalebox{0.8}{0.235}}&\secondres{\scalebox{0.78}{0.267}}&
    {\scalebox{0.85}{0.270}}&{\scalebox{0.85}{0.303}}&
    {\scalebox{0.85}{0.302}}&{\scalebox{0.85}{0.323}}&
    {\scalebox{0.85}{0.252}}&{\scalebox{0.85}{0.304}} \\
    \midrule

    \multirow{5}{*}{\rotatebox{90}{\scalebox{0.95}{PEMS03}}} 
    & \scalebox{0.85}{96} &
    \boldres{\scalebox{0.85}{0.25}}&\boldres{\scalebox{0.85}{0.343}}&
    {\scalebox{0.85}{0.337}}&{\scalebox{0.85}{0.413}}&
    {\scalebox{0.85}{0.504}}&{\scalebox{0.85}{0.516}}&
    {\scalebox{0.85}{0.458}}&{\scalebox{0.85}{0.517}}&
    {\scalebox{0.85}{0.325}}&{\scalebox{0.85}{0.423}}&
    \secondres{\scalebox{0.78}{0.302}}&\secondres{\scalebox{0.78}{0.395}}&
    {\scalebox{0.85}{0.514}}&{\scalebox{0.85}{0.505}}&
    {\scalebox{0.85}{0.731}}&{\scalebox{0.85}{0.625}}&
    {\scalebox{0.85}{0.381}}&{\scalebox{0.85}{0.438}} \\
    
    & \scalebox{0.85}{192} &
    \boldres{\scalebox{0.85}{0.297}}&\boldres{\scalebox{0.85}{0.378}}&
    {\scalebox{0.85}{0.374}}&{\scalebox{0.85}{0.436}}&
    {\scalebox{0.85}{0.547}}&{\scalebox{0.85}{0.544}}&
    {\scalebox{0.85}{0.477}}&{\scalebox{0.85}{0.527}}&
    {\scalebox{0.85}{0.475}}&{\scalebox{0.85}{0.515}}&
    \secondres{\scalebox{0.78}{0.349}}&\secondres{\scalebox{0.78}{0.428}}&
    {\scalebox{0.85}{0.557}}&{\scalebox{0.85}{0.526}}&
    {\scalebox{0.85}{0.797}}&{\scalebox{0.85}{0.648}}&
    {\scalebox{0.85}{0.447}}&{\scalebox{0.85}{0.478}} \\
    
    & \scalebox{0.85}{336} &
    \boldres{\scalebox{0.85}{0.274}}&\boldres{\scalebox{0.85}{0.356}}&
    {\scalebox{0.85}{0.332}}&{\scalebox{0.85}{0.395}}&
    {\scalebox{0.85}{0.432}}&{\scalebox{0.85}{0.460}}&
    {\scalebox{0.85}{0.396}}&{\scalebox{0.85}{0.456}}&
    {\scalebox{0.85}{0.425}}&{\scalebox{0.85}{0.480}}&
    \secondres{\scalebox{0.78}{0.313}}&\secondres{\scalebox{0.78}{0.388}}&
    {\scalebox{0.85}{0.440}}&{\scalebox{0.85}{0.450}}&
    {\scalebox{0.85}{0.614}}&{\scalebox{0.85}{0.545}}&
    {\scalebox{0.85}{0.380}}&{\scalebox{0.85}{0.432}} \\
    
    & \scalebox{0.85}{720} &
    \boldres{\scalebox{0.85}{0.336}}&\boldres{\scalebox{0.85}{0.401}}&
    {\scalebox{0.85}{0.397}}&\secondres{\scalebox{0.78}{0.440}}&
    {\scalebox{0.85}{0.527}}&{\scalebox{0.85}{0.521}}&
    {\scalebox{0.85}{0.439}}&{\scalebox{0.85}{0.492}}&
    {\scalebox{0.85}{0.644}}&{\scalebox{0.85}{0.596}}&
    \secondres{\scalebox{0.78}{0.383}}&{\scalebox{0.85}{0.441}}&
    {\scalebox{0.85}{0.533}}&{\scalebox{0.85}{0.509}}&
    {\scalebox{0.85}{0.762}}&{\scalebox{0.85}{0.638}}&
    {\scalebox{0.85}{0.447}}&{\scalebox{0.85}{0.478}} \\
    
    \cmidrule(lr){2-20}
    
    & \scalebox{0.85}{Avg} &
    \boldres{\scalebox{0.8}{0.289}}&\boldres{\scalebox{0.8}{0.369}}&
    {\scalebox{0.85}{0.360}}&{\scalebox{0.85}{0.421}}&
    {\scalebox{0.85}{0.502}}&{\scalebox{0.85}{0.510}}&
    {\scalebox{0.85}{0.442}}&{\scalebox{0.85}{0.498}}&
    {\scalebox{0.85}{0.467}}&{\scalebox{0.85}{0.503}}&
    \secondres{\scalebox{0.78}{0.336}}&\secondres{\scalebox{0.78}{0.413}}&
    {\scalebox{0.85}{0.511}}&{\scalebox{0.85}{0.497}}&
    {\scalebox{0.85}{0.726}}&{\scalebox{0.85}{0.614}}&
    {\scalebox{0.85}{0.413}}&{\scalebox{0.85}{0.456}} \\
    \midrule

    \multirow{5}{*}{\rotatebox{90}{\scalebox{0.95}{PEMS04}}} 
    & \scalebox{0.85}{96} &
    \boldres{\scalebox{0.85}{0.226}}&\boldres{\scalebox{0.85}{0.335}}&
    {\scalebox{0.85}{0.387}}&{\scalebox{0.85}{0.446}}&
    {\scalebox{0.85}{0.645}}&{\scalebox{0.85}{0.594}}&
    {\scalebox{0.85}{0.452}}&{\scalebox{0.85}{0.504}}&
    \secondres{\scalebox{0.78}{0.300}}&\secondres{\scalebox{0.78}{0.411}}&
    {\scalebox{0.85}{0.356}}&{\scalebox{0.85}{0.428}}&
    {\scalebox{0.85}{0.584}}&{\scalebox{0.85}{0.554}}&
    {\scalebox{0.85}{0.794}}&{\scalebox{0.85}{0.658}}&
    {\scalebox{0.85}{0.415}}&{\scalebox{0.85}{0.468}} \\
    
    & \scalebox{0.85}{192} &
    \boldres{\scalebox{0.85}{0.286}}&\boldres{\scalebox{0.85}{0.383}}&
    {\scalebox{0.85}{0.429}}&{\scalebox{0.85}{0.473}}&
    {\scalebox{0.85}{0.688}}&{\scalebox{0.85}{0.615}}&
    {\scalebox{0.85}{0.477}}&{\scalebox{0.85}{0.527}}&
    {\scalebox{0.85}{0.458}}&{\scalebox{0.85}{0.513}}&
    \secondres{\scalebox{0.78}{0.398}}&\secondres{\scalebox{0.78}{0.459}}&
    {\scalebox{0.85}{0.650}}&{\scalebox{0.85}{0.586}}&
    {\scalebox{0.85}{0.891}}&{\scalebox{0.85}{0.697}}&
    {\scalebox{0.85}{0.496}}&{\scalebox{0.85}{0.510}} \\
    
    & \scalebox{0.85}{336} &
    \boldres{\scalebox{0.85}{0.286}}&\boldres{\scalebox{0.85}{0.372}}&
    {\scalebox{0.85}{0.369}}&{\scalebox{0.85}{0.426}}&
    {\scalebox{0.85}{0.536}}&{\scalebox{0.85}{0.516}}&
    {\scalebox{0.85}{0.396}}&{\scalebox{0.85}{0.456}}&
    {\scalebox{0.85}{0.411}}&{\scalebox{0.85}{0.474}}&
    \secondres{\scalebox{0.78}{0.348}}&\secondres{\scalebox{0.78}{0.413}}&
    {\scalebox{0.85}{0.511}}&{\scalebox{0.85}{0.499}}&
    {\scalebox{0.85}{0.703}}&{\scalebox{0.85}{0.596}}&
    {\scalebox{0.85}{0.420}}&{\scalebox{0.85}{0.461}} \\
    
    & \scalebox{0.85}{720} &
    \boldres{\scalebox{0.85}{0.338}}&\boldres{\scalebox{0.85}{0.414}}&
    {\scalebox{0.85}{0.440}}&{\scalebox{0.85}{0.476}}&
    {\scalebox{0.85}{0.629}}&{\scalebox{0.85}{0.578}}&
    {\scalebox{0.85}{0.439}}&{\scalebox{0.85}{0.492}}&
    {\scalebox{0.85}{0.718}}&{\scalebox{0.85}{0.630}}&
    \secondres{\scalebox{0.78}{0.423}}&\secondres{\scalebox{0.78}{0.467}}&
    {\scalebox{0.85}{0.601}}&{\scalebox{0.85}{0.556}}&
    {\scalebox{0.85}{0.812}}&{\scalebox{0.85}{0.660}}&
    {\scalebox{0.85}{0.479}}&{\scalebox{0.85}{0.502}} \\
    
    \cmidrule(lr){2-20}
    
    & \scalebox{0.85}{Avg} &
    \boldres{\scalebox{0.8}{0.284}}&\boldres{\scalebox{0.8}{0.376}}&
    {\scalebox{0.85}{0.406}}&{\scalebox{0.85}{0.455}}&
    {\scalebox{0.85}{0.624}}&{\scalebox{0.85}{0.575}}&
    {\scalebox{0.85}{0.441}}&{\scalebox{0.85}{0.494}}&
    {\scalebox{0.85}{0.471}}&{\scalebox{0.85}{0.507}}&
    \secondres{\scalebox{0.78}{0.381}}&\secondres{\scalebox{0.78}{0.441}}&
    {\scalebox{0.85}{0.586}}&{\scalebox{0.85}{0.548}}&
    {\scalebox{0.85}{0.800}}&{\scalebox{0.85}{0.652}}&
    {\scalebox{0.85}{0.452}}&{\scalebox{0.85}{0.485}} \\
    \midrule

    \multirow{5}{*}{\rotatebox{90}{\scalebox{0.95}{PEMS08}}} 
    & \scalebox{0.85}{96} &
    \boldres{\scalebox{0.85}{0.365}}&\boldres{\scalebox{0.85}{0.409}}&
    {\scalebox{0.85}{0.477}}&{\scalebox{0.85}{0.470}}&
    {\scalebox{0.85}{0.567}}&{\scalebox{0.85}{0.534}}&
    {\scalebox{0.85}{0.672}}&{\scalebox{0.85}{0.564}}&
    {\scalebox{0.85}{0.504}}&{\scalebox{0.85}{0.506}}&
    \secondres{\scalebox{0.78}{0.406}}&\secondres{\scalebox{0.78}{0.437}}&
    {\scalebox{0.85}{0.648}}&{\scalebox{0.85}{0.556}}&
    {\scalebox{0.85}{0.854}}&{\scalebox{0.85}{0.657}}&
    {\scalebox{0.85}{0.516}}&{\scalebox{0.85}{0.487}} \\
    
    & \scalebox{0.85}{192} &
    \boldres{\scalebox{0.85}{0.513}}&\boldres{\scalebox{0.85}{0.462}}&
    {\scalebox{0.85}{0.627}}&{\scalebox{0.85}{0.518}}&
    {\scalebox{0.85}{0.794}}&{\scalebox{0.85}{0.620}}&
    {\scalebox{0.85}{0.726}}&{\scalebox{0.85}{0.580}}&
    {\scalebox{0.85}{0.726}}&{\scalebox{0.85}{0.608}}&
    \secondres{\scalebox{0.78}{0.539}}&\secondres{\scalebox{0.78}{0.479}}&
    {\scalebox{0.85}{0.766}}&{\scalebox{0.85}{0.581}}&
    {\scalebox{0.85}{1.048}}&{\scalebox{0.85}{0.702}}&
    {\scalebox{0.85}{0.646}}&{\scalebox{0.85}{0.528}} \\
    
    & \scalebox{0.85}{336} &
    \boldres{\scalebox{0.85}{0.530}}&\boldres{\scalebox{0.85}{0.435}}&
    {\scalebox{0.85}{0.599}}&{\scalebox{0.85}{0.472}}&
    {\scalebox{0.85}{0.709}}&{\scalebox{0.85}{0.539}}&
    {\scalebox{0.85}{0.650}}&{\scalebox{0.85}{0.518}}&
    {\scalebox{0.85}{0.803}}&{\scalebox{0.85}{0.615}}&
    \secondres{\scalebox{0.78}{0.538}}&\secondres{\scalebox{0.78}{0.444}}&
    {\scalebox{0.85}{0.696}}&{\scalebox{0.85}{0.518}}&
    {\scalebox{0.85}{0.906}}&{\scalebox{0.85}{0.615}}&
    {\scalebox{0.85}{0.617}}&{\scalebox{0.85}{0.490}} \\
    
    & \scalebox{0.85}{720} &
    \secondres{\scalebox{0.78}{0.624}}&\boldres{\scalebox{0.85}{0.487}}&
    {\scalebox{0.85}{0.691}}&{\scalebox{0.85}{0.532}}&
    {\scalebox{0.85}{0.807}}&{\scalebox{0.85}{0.599}}&
    {\scalebox{0.85}{0.713}}&{\scalebox{0.85}{0.562}}&
    {\scalebox{0.85}{1.000}}&{\scalebox{0.85}{0.711}}&
    \boldres{\scalebox{0.85}{0.603}}&\secondres{\scalebox{0.78}{0.489}}&
    {\scalebox{0.85}{0.789}}&{\scalebox{0.85}{0.575}}&
    {\scalebox{0.85}{1.019}}&{\scalebox{0.85}{0.682}}&
    {\scalebox{0.85}{0.707}}&{\scalebox{0.85}{0.541}} \\
    
    \cmidrule(lr){2-20}
    
    & \scalebox{0.85}{Avg} &
    \boldres{\scalebox{0.8}{0.508}}&\boldres{\scalebox{0.8}{0.448}}&
    {\scalebox{0.85}{0.598}}&{\scalebox{0.85}{0.498}}&
    {\scalebox{0.85}{0.719}}&{\scalebox{0.85}{0.573}}&
    {\scalebox{0.85}{0.690}}&{\scalebox{0.85}{0.556}}&
    {\scalebox{0.85}{0.758}}&{\scalebox{0.85}{0.610}}&
    \secondres{\scalebox{0.78}{0.521}}&\secondres{\scalebox{0.78}{0.462}}&
    {\scalebox{0.85}{0.724}}&{\scalebox{0.85}{0.557}}&
    {\scalebox{0.85}{0.956}}&{\scalebox{0.85}{0.664}}&
    {\scalebox{0.85}{0.621}}&{\scalebox{0.85}{0.511}} \\
    \midrule

    \multirow{1}{*}{\rotatebox{90}{\scalebox{0.95}{-}}} 
    & \scalebox{0.85}{Count} &
    \boldres{\scalebox{0.85}{26}}&\boldres{\scalebox{0.85}{29}}&
    {\scalebox{0.85}{0}}&{\scalebox{0.85}{0}}&
    {\scalebox{0.85}{2}}&{\scalebox{0.85}{2}}&
    {\scalebox{0.85}{0}}&{\scalebox{0.85}{0}}&
    {\scalebox{0.78}{0}}&{\scalebox{0.78}{0}}&
    \secondres{\scalebox{0.85}{10}}&\secondres{\scalebox{0.85}{9}}&
    {\scalebox{0.85}{6}}&{\scalebox{0.85}{3}}&
    {\scalebox{0.85}{0}}&{\scalebox{0.85}{0}}&
    {\scalebox{0.85}{0}}&{\scalebox{0.85}{0}} \\
  \bottomrule
  \end{tabular}
    \end{small}
  \end{threeparttable}
  }
\end{table}

\begin{table*}[htbp]
\centering
\caption{Robustness of HTMformer performance is evaluated over five random seeds.}
\vspace{5.5pt}
\label{tab:seed1}
\setlength{\tabcolsep}{8pt}
\renewcommand{\arraystretch}{1.1}

\begin{adjustbox}{width=\linewidth}
\begin{tabular}{l|cc|cc|cc|cc}
\toprule
Dataset & \multicolumn{2}{c|}{ECL} & \multicolumn{2}{c|}{Weather} & \multicolumn{2}{c}{Traffic} & \multicolumn{2}{c}{ETTh2} \\
\cmidrule(lr){2-3}\cmidrule(lr){4-5}\cmidrule(lr){6-7}\cmidrule(lr){8-9}
Horizon & MSE & MAE & MSE & MAE & MSE & MAE & MSE & MAE \\
\midrule
96  & $0.157\pm0.003$ & $0.247\pm0.002$ & $0.164\pm0.004$ & $0.209\pm0.002$ & $0.439\pm0.002$ & $0.300\pm0.002$ & $0.300\pm0.003$ & $0.348\pm0.004$ \\
192 & $0.171\pm0.003$ & $0.260\pm0.002$ & $0.222\pm0.005$ & $0.261\pm0.004$ & $0.450\pm0.002$ & $0.302\pm0.003$ & $0.389\pm0.004$ & $0.402\pm0.003$ \\
336 & $0.192\pm0.005$ & $0.277\pm0.005$ & $0.272\pm0.005$ & $0.290\pm0.010$ & $0.462\pm0.004$ & $0.310\pm0.005$ & $0.409\pm0.004$ & $0.408\pm0.004$ \\
720 & $0.220\pm0.006$ & $0.304\pm0.007$ & $0.359\pm0.004$ & $0.352\pm0.002$ & $0.517\pm0.009$ & $0.339\pm0.007$ & $0.421\pm0.007$ & $0.439\pm0.007$ \\
\bottomrule
\end{tabular}
\end{adjustbox}

\vspace{0.6em}

\begin{adjustbox}{width=\linewidth}
\begin{tabular}{l|cc|cc|cc|cc} 
\toprule
Dataset & \multicolumn{2}{c|}{Solar-Energy} & \multicolumn{2}{c|}{PEMS03} & \multicolumn{2}{c}{PEMS04} & \multicolumn{2}{c}{PEMS08} \\
\cmidrule(lr){2-3}\cmidrule(lr){4-5}\cmidrule(lr){6-7}\cmidrule(lr){8-9}
Horizon & MSE & MAE & MSE & MAE & MSE & MAE & MSE & MAE \\
\midrule
96  & $0.199\pm0.004$ & $0.239\pm0.003$ & $0.250\pm0.003$ & $0.343\pm0.002$ & $0.226\pm0.008$ & $0.335\pm0.004$ & $0.365\pm0.003$ & $0.409\pm0.005$ \\
192 & $0.233\pm0.004$ & $0.263\pm0.006$ & $0.297\pm0.005$ & $0.378\pm0.010$ & $0.286\pm0.010$ & $0.383\pm0.005$ & $0.513\pm0.005$ & $0.462\pm0.006$ \\
336 & $0.252\pm0.006$ & $0.279\pm0.007$ & $0.274\pm0.010$ & $0.356\pm0.009$ & $0.286\pm0.010$ & $0.372\pm0.005$ & $0.530\pm0.006$ & $0.435\pm0.010$ \\
720 & $0.259\pm0.006$ & $0.280\pm0.009$ & $0.336\pm0.007$ & $0.401\pm0.008$ & $0.338\pm0.009$ & $0.414\pm0.007$ & $0.624\pm0.003$ & $0.487\pm0.006$ \\
\bottomrule
\end{tabular}
\end{adjustbox}
\end{table*}

\subsection{Short-term Forecasting Results}\label{app:F 3}
Table~\ref{tab:shortterm2} presents the short-term forecasting results. 
The evaluation metrics comprise the Mean Squared Error (MSE) and Mean Absolute Error (MAE) computed over four prediction horizons. 
Our approach consistently achieves substantial improvements across all horizons on the selected datasets.
In particular, on the PEMS04 dataset, our proposed model achieved a \emph{33.3\%} reduction in average MSE and an \emph{18.3\%} reduction in average MAE compared to the second-best results. The standard deviations, computed over five independent runs with different random seeds, are reported in Table~\ref{tab:seed2}. 
For short-term forecasting, the limited length of the input sequence constrains the model's capacity to capture sufficient temporal dependencies. 
Consequently, modeling multivariate correlations becomes crucial, in which HTMformer demonstrates superior performance.

\begin{table}[htbp]
  \caption{Full results for the short-term forecasting task. We compare extensive competitive models on PEMS datasets. \emph{Avg} means the average results from all four prediction lengths.}
  \vspace{5.5pt}
  \centering
  \label{tab:shortterm2}
  \resizebox{1.0\columnwidth}{!}{
  \begin{threeparttable}
  \begin{small}
  \renewcommand{\multirowsetup}{\centering}
  \setlength{\tabcolsep}{1pt}
\begin{tabular}{c|c|cc|cc|cc|cc|cc|cc|cc|cc}
    \toprule
    \multicolumn{2}{c}{\multirow{2}{*}{Models}} & 
    \multicolumn{2}{c}{\rotatebox{0}{\scalebox{0.8}{\textbf{HTMformer}}}} &
    \multicolumn{2}{c}{\rotatebox{0}{\scalebox{0.8}{iTransformer}}} &
    \multicolumn{2}{c}{\rotatebox{0}{\scalebox{0.8}{PatchTST}}} &
    \multicolumn{2}{c}{\rotatebox{0}{\scalebox{0.8}{DLinear}}} &
    \multicolumn{2}{c}{\rotatebox{0}{\scalebox{0.8}{FEDformer}}} &
    \multicolumn{2}{c}{\rotatebox{0}{\scalebox{0.8}{MPFormer}}} &
    \multicolumn{2}{c}{\rotatebox{0}{\scalebox{0.8}{WPMixer}}} & 
    \multicolumn{2}{c}{\rotatebox{0}{\scalebox{0.8}{TimeMixer}}}  \\

    \multicolumn{2}{c}{} & 
    \multicolumn{2}{c}{\scalebox{0.8}{(\textbf{Ours})}} &
    \multicolumn{2}{c}{\scalebox{0.8}{(\citeyear{liu2024itransformer})}} &
    \multicolumn{2}{c}{\scalebox{0.8}{\citeyear{nie2023time}}} &
    \multicolumn{2}{c}{\scalebox{0.8}{\citeyear{zeng2023dlinear}}}&
    \multicolumn{2}{c}{\scalebox{0.8}{\citeyear{zhou2022fedformer}}}&
    \multicolumn{2}{c}{\scalebox{0.8}{\citeyear{naghashi2025multipatchformer}}}&
    \multicolumn{2}{c}{\scalebox{0.8}{\citeyear{murad2024wp}}}&
    \multicolumn{2}{c}{\scalebox{0.8}{\citeyear{wang2023timemixer}}} \\

    \cmidrule(lr){3-4} \cmidrule(lr){5-6} \cmidrule(lr){7-8} \cmidrule(lr){9-10} 
    \cmidrule(lr){11-12} \cmidrule(lr){13-14} \cmidrule(lr){15-16} \cmidrule(lr){17-18}

    \multicolumn{2}{c}{Metric} & 
    \scalebox{0.78}{MSE} & \scalebox{0.78}{MAE} & 
    \scalebox{0.78}{MSE} & \scalebox{0.78}{MAE} & 
    \scalebox{0.78}{MSE} & \scalebox{0.78}{MAE} & 
    \scalebox{0.78}{MSE} & \scalebox{0.78}{MAE} & 
    \scalebox{0.78}{MSE} & \scalebox{0.78}{MAE} & 
    \scalebox{0.78}{MSE} & \scalebox{0.78}{MAE} & 
    \scalebox{0.78}{MSE} & \scalebox{0.78}{MAE} & 
    \scalebox{0.78}{MSE} & \scalebox{0.78}{MAE}  \\ 
    
    \toprule
  
    \multirow{5}{*}{\rotatebox{90}{\scalebox{0.95}{PEMS03}}} 
    & \scalebox{0.85}{12} &
    \boldres{\scalebox{0.85}{0.067}} & \boldres{\scalebox{0.85}{0.174}} &
    \secondres{\scalebox{0.85}{0.075}} & \secondres{\scalebox{0.85}{0.184}} &
    {\scalebox{0.85}{0.102}} & {\scalebox{0.85}{0.217}} &
    {\scalebox{0.85}{0.122}} & {\scalebox{0.85}{0.245}} &
    {\scalebox{0.85}{0.125}} & {\scalebox{0.85}{0.251}} &
    {\scalebox{0.85}{0.089}} & {\scalebox{0.85}{0.199}} &
    {\scalebox{0.85}{0.093}} & {\scalebox{0.85}{0.204}} &
    {\scalebox{0.85}{0.107}} & {\scalebox{0.85}{0.220}} \\
    & \scalebox{0.85}{24} &
    \boldres{\scalebox{0.85}{0.093}} & \boldres{\scalebox{0.85}{0.204}} &
    {\scalebox{0.85}{0.115}} & {\scalebox{0.85}{0.229}} &
    {\scalebox{0.85}{0.187}} & {\scalebox{0.85}{0.298}} &
    {\scalebox{0.85}{0.201}} & {\scalebox{0.85}{0.320}} &
    {\scalebox{0.85}{0.152}} & {\scalebox{0.85}{0.279}} &
    \secondres{\scalebox{0.85}{0.114}} & \secondres{\scalebox{0.85}{0.225}} &
    {\scalebox{0.85}{0.151}} & {\scalebox{0.85}{0.263}} &
    {\scalebox{0.85}{0.185}} & {\scalebox{0.85}{0.292}} \\
    & \scalebox{0.85}{48} &
    \boldres{\scalebox{0.85}{0.151}} & \boldres{\scalebox{0.85}{0.261}} &
    {\scalebox{0.85}{0.195}} & {\scalebox{0.85}{0.305}} &
    {\scalebox{0.85}{0.278}} & {\scalebox{0.85}{0.369}} &
    {\scalebox{0.85}{0.334}} & {\scalebox{0.85}{0.428}} &
    {\scalebox{0.85}{0.222}} & {\scalebox{0.85}{0.347}} &
    \secondres{\scalebox{0.85}{0.186}} & \secondres{\scalebox{0.85}{0.296}} &
    {\scalebox{0.85}{0.292}} & {\scalebox{0.85}{0.373}} &
    {\scalebox{0.85}{0.375}} & {\scalebox{0.85}{0.428}} \\
    & \scalebox{0.85}{96} &
    \boldres{\scalebox{0.85}{0.241}} & \boldres{\scalebox{0.85}{0.341}} &
    {\scalebox{0.85}{0.340}} & {\scalebox{0.85}{0.417}} &
    {\scalebox{0.85}{0.504}} & {\scalebox{0.85}{0.516}} &
    {\scalebox{0.85}{0.458}} & {\scalebox{0.85}{0.517}} &
    {\scalebox{0.85}{0.325}} & {\scalebox{0.85}{0.423}} &
    \secondres{\scalebox{0.85}{0.300}} & \secondres{\scalebox{0.85}{0.393}} &
    {\scalebox{0.85}{0.514}} & {\scalebox{0.85}{0.505}} &
    {\scalebox{0.85}{0.731}} & {\scalebox{0.85}{0.625}} \\
    
    \cmidrule(lr){2-18}
    
    & \scalebox{0.85}{Avg} &
    \boldres{\scalebox{0.8}{0.138}} & \boldres{\scalebox{0.8}{0.245}} &
    {\scalebox{0.85}{0.181}} & {\scalebox{0.85}{0.283}} &
    {\scalebox{0.85}{0.267}} & {\scalebox{0.85}{0.350}} &
    {\scalebox{0.85}{0.278}} & {\scalebox{0.85}{0.377}} &
    {\scalebox{0.85}{0.206}} & {\scalebox{0.85}{0.325}} &
    \secondres{\scalebox{0.78}{0.172}} & \secondres{\scalebox{0.78}{0.278}} &
    {\scalebox{0.85}{0.262}} & {\scalebox{0.85}{0.336}} &
    {\scalebox{0.85}{0.348}} & {\scalebox{0.85}{0.391}} \\
    \midrule

    \multirow{5}{*}{\rotatebox{90}{\scalebox{0.95}{PEMS04}}} 
    & \scalebox{0.85}{12} &
    \boldres{\scalebox{0.85}{0.076}} & \boldres{\scalebox{0.85}{0.183}} &
    \secondres{\scalebox{0.85}{0.095}} & \secondres{\scalebox{0.85}{0.202}} &
    {\scalebox{0.85}{0.112}} & {\scalebox{0.85}{0.231}} &
    {\scalebox{0.85}{0.147}} & {\scalebox{0.85}{0.272}} &
    {\scalebox{0.85}{0.136}} & {\scalebox{0.85}{0.263}} &
    {\scalebox{0.85}{0.109}} & {\scalebox{0.85}{0.218}} &
    {\scalebox{0.85}{0.111}} & {\scalebox{0.85}{0.223}} &
    {\scalebox{0.85}{0.126}} & {\scalebox{0.85}{0.239}} \\
    & \scalebox{0.85}{24} &
    \boldres{\scalebox{0.85}{0.097}} & \boldres{\scalebox{0.85}{0.209}} &
    {\scalebox{0.85}{0.140}} & {\scalebox{0.85}{0.249}} &
    {\scalebox{0.85}{0.187}} & {\scalebox{0.85}{0.301}} &
    {\scalebox{0.85}{0.224}} & {\scalebox{0.85}{0.340}} &
    {\scalebox{0.85}{0.156}} & {\scalebox{0.85}{0.284}} &
    \secondres{\scalebox{0.85}{0.138}} & \secondres{\scalebox{0.85}{0.248}} &
    {\scalebox{0.85}{0.181}} & {\scalebox{0.85}{0.289}} &
    {\scalebox{0.85}{0.208}} & {\scalebox{0.85}{0.312}} \\
    & \scalebox{0.85}{48} &
    \boldres{\scalebox{0.85}{0.146}} & \boldres{\scalebox{0.85}{0.263}} &
    {\scalebox{0.85}{0.238}} & {\scalebox{0.85}{0.333}} &
    {\scalebox{0.85}{0.355}} & {\scalebox{0.85}{0.422}} &
    {\scalebox{0.85}{0.356}} & {\scalebox{0.85}{0.437}} &
    {\scalebox{0.85}{0.226}} & {\scalebox{0.85}{0.351}} &
    \secondres{\scalebox{0.85}{0.226}} & \secondres{\scalebox{0.85}{0.326}} &
    {\scalebox{0.85}{0.343}} & {\scalebox{0.85}{0.411}} &
    {\scalebox{0.85}{0.422}} & {\scalebox{0.85}{0.456}} \\
    & \scalebox{0.85}{96} &
    \boldres{\scalebox{0.85}{0.233}} & \boldres{\scalebox{0.85}{0.342}} &
    {\scalebox{0.85}{0.394}} & {\scalebox{0.85}{0.449}} &
    {\scalebox{0.85}{0.638}} & {\scalebox{0.85}{0.587}} &
    {\scalebox{0.85}{0.453}} & {\scalebox{0.85}{0.504}} &
    {\scalebox{0.85}{0.308}} & {\scalebox{0.85}{0.417}} &
    \secondres{\scalebox{0.85}{0.358}} & \secondres{\scalebox{0.85}{0.430}} &
    {\scalebox{0.85}{0.584}} & {\scalebox{0.85}{0.554}} &
    {\scalebox{0.85}{0.794}} & {\scalebox{0.85}{0.658}} \\
    \cmidrule(lr){2-18}
    & \scalebox{0.85}{Avg} &
    \boldres{\scalebox{0.85}{0.138}} & \boldres{\scalebox{0.85}{0.249}} &
    {\scalebox{0.85}{0.216}} & {\scalebox{0.85}{0.308}} &
    {\scalebox{0.85}{0.323}} & {\scalebox{0.85}{0.385}} &
    {\scalebox{0.85}{0.295}} & {\scalebox{0.85}{0.388}} &
    {\scalebox{0.85}{0.206}} & {\scalebox{0.85}{0.328}} &
    \secondres{\scalebox{0.85}{0.207}} & \secondres{\scalebox{0.85}{0.305}} &
    {\scalebox{0.85}{0.304}} & {\scalebox{0.85}{0.369}} &
    {\scalebox{0.85}{0.387}} & {\scalebox{0.85}{0.426}} \\
    \midrule

\multirow{5}{*}{\rotatebox{90}{\scalebox{0.95}{PEMS07}}} 
    & \scalebox{0.85}{12} &
    \boldres{\scalebox{0.85}{0.081}} & \boldres{\scalebox{0.85}{0.187}} &
    \secondres{\scalebox{0.85}{0.086}} & \secondres{\scalebox{0.85}{0.190}} &
    {\scalebox{0.85}{0.108}} & {\scalebox{0.85}{0.225}} &
    {\scalebox{0.85}{0.152}} & {\scalebox{0.85}{0.274}} &
    {\scalebox{0.85}{0.175}} & {\scalebox{0.85}{0.273}} &
    {\scalebox{0.85}{0.099}} & {\scalebox{0.85}{0.205}} &
    {\scalebox{0.85}{0.103}} & {\scalebox{0.85}{0.212}} &
    {\scalebox{0.85}{0.117}} & {\scalebox{0.85}{0.230}} \\
    & \scalebox{0.85}{24} &
    \boldres{\scalebox{0.85}{0.120}} & \boldres{\scalebox{0.85}{0.227}} &
    {\scalebox{0.85}{0.135}} & {\scalebox{0.85}{0.240}} &
    {\scalebox{0.85}{0.173}} & {\scalebox{0.85}{0.284}} &
    {\scalebox{0.85}{0.246}} & {\scalebox{0.85}{0.351}} &
    {\scalebox{0.85}{0.212}} & {\scalebox{0.85}{0.307}} &
    \secondres{\scalebox{0.85}{0.127}} & \secondres{\scalebox{0.85}{0.234}} &
    {\scalebox{0.85}{0.174}} & {\scalebox{0.85}{0.279}} &
    {\scalebox{0.85}{0.196}} & {\scalebox{0.85}{0.302}} \\
    & \scalebox{0.85}{48} &
    \boldres{\scalebox{0.85}{0.198}} & \boldres{\scalebox{0.85}{0.297}} &
    {\scalebox{0.85}{0.247}} & {\scalebox{0.85}{0.333}} &
    {\scalebox{0.85}{0.341}} & {\scalebox{0.85}{0.404}} &
    {\scalebox{0.85}{0.438}} & {\scalebox{0.85}{0.469}} &
    {\scalebox{0.85}{0.296}} & {\scalebox{0.85}{0.375}} &
    \secondres{\scalebox{0.85}{0.219}} & \secondres{\scalebox{0.85}{0.312}} &
    {\scalebox{0.85}{0.332}} & {\scalebox{0.85}{0.396}} &
    {\scalebox{0.85}{0.402}} & {\scalebox{0.85}{0.443}} \\
    & \scalebox{0.85}{96} &
    \boldres{\scalebox{0.85}{0.352}} & \boldres{\scalebox{0.85}{0.401}} &
    {\scalebox{0.85}{0.486}} & {\scalebox{0.85}{0.478}} &
    {\scalebox{0.85}{0.567}} & {\scalebox{0.85}{0.534}} &
    {\scalebox{0.85}{0.672}} & {\scalebox{0.85}{0.564}} &
    {\scalebox{0.85}{0.463}} & {\scalebox{0.85}{0.481}} &
    \secondres{\scalebox{0.85}{0.410}} & \secondres{\scalebox{0.85}{0.434}} &
    {\scalebox{0.85}{0.648}} & {\scalebox{0.85}{0.556}} &
    {\scalebox{0.85}{0.854}} & {\scalebox{0.85}{0.657}} \\
    \cmidrule(lr){2-18}
    & \scalebox{0.85}{Avg} &
    \boldres{\scalebox{0.85}{0.187}} & \boldres{\scalebox{0.85}{0.278}} &
    {\scalebox{0.85}{0.238}} & {\scalebox{0.85}{0.310}} &
    {\scalebox{0.85}{0.297}} & {\scalebox{0.85}{0.361}} &
    {\scalebox{0.85}{0.377}} & {\scalebox{0.85}{0.414}} &
    {\scalebox{0.85}{0.286}} & {\scalebox{0.85}{0.359}} &
    \secondres{\scalebox{0.85}{0.213}} & \secondres{\scalebox{0.85}{0.296}} &
    {\scalebox{0.85}{0.314}} & {\scalebox{0.85}{0.360}} &
    {\scalebox{0.85}{0.392}} & {\scalebox{0.85}{0.408}} \\
    \midrule

    \multirow{1}{*}{\rotatebox{90}{\scalebox{0.95}{-}}} 
    & \scalebox{0.85}{Count} &
    \boldres{\scalebox{0.85}{15}}&\boldres{\scalebox{0.85}{15}}&
    {\scalebox{0.85}{0}}&{\scalebox{0.85}{0}}&
    {\scalebox{0.85}{0}}&{\scalebox{0.85}{0}}&
    {\scalebox{0.85}{0}}&{\scalebox{0.85}{0}}&
    {\scalebox{0.78}{0}}&{\scalebox{0.78}{0}}&
    {\scalebox{0.85}{0}}&{\scalebox{0.85}{0}}&
    {\scalebox{0.85}{0}}&{\scalebox{0.85}{0}}&
    {\scalebox{0.85}{0}}&{\scalebox{0.85}{0}} \\
  \bottomrule
  \end{tabular}
    \end{small}
  \end{threeparttable}
  }
\end{table}

\begin{table}[htbp]
\caption{Results on short-term time series forecasting are obtained from five random seeds.}
\label{tab:seed2}
\vspace{5.5pt}
\centering
\renewcommand{\arraystretch}{1.1}
\begin{adjustbox}{width=\linewidth}
\begin{tabular}{l|cc|cc|cc}
\toprule
Dataset & \multicolumn{2}{c|}{PEMS03} & \multicolumn{2}{c|}{PEMS04} & \multicolumn{2}{c}{PEMS08} \\
\cmidrule(lr){2-3}\cmidrule(lr){4-5}\cmidrule(lr){6-7}
Horizon & MSE & MAE & MSE & MAE & MSE & MAE \\
\midrule
12  & $0.067\pm0.000$ & $0.174\pm0.000$ & $0.076\pm0.000$ & $0.183\pm0.000$ & $0.081\pm0.002$ & $0.187\pm0.002$ \\
24 & $0.093\pm0.001$ & $0.204\pm0.000$ & $0.097\pm0.001$ & $0.209\pm0.002$ & $0.120\pm0.001$ & $0.227\pm0.001$ \\
48 & $0.151\pm0.003$ & $0.261\pm0.001$ & $0.146\pm0.004$ & $0.263\pm0.004$ & $0.198\pm0.002$ & $0.297\pm0.002$ \\
96 & $0.241\pm0.000$ & $0.341\pm0.003$ & $0.233\pm0.005$ & $0.342\pm0.005$ & $0.352\pm0.006$ & $0.401\pm0.002$ \\
\bottomrule
\end{tabular}
\end{adjustbox}
\end{table}

\section{Additional Ablation Studies}\label{app:G}
To investigate the functional rationale of HTME components, We examined three different variants of iTransformer, DLinear, and SegRNN, as representatives of Transformer-based, Linear-based, and RNN-based models, respectively.
1) \textbf{Productor}, which integrates the multivariate feature extraction module of the HTME extractor into the embedding layer of iTransformer.
2) \textbf{ProductorV1}, which utilizes only the multivariate feature extraction module of the HTME extractor. 
We also examined another variant 3) \textbf{ProductorV2}, which utilizes the multivariate feature extraction module of the HTME extractor to supplement the embedding representations.
We introduce two Additional variants: \textbf{HTMformer}, builds upon iTransformer by incorporating the complete HTME extractor; \textbf{HTMformerV1}, replaces the original embedding layer in iTransformer with the temporal feature extractor from HTME.
The detailed results are reported in Table~\ref{tab:ablation} of the main text. 
As all experiments were conducted under identical configurations, the HTMPredictors are directly comparable to other models.

\paragraph{Effect of the temporal feature extraction module.} By comparing iTransformer and HTMformerV1, \textcolor{blue}{blue} numbers are used to denote cases where HTMformerV1 outperforms iTransformer. 
Observations indicate that HTMformerV1 demonstrates improvements in 63/80 cases.
\paragraph{Effect of the temporal feature.} 
By comparing Predictor and PredictorV2, numbers prefixed with an \underline{underline} indicate that Predictor outperforms PredictorV2. Notably, Predictor performs better in most cases. The majority of time series features are stored in the temporal dimension, and the depth of temporal feature mining substantially influences prediction accuracy.
This indicates that the temporal feature extraction module of HTME is more effective.
\paragraph{Effect of the multivariate features.}  Using Predictor as baselines, we compare them against PredictorV1. 
\textcolor{red}{Red} highlighted numbers indicate cases where the multivariate-integrated versions outperform their original counterparts.
Experimental results show that PredictorV1 achieves superior performance most cases. 
We also find that PredictorV2 outperforms Predictor in some cases.
The spatial dimension (multivariate dimension) also contains abundant features, and comprehensive multivariate feature mining can further reduce errors.

\paragraph{Effect of the hybrid strategy.} Among all models and their variants, PredictorV1 achieves superior performance.
This demonstrates the effectiveness of the HTME strategy.
In most cases, temporal features are the primary characteristics of time series data, but the correlations between multiple variables cannot be overlooked. 
The HTME strategy has successfully suppressed noise and extracted effective patterns from the raw sequence.
In conclusion, the strategy of fusing temporal features and multivariate features can enhance the accuracy and scalability of Transformer-based predictors.

\begin{table}[htbp]
  \caption{We compare different variants under different prediction lengths on multiple datasets (MSE and MAE). The input sequence length is set to 96. }
  \vspace{3pt}
  \centering
  \label{tab:ablation}
  \resizebox{1\columnwidth}{!}{
  \begin{threeparttable}
  \begin{small}
  \renewcommand{\multirowsetup}{\centering}
  \begin{tabular}{c|c|cc|cc|cc|cc|cc|cc|cc|cc|cc|cc|cc}
    \toprule
    \multicolumn{2}{c}{\multirow{1}{*}{Models}} & 
    \multicolumn{2}{c}{\rotatebox{0}{\scalebox{1}{HTMformer}}} &
    \multicolumn{2}{c}{\rotatebox{0}{\scalebox{1}{HTMformerV1}}} &
    \multicolumn{2}{c}{\rotatebox{0}{\scalebox{1}{iTransV2}}} &
    \multicolumn{2}{c}{\rotatebox{0}{\scalebox{1}{iTransformer}}} &
    \multicolumn{2}{c}{\rotatebox{0}{\scalebox{1}{iTransV1}}} &
    \multicolumn{2}{c}{\rotatebox{0}{\scalebox{1}{DLinearV2}}}&
    \multicolumn{2}{c}{\rotatebox{0}{\scalebox{1}{DLinear}}}&
    \multicolumn{2}{c}{\rotatebox{0}{\scalebox{1}{DLinearV1}}}&
    \multicolumn{2}{c}{\rotatebox{0}{\scalebox{1}{SegRNNV2}}}&
    \multicolumn{2}{c}{\rotatebox{0}{\scalebox{1}{SegRNN}}}&
    \multicolumn{2}{c}{\rotatebox{0}{\scalebox{1}{SegRNNV1}}}\\
    \midrule
    \multicolumn{2}{c|}{Metric} & MSE & MAE & MSE & MAE & MSE & MAE& MSE & MAE& MSE & MAE& MSE & MAE& MSE & MAE& MSE & MAE& MSE & MAE& MSE & MAE& MSE & MAE\\
    \midrule
    \multirow{5}{*}{ECL} 
    & 96 & 0.157 & 0.247 & \textcolor{blue}{\underline{0.153}} & \textcolor{blue}{\underline{0.246}} & 0.255 & 0.34 & 0.163 & 0.252 & 0.165 & 0.262 &0.499	&0.533& 0.210 & 0.301&\textcolor{red}{0.204} &\textcolor{red}{0.299} &0.390 &0.454&0.168 &0.264&\textcolor{red}{0.164} &\textcolor{red}{0.258}\\
    & 192 & 0.171 & \textcolor{red}{0.260} & \textcolor{blue}{\underline{0.169}} & \textcolor{blue}{\underline{0.260}} & 0.264 & 0.349 & 0.175 & 0.263 & 0.184 & 0.278 &0.505	&0.535& 0.210 & 0.304 &0.211 &0.307&0.407 &0.464&0.183 &0.279&\textcolor{red}{0.175} &\textcolor{red}{0.269}\\
    & 336 & 0.192 & \textcolor{red}{0.277} & \textcolor{blue}{\underline{0.187}} & \textcolor{blue}{\underline{0.277}} & 0.273 & 0.357 & 0.282 & 0.299 & \textcolor{red}{0.199} & \textcolor{red}{0.294} &0.504	&0.531& 0.223 & 0.319 &\textcolor{red}{0.222} &\textcolor{red}{0.318}&0.415 &0.470&0.192 &0.289&\textcolor{red}{0.192} &\textcolor{red}{0.285}\\
    & 720 & \textcolor{red}{0.220} & \textcolor{red}{0.304} & \textcolor{blue}{\underline{0.227}} & \textcolor{blue}{\underline{0.311}} & 0.31 & 0.384 & 0.357 & 0.348 & \textcolor{red}{0.231} & \textcolor{red}{0.318} &0.491	&0.526& 0.257 & 0.349 &\textcolor{red}{0.240} &\textcolor{red}{0.331}&0.452 &0.491&0.231 &0.321&\textcolor{red}{0.230} &\textcolor{red}{0.320}\\
    \cmidrule(lr){2-24}
    & AVG & 0.185 & \textcolor{red}{0.272} & \textcolor{blue}{\underline{0.184}} & \textcolor{blue}{\underline{0.273}} & 0.275 & 0.357 & 0.244 & 0.29 & \textcolor{red}{0.194} & \textcolor{red}{0.288} &0.500	&0.531& 0.225 & 0.318 &\textcolor{red}{0.219} &\textcolor{red}{0.314}&0.416 &0.470&0.194 &0.288&\textcolor{red}{0.190} &\textcolor{red}{0.283}\\
    \midrule
    \multirow{5}{*}{Weather} 
    & 96 & \textcolor{red}{0.164} & \textcolor{red}{0.209} & \textcolor{blue}{\underline{0.169}} & \textcolor{blue}{\underline{0.212}} & 0.175 & 0.229 & 0.176 & 0.216 & \textcolor{red}{0.167} & \textcolor{red}{0.167} &0.18	&0.272 &0.193	&0.233 &\textcolor{red}{0.177} &\textcolor{red}{0.223}&0.165 &0.231&0.168 &0.230&\textcolor{red}{0.161} &\textcolor{red}{0.226}\\
    & 192 & 0.222 & 0.261 & \textcolor{blue}{\underline{0.218}} & \textcolor{blue}{\underline{0.256}} & 0.223 & 0.264 & 0.223 & 0.255 & \textcolor{red}{0.217} & 0.256 &0.232	&0.320&0.236	&0.268 &\textcolor{red}{0.227} &\textcolor{red}{0.266}&0.221 &0.285&0.215 &0.277&\textcolor{red}{0.211} &\textcolor{red}{0.273}\\
    & 336 & \textcolor{red}{0.272} & \textcolor{red}{0.290} & \textcolor{blue}{\underline{0.275}} & \textcolor{blue}{\underline{0.298}} & 0.287 & 0.31 & 0.28 & 0.298 & \textcolor{red}{0.272} & \textcolor{red}{0.297} &0.294	&0.365&0.288	&0.304 &\textcolor{red}{0.281} &\textcolor{red}{0.303}&0.281 &0.330&0.274 &0.319&\textcolor{red}{0.269} &\textcolor{red}{0.315}\\
    & 720 & 0.359 & 0.352 & \textcolor{blue}{\underline{0.353}} & \textcolor{blue}{\underline{0.349}} & 0.364 & 0.358 & 0.357 & 0.349 & \textcolor{red}{0.353} & \textcolor{red}{0.349} &0.391	&0.433&0.359	&0.349 &\textcolor{red}{0.358} &\textcolor{red}{0.352}&0.364 &0.385&0.359 &0.378&\textcolor{red}{0.355} &\textcolor{red}{0.371}\\
    \cmidrule(lr){2-24}
    & AVG & 0.254 & \textcolor{red}{0.277} & \textcolor{blue}{\underline{0.253}} & \textcolor{blue}{\underline{0.278}} & 0.262 & 0.29 & 0.259 & 0.279 & \textcolor{red}{0.252} & \textcolor{red}{0.267} &0.274	&0.348&0.269	&0.289 &\textcolor{red}{0.261} &\textcolor{red}{0.286}&0.258 &0.308&0.254 &0.301&\textcolor{red}{0.249} &\textcolor{red}{0.296}\\
    \midrule
    \multirow{5}{*}{Traffic} 
    & 96 & \textcolor{red}{0.439} & \textcolor{red}{0.300} & \underline{0.447} & \underline{0.303} & 0.648 & 0.405 & 0.442 & 0.302 & \textcolor{red}{0.440} & 0.309 &0.955	&0.544&0.696	&0.428 &\textcolor{red}{0.626} &\textcolor{red}{0.393}&1.224 &0.583&0.723 &0.376&\textcolor{red}{0.696} &\textcolor{red}{0.366}\\
    & 192 & \textcolor{red}{0.450} & \textcolor{red}{0.302} & \underline{0.460} & \textcolor{blue}{\underline{0.307}} & 0.737 & 0.438 & 0.459 & 0.308 & \textcolor{red}{0.457} & \textcolor{red}{0.305} &0.970	&0.543&0.646	&0.407&\textcolor{red}{0.583} &\textcolor{red}{0.369} &1.201 &0.568&0.737 &0.385&\textcolor{red}{0.717} &\textcolor{red}{0.382}\\
    & 336 & \textcolor{red}{0.462} & \textcolor{red}{0.310} & \underline{0.482} & \textcolor{blue}{\underline{0.319}} & 0.745 & 0.435 & 0.479 & 0.319 & \textcolor{red}{0.478} & 0.321 &0.982	&0.550&0.653	&0.409 &\textcolor{red}{0.591} &\textcolor{red}{0.373}&1.285 &0.595&0.766 &0.405&\textcolor{red}{0.754} &\textcolor{red}{0.401}\\
    & 720 & \textcolor{red}{0.517} & \textcolor{red}{0.339} & \underline{0.522} & \underline{0.343} & 0.863 & 0.484 & 0.516 & 0.342 & 0.520 & 0.349 &0.997	&0.548&0.694	&0.428 &\textcolor{red}{0.636} &\textcolor{red}{0.395}&1.343 &0.614&0.818 &0.423&\textcolor{red}{0.803} &\textcolor{red}{0.419}\\
    \cmidrule(lr){2-24}
    & AVG & \textcolor{red}{0.467} & \textcolor{red}{0.312} & \underline{0.477} & \underline{0.318} & 0.748 & 0.44 & 0.474 & 0.317 & \textcolor{red}{0.473} & 0.321 &0.976	&0.546&0.672	&0.418&\textcolor{red}{0.609} &\textcolor{red}{0.383}&1.263 &0.590 &0.761 &0.397&\textcolor{red}{0.750} &\textcolor{red}{0.392}\\
    \midrule
    \multirow{5}{*}{ETTh2} 
    & 96 & 0.300 & \textcolor{red}{0.348} & \underline{0.298} & \underline{0.348} & 0.375 & 0.399 & 0.295 & 0.344 & 0.304 & 0.352 &0.357 &0.409&0.341	&0.395 &\textcolor{red}{0.343} &\textcolor{red}{0.399}&0.339 &0.389&0.295 &0.355&\textcolor{red}{0.292} &\textcolor{red}{0.345}\\
    & 192 & 0.389 & \textcolor{red}{0.402} & \underline{0.387} & \underline{0.402} & 0.45 & 0.435 & 0.375 & 0.398 & 0.391 & 0.406 &0.497 &0.491 &0.481	&0.479&\textcolor{red}{0.480} &\textcolor{red}{0.477}&0.417 &0.431&0.375 &0.397&\textcolor{red}{0.371} &\textcolor{red}{0.396}\\
    & 336 & \textcolor{red}{0.409} & \textcolor{red}{0.408} & \textcolor{blue}{\underline{0.416}} & \textcolor{blue}{\underline{0.427}} & 0.481 & 0.469 & 0.436 & 0.435 & \textcolor{red}{0.419} & \textcolor{red}{0.428} &0.610 &0.553 &0.592	&0.542&\textcolor{red}{0.597} &\textcolor{red}{0.540}&0.476 &0.476&0.414 &0.436&0.416 &\textcolor{red}{0.430}\\
    & 720 & \textcolor{red}{0.421} & \textcolor{red}{0.439} & \textcolor{blue}{\underline{0.426}} & \textcolor{blue}{\underline{0.443}} & 0.471 & 0.474 & 0.429 & 0.447 & \textcolor{red}{0.422} & \textcolor{red}{0.438} &0.816 &0.658 &0.840	&0.661&0.820 &\textcolor{red}{0.652}&0.474 &0.489&0.443 &0.469&\textcolor{red}{0.436} &\textcolor{red}{0.462}\\
    \cmidrule(lr){2-24}
    & AVG & \textcolor{red}{0.379} & \textcolor{red}{0.399} & \underline{0.381} & \underline{0.405} & 0.444 & 0.444 & 0.383 & 0.406 & 0.384 & \textcolor{red}{0.406} &0.570 &0.528 &0.563	&0.519&\textcolor{red}{0.560} &\textcolor{red}{0.517}&0.427 &0.446&0.382 &0.414&\textcolor{red}{0.379} &\textcolor{red}{0.408}\\
    \midrule
    \multirow{5}{*}{Solar} 
    & 96 & \textcolor{red}{0.199} & \textcolor{red}{0.239} & \textcolor{blue}{\underline{0.209}} & \textcolor{blue}{\underline{0.252}} & 0.244 & 0.284 & 0.213 & 0.253 & \textcolor{red}{0.207} & \textcolor{red}{0.249} &0.231 &0.326 &0.289	&0.377&\textcolor{red}{0.225} &\textcolor{red}{0.317}&0.353 &0.416&0.232 &0.294&\textcolor{red}{0.203} &\textcolor{red}{0.291}\\
    & 192 & \textcolor{red}{0.233} & \textcolor{red}{0.263} & \textcolor{blue}{\underline{0.238}} & \textcolor{blue}{\underline{0.274}} & 0.277 & 0.306 & 0.242 & 0.274 & \textcolor{red}{0.240} & 0.279 &0.262 &0.348 &0.319	&0.397&\textcolor{red}{0.260} &\textcolor{red}{0.342}&0.404 &0.448&0.253 &0.306&\textcolor{red}{0.230} &\textcolor{red}{0.303}\\
    & 336 & \textcolor{red}{0.252} & \textcolor{red}{0.279} & \textcolor{blue}{\underline{0.261}} & \underline{0.292} & 0.323 & 0.326 & 0.262 & 0.29 & \textcolor{red}{0.256} & \textcolor{red}{0.290} &0.302 &0.373 &0.352	&0.415&\textcolor{red}{0.296} &\textcolor{red}{0.367}&0.418 &0.455&0.264 &0.310&\textcolor{red}{0.257} &\textcolor{red}{0.321}\\
    & 720 & \textcolor{red}{0.259} & \textcolor{red}{0.280} & \textcolor{blue}{\underline{0.265}} & \textcolor{blue}{\underline{0.295}} & 0.311 & 0.329 & 0.27 & 0.296 & 0.275 & 0.303 &0.313 &0.378 &0.356	&0.412&\textcolor{red}{0.309} &\textcolor{red}{0.373}&0.482 &0.516&0.260 &0.306&\textcolor{red}{0.281} &\textcolor{red}{0.349}\\
    \cmidrule(lr){2-24}
    & AVG & \textcolor{red}{0.235} & \textcolor{red}{0.265} & \textcolor{blue}{\underline{0.243}} & \textcolor{blue}{\underline{0.278}} & 0.288 & 0.311 & 0.246 & 0.278 & \textcolor{red}{0.244} & 0.280 &0.277 &0.356 &0.329	&0.400&\textcolor{red}{0.273} &\textcolor{red}{0.350} &0.414 &0.459&0.252 &0.304&\textcolor{red}{0.243} &0.316\\
    \midrule
    \multirow{5}{*}{PMES03} 
    & 96 & \textcolor{red}{0.250} & \textcolor{red}{0.343} & \textcolor{blue}{0.304} & \textcolor{blue}{0.393} & 0.302 & 0.363 & 0.337 & 0.413 & \textcolor{red}{0.240} & \textcolor{red}{0.340} &0.315 &0.418 &0.458	&0.517&\textcolor{red}{0.313} &\textcolor{red}{0.416}&0.389 &0.446&0.381 &0.438&\textcolor{red}{0.287} &\textcolor{red}{0.374}\\
    & 192 & \textcolor{red}{0.297} & \textcolor{red}{0.378} & \textcolor{blue}{\underline{0.353}} & \textcolor{blue}{0.424} & 0.373 & 0.422 & 0.374 & 0.436 & \textcolor{red}{0.300} & \textcolor{red}{0.384} &0.410 &0.477 &0.477	&0.527&\textcolor{red}{0.410} &\textcolor{red}{0.477}&0.620 &0.602&0.447 &0.478&\textcolor{red}{0.368} &\textcolor{red}{0.433}\\
    & 336 & \textcolor{red}{0.274} & \textcolor{red}{0.356} & \textcolor{blue}{\underline{0.318}} & \textcolor{blue}{\underline{0.388}} & 0.32 & 0.388 & 0.332 & 0.395 & \textcolor{red}{0.286} & \textcolor{red}{0.365} &0.337 &0.422 &0.396	&0.456&0.344 &\textcolor{red}{0.420}&0.729 &0.641&0.380 &0.432&\textcolor{red}{0.349} &\textcolor{red}{0.418}\\
    & 720 & \textcolor{red}{0.336} & \textcolor{red}{0.401} & \textcolor{blue}{\underline{0.387}} & \textcolor{blue}{\underline{0.438}} & 0.409 & 0.44 & 0.397 & 0.44 & \textcolor{red}{0.352} & \textcolor{red}{0.414} &0.391 &0.469 &0.439	&0.492&0.396 &\textcolor{red}{0.462}&0.774 &0.683&0.447 &0.478&\textcolor{red}{0.406} &\textcolor{red}{0.472}\\
    \cmidrule(lr){2-24}
    & AVG & \textcolor{red}{0.289} & \textcolor{red}{0.369} & \textcolor{blue}{\underline{0.340}} & \textcolor{blue}{0.410} & 0.351 & 0.403 & 0.36 & 0.421 & \textcolor{red}{0.294} & \textcolor{red}{0.375} &0.363 &0.447 &0.442	&0.498&0.366 &\textcolor{red}{0.444}&0.628 &0.593&0.413 &0.456&\textcolor{red}{0.353} &\textcolor{red}{0.424}\\
    \midrule
    \multirow{5}{*}{PMES04} 
    & 96 & \textcolor{red}{0.226} & \textcolor{red}{0.335} & \textcolor{blue}{0.364} & \textcolor{blue}{0.433} & 0.236 & 0.348 & 0.387 & 0.446 & \textcolor{red}{0.235} & \textcolor{red}{0.344} &0.315 &0.404 &0.452	&0.504&\textcolor{red}{0.312} &\textcolor{red}{0.401}&0.370 &0.457&0.415 &0.468&\textcolor{red}{0.291} &\textcolor{red}{0.398}\\
    & 192 & \textcolor{red}{0.286} & \textcolor{red}{0.383} & \textcolor{blue}{0.406} & \textcolor{blue}{\underline{0.406}} & 0.326 & 0.413 & 0.429 & 0.473 & \textcolor{red}{0.307} & \textcolor{red}{0.395} &0.377 &0.451 &0.477	&0.527&\textcolor{red}{0.372} &\textcolor{red}{0.451}&0.723 &0.665&0.496 &0.510&\textcolor{red}{0.463} &\textcolor{red}{0.507}\\
    & 336 & \textcolor{red}{0.286} & \textcolor{red}{0.372} & \textcolor{blue}{0.359} & \textcolor{blue}{0.417} & 0.312 & 0.398 & 0.369 & 0.426 & \textcolor{red}{0.297} & \textcolor{red}{0.378} &0.330 &0.411 &0.396	&0.456&\textcolor{red}{0.330} &\textcolor{red}{0.410}&0.789 &0.694&0.420 &0.461&0.453 &0.508\\
    & 720 & \textcolor{red}{0.338} & \textcolor{red}{0.414} & \textcolor{blue}{0.436} & \textcolor{blue}{0.471} & 0.387 & 0.451 & 0.44 & 0.476 & \textcolor{red}{0.362} & \textcolor{red}{0.428} &0.401 &0.460 &0.439	&0.492&\textcolor{red}{0.400} &\textcolor{red}{0.460}&0.850 &0.729&0.479 &0.502&0.714 &0.677\\
    \cmidrule(lr){2-24}
    & AVG & \textcolor{red}{0.284} & \textcolor{red}{0.376} & \textcolor{blue}{0.391} & \textcolor{blue}{0.431} & 0.315 & 0.402 & 0.406 & 0.455 & \textcolor{red}{0.300} & \textcolor{red}{0.386} &0.356 &0.432 &0.441	&0.494&\textcolor{red}{0.354} &\textcolor{red}{0.431}&0.683 &0.636&0.452 &0.485&0.480 &0.522\\
    \midrule
    \multirow{5}{*}{PMES08} 
    & 96 & \textcolor{red}{0.365} & \textcolor{red}{0.409} & \textcolor{blue}{0.469} & \textcolor{blue}{0.461} & 0.401 & 0.416 & 0.477 & 0.47 & \textcolor{red}{0.338} & \textcolor{red}{0.388} &0.449 &0.446 &0.672	&0.564&0.454 &0.451&0.429 &0.477&0.516 &0.487&\textcolor{red}{0.364} &\textcolor{red}{0.425}\\
    & 192 & \textcolor{red}{0.513} & \textcolor{red}{0.462} & \textcolor{blue}{0.612} & \textcolor{blue}{0.507} & 0.567 & 0.481 & 0.627 & 0.518 & \textcolor{red}{0.487} & \textcolor{red}{0.444} &0.599 &0.536 &0.726	&0.580&\textcolor{red}{0.597} &\textcolor{red}{0.535}&0.811 &0.674&0.646 &0.528&\textcolor{red}{0.568} &\textcolor{red}{0.523}\\
    & 336 & \textcolor{red}{0.530} & \textcolor{red}{0.435} & 0.606 & \textcolor{blue}{0.471} & 0.567 & 0.452 & 0.599 & 0.472 & \textcolor{red}{0.516} & \textcolor{red}{0.425} &0.569 &0.476 &0.650	&0.518&\textcolor{red}{0.564} &\textcolor{red}{0.469}&0.949 &0.712&0.617 &0.490&0.621 &0.511\\
    & 720 & \textcolor{red}{0.624} & \textcolor{red}{0.487} & 0.692 & \textcolor{blue}{0.531} & 0.697 & 0.512 & 0.691 & 0.532 & \textcolor{red}{0.600} & \textcolor{red}{0.479} &0.633 &0.531 &0.713	&0.562&\textcolor{red}{0.632} &\textcolor{red}{0.521}&1.034 &0.743&0.707 &0.541&0.874 &0.652\\
    \cmidrule(lr){2-24}
    & AVG & \textcolor{red}{0.508} & \textcolor{red}{0.448} & \textcolor{blue}{0.594} & \textcolor{blue}{0.492} & 0.558 & 0.465 & 0.598 & 0.498 & \textcolor{red}{0.485} & \textcolor{red}{0.434} &0.563 &0.497 &0.690	&0.556&\textcolor{red}{0.562} &\textcolor{red}{0.494}&0.806 &0.652&0.621 &0.511&\textcolor{red}{0.607} &0.528\\
    \bottomrule
  \end{tabular}
    \end{small}
  \end{threeparttable}
}
\end{table}

\section{Details of Model Efficiency}\label{app:H}
For a comprehensive efficiency comparison, we evaluate HTMformer against three highly competitive baselines, including MultiPatchFormer~\citep{naghashi2025multipatchformer}, WPMixer~\citep{murad2024wp}, and PatchTST~\citep{nie2023time}, across eight datasets.
The evaluation considers three key metrics: training time, GPU memory footprint, with the input sequence length fixed at 96 and the prediction length at 192. 
Moreover, the experimental hardware conditions and parameter configurations follow those described in \underline{Appendix~\ref{app:E}}, and the best results are marked in \textcolor{red}{red}.
The full results are shown in Table~\ref{fig:time_memory} (\textbf{Left}) and Table~\ref{fig:time_memory} (\textbf{Right}).
Across all datasets, HTMformer achieves the best training time and GPU memory footprint.
Notably, for high-dimensional datasets, particularly when the number of variate dimensions exceeds one hundred, HTMformer delivers nearly three times the training speed of the second-best model and requires only one-third of its memory footprint. 
Moreover, this advantage becomes increasingly pronounced as dimensionality grows.
HTME is a lightweight module that enables simple models focused on temporal feature modeling to incorporate the influence of multivariate correlations, thereby achieving performance comparable to that of complex models.
\vspace{-5pt}
\begin{table}[htbp]
    \centering
    \caption{Comparison of training time~(s \textbackslash iter) and GPU memory footprint~(MB).}
    \label{fig:time_memory}
    \vspace{5.5pt}
    \begin{minipage}{0.49\textwidth}
        \resizebox{\linewidth}{!}{
            \begin{threeparttable}
            \begin{small}
            \setlength{\tabcolsep}{1pt}
            \begin{tabular}{l|c|c|c|c}
            \toprule
            Datasets & \textbf{HTMFormer} & MPFormer & WPMixer & PatchTST \\
            Models & \textbf{Ours} & \citeyear{naghashi2025multipatchformer} & \citeyear{murad2024wp} & \citeyear{nie2023time} \\
            \midrule
            ECL     & \textcolor{red}{0.1004} & 0.3319 & 0.4143 & 0.5088 \\
            Weather & \textcolor{red}{0.0282} & 0.0320 & 0.0392 & 0.0292 \\
            Traffic & \textcolor{red}{0.2682} & 0.9648 & 1.2374 & 1.0294 \\
            ETTh2   & 0.0199 & 0.0279 & \textcolor{red}{0.0183} & 0.0190 \\
            Solar   & \textcolor{red}{0.0502} & 0.1435 & 0.1745 & 0.2200 \\
            pems03  & \textcolor{red}{0.0982} & 0.3868 & 8344.82 & 0.5797 \\
            pems04  & \textcolor{red}{0.1514} & 0.5801 & 0.3894 & 0.4857 \\
            pems07  & \textcolor{red}{0.0915} & 0.3108 & 0.2154 & 0.2552 \\
            AVG     & \textcolor{red}{0.1010} & 0.3472 & 0.3849 & 0.3909 \\
            \bottomrule
            \end{tabular}
            \end{small}
            \end{threeparttable}
        }
    \end{minipage}%
    \hfill
    \begin{minipage}{0.49\textwidth}
        \resizebox{\linewidth}{!}{
            \begin{threeparttable}
            \begin{small}
            \setlength{\tabcolsep}{1pt}
            \begin{tabular}{l|c|c|c|c}
            \toprule
            Datasets & \textbf{HTMFormer} & MPFormer & WPMixer & PatchTST \\
            Models & \textbf{Ours} & \citeyear{naghashi2025multipatchformer} & \citeyear{murad2024wp} & \citeyear{nie2023time} \\
            \midrule
            ECL     & \textcolor{red}{1623.47} & 7471.94 & 13784.07  & 7195.99  \\
            Weather & \textcolor{red}{151.83}  & 584.33  & 651.45    & 538.76   \\
            Traffic & \textcolor{red}{6221.31} & 20860.23& 29560.07  & 19152.58 \\
            ETTh2   & \textcolor{red}{102.48}  & 265.89  & 455.10    & 430.54   \\
            Solar   & \textcolor{red}{647.04}  & 3219.13 & 5970.26   & 3130.73  \\
            pems03  & \textcolor{red}{1849.57} & 8344.82 & 8015.67   & 8015.67  \\
            pems04  & \textcolor{red}{1536.60} & 7141.78 & 13195.91  & 6882.86  \\
            pems07  & \textcolor{red}{778.26}  & 3979.58 & 7369.67   & 3869.06  \\
            AVG     & \textcolor{red}{1613.82} & 6483.46 & 10125.27  & 6152.02  \\
            \bottomrule
            \end{tabular}
            \end{small}
            \end{threeparttable}
        }
    \end{minipage}
\end{table}
\newpage
\section{Analysis of HTME}\label{app:feature}
The vast majority of features in time series are contained within the temporal dimension, and the depth of exploration in this dimension significantly influences prediction accuracy~\citep{hyndman2021forecasting}. Therefore, existing models can achieve superior performance even by focusing solely on the temporal dimension. Moreover, thanks to their simple model architectures, they often have a low computational overhead.

The multivariate dimension also contains a large number of features~\citep{Pravilovic2013SpatialForecasting}, and fully exploiting multivariate features can further boost performance. 
For time series data, the vast majority of features are latent within the temporal dimension, making the deep extraction of temporal features the key to effective modeling. Overemphasizing multivariate features—particularly strategies that explicitly model cross-variable correlations within the backbone network—introduces significant computational overhead without a corresponding performance gain.
Most importantly, introducing additional complex modules into the backbone network to model multivariate correlations forces temporal and multivariate features to share the same feature space. Consequently, while the model is capturing cross-variable relationships, the representation of temporal features is inevitably compromised.

We visualize the temporal and multivariate feature spaces of HTME, as well as the weight parameter alpha. The value of alpha is approximately 0.65, indicating that subsequent modules are more inclined to select temporal features. Additionally, temporal feature weights are considerably higher than spatial feature weights. This emphasizes the primary role of temporal information. 

We also observe that the temporal features exhibit a strip-like distribution, showing significant variation along the temporal dimension but minimal differences in the multivariate dimension.
This suggests that the temporal features possess a high degree of consistency across the multivariate dimension, which in turn underscores their importance.
The multivariate features exhibit a point-like and uniform distribution. For the most part, they take on small values, which ensures that they do not disrupt the temporal feature space.
A few points within the multivariate features exhibit large values. These points serve to emphasize their corresponding temporal features, thereby creating a more information-rich temporal feature space. Consequently, subsequent modules can utilize this enhanced feature space more efficiently, improving the model's capability to capture complex temporal dynamics.
\section{Hyperparameter Sensitivity}\label{app:I}
\begin{figure}[ht]
    \centering

    \begin{minipage}{\textwidth}
        \centering
        \includegraphics[width=\textwidth]{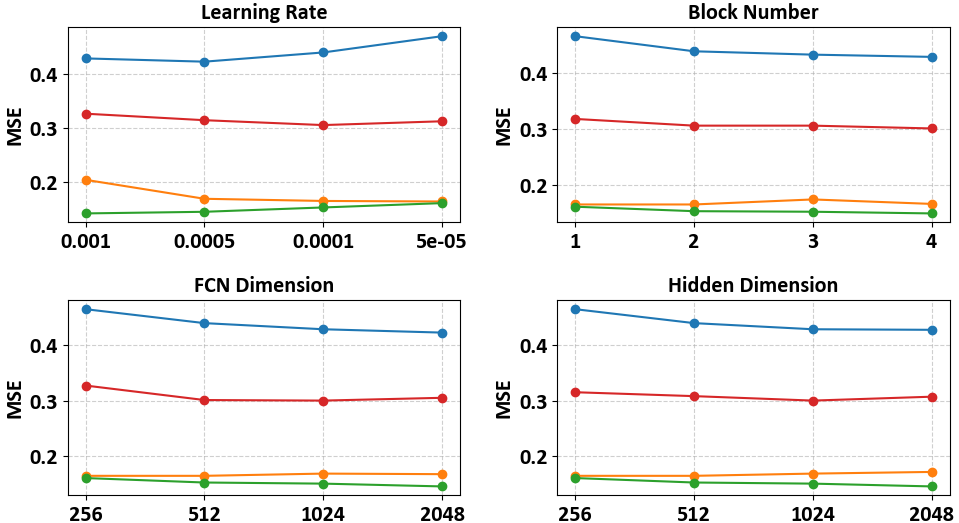}
    \end{minipage}
    
    \vspace{0.5em} 

    \begin{minipage}{\textwidth}
        \centering
        \includegraphics[width=\textwidth]{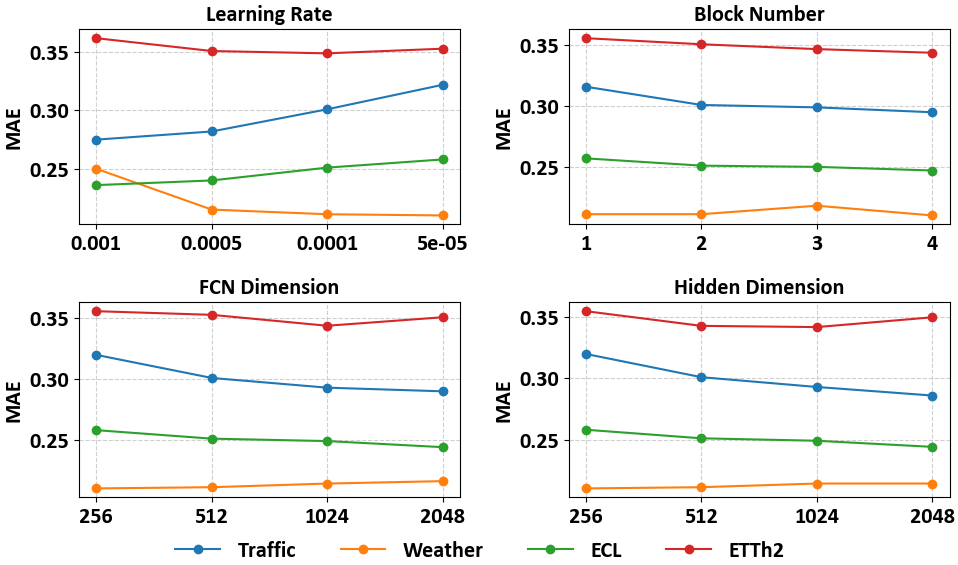}
    \end{minipage}

    \caption{Hyperparameter sensitivity with respect to $lr$, $L$, $K$, and $D$ . All results are obtained using a lookback window of $T = 96$ time steps and a forecast horizon of $S = 96$ time steps.}
    \label{fig:myplot6and7}
\end{figure}

We assess the sensitivity of HTMformer to variations in key hyperparameters, including the learning rate $lr$, the number of Transformer blocks $L$, the FCN dimension $K$, and the hidden dimension $D$ of the variate tokens.
This study can provide guidance for hyperparameter selection of HTMformer in practical applications.
As shown in Figure~\ref{fig:myplot6and7}, our analysis leads to the following key observations:
Increasing hyperparameter values does not necessarily translate into improved forecasting accuracy.  The model attains its best performance on most datasets when configured with $lr=0.0005$, $L=2$, $K=1024$, and $D=512$.
\section{Visualization of Prediction Results}\label{app:J}
\begin{figure}[htbp]  
    \centering  
    \includegraphics[width=1.0\textwidth]{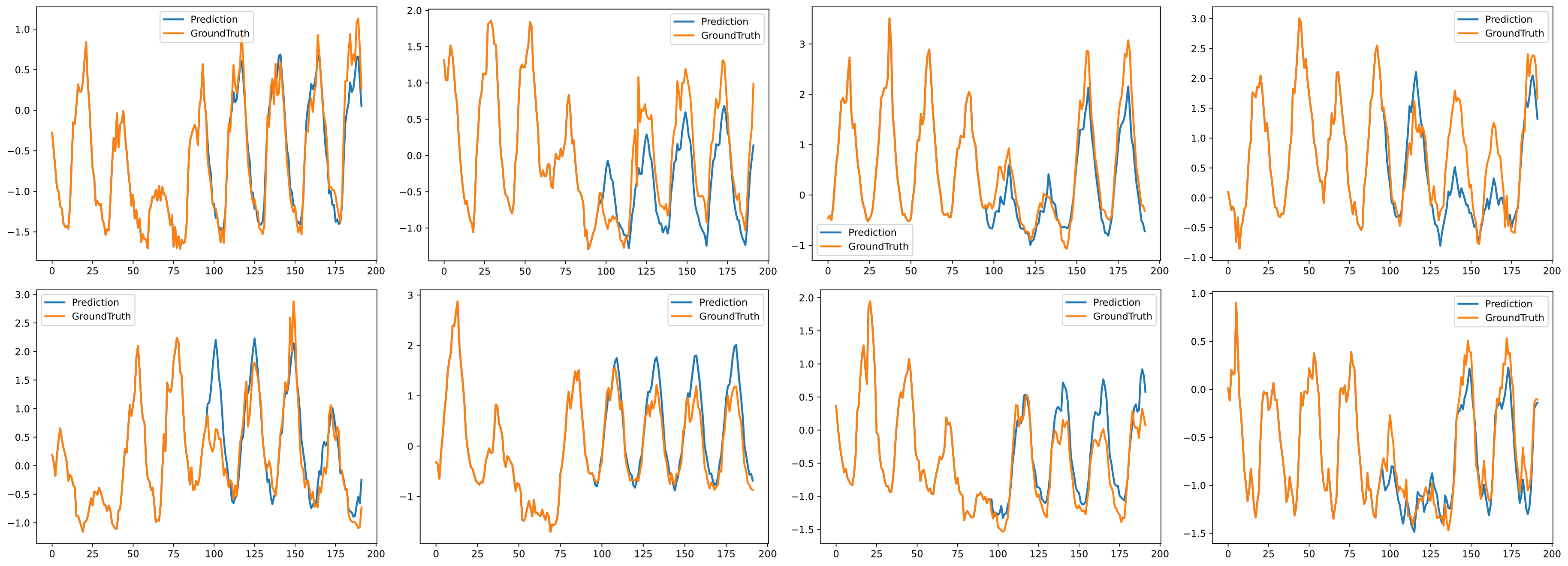}  
    \caption{Examples of forecasts for the ECL dataset with a 96-step predictions.}  
    \label{fig:ECL}  
\end{figure}

\begin{figure}[htbp] 
    \centering  
    \includegraphics[width=1.0\textwidth]{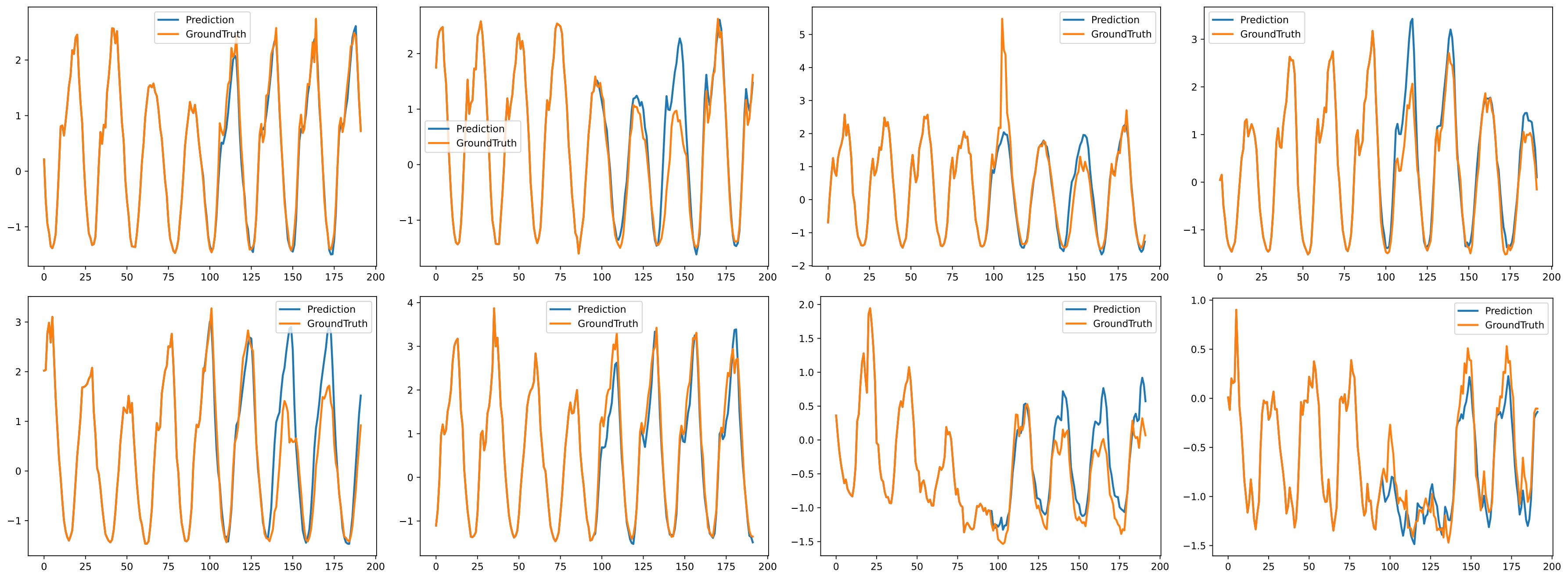}  
    \caption{Examples of forecasts for the Traffic dataset with a 96-step predictions.}  
    \label{fig:traffic}  
\end{figure}

\begin{figure}[htbp]  
    \centering  
    \includegraphics[width=1.0\textwidth]{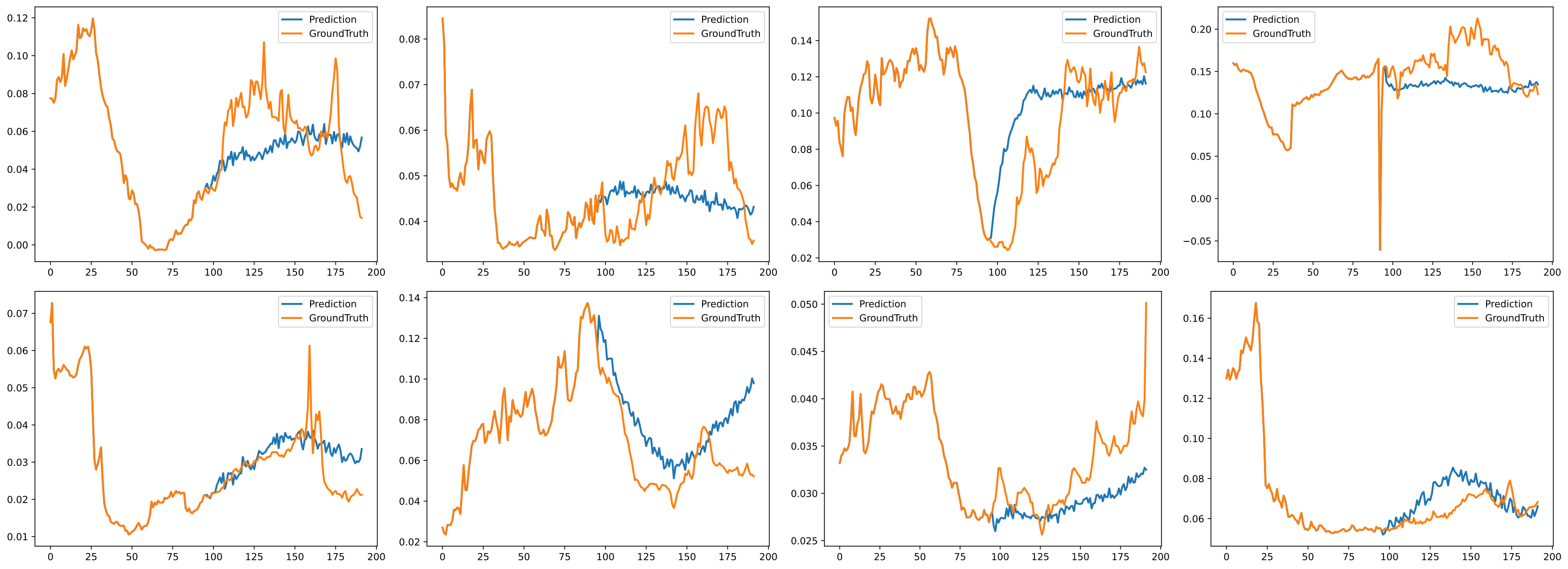} 
    \caption{Examples of forecasts for the Weather dataset with a 96-step predictions.}  
    \label{fig:weather}  
\end{figure}

\begin{figure}[htbp]  
    \centering  
    \includegraphics[width=1.0\textwidth]{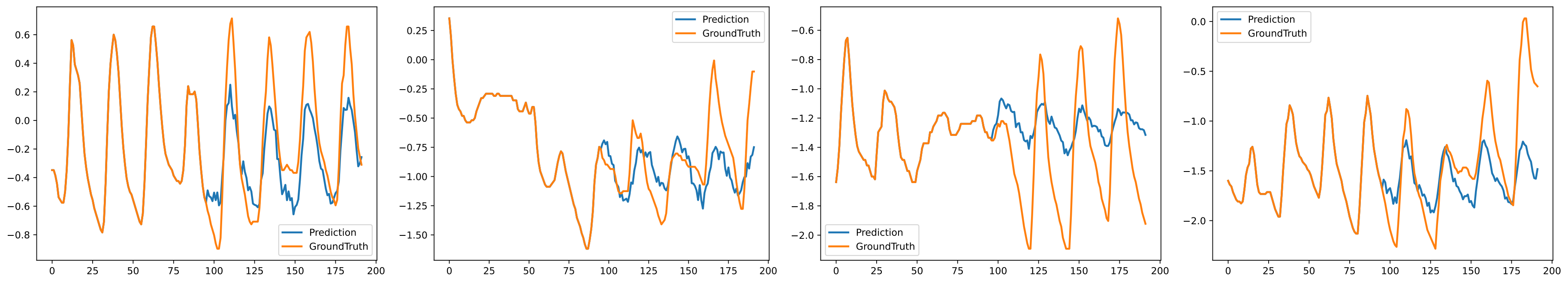}  
    \caption{Examples of forecasts for the ETTh2 dataset with a 96-step predictions.}  
    \label{fig:ETTh2}  
\end{figure}

\begin{figure}[htbp]  
    \centering 
    \includegraphics[width=1.0\textwidth]{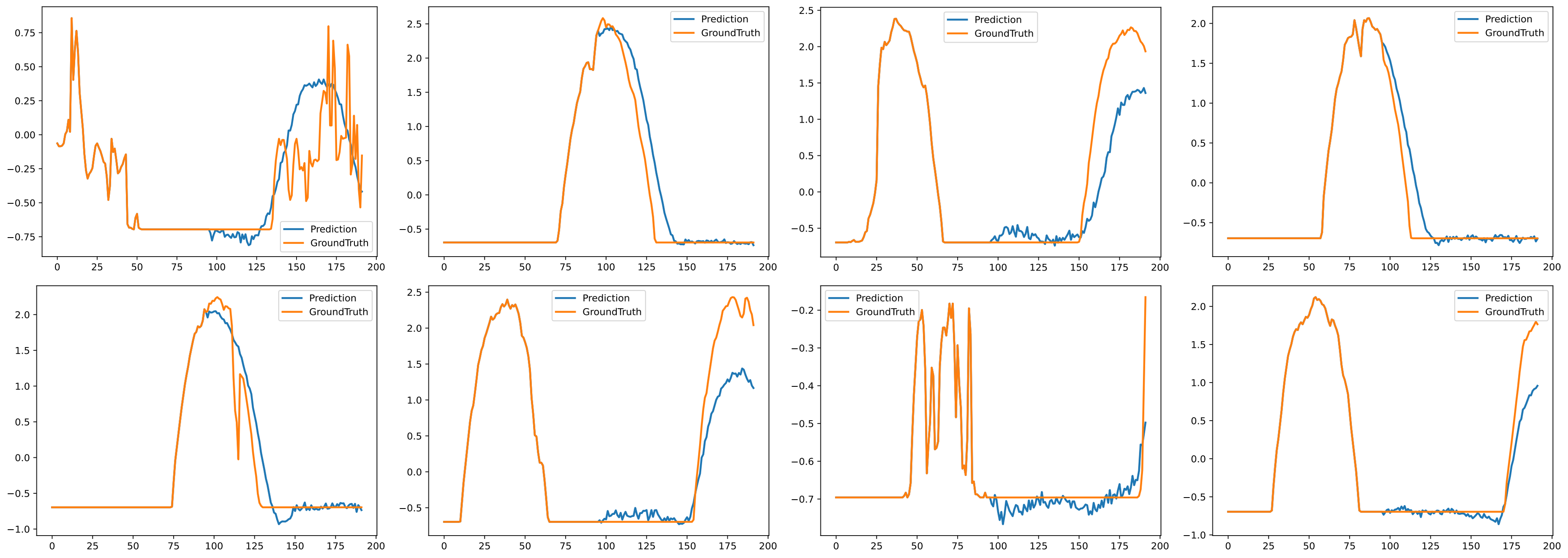}  
    \caption{Examples of forecasts for the Solar dataset with a 96-step predictions.}  
    \label{fig:solar}  
\end{figure}

\begin{figure}[htbp]  
    \centering  
    \includegraphics[width=1.0\textwidth]{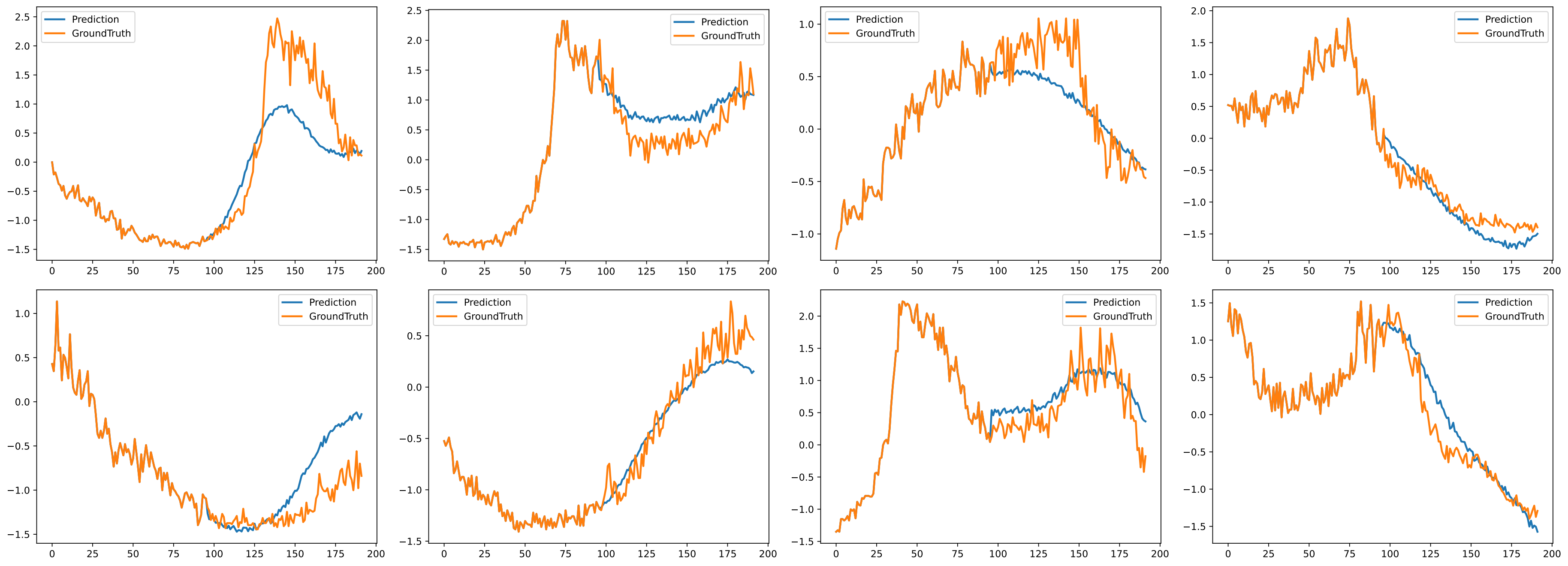}  
    \caption{Examples of forecasts for the PEMS03 dataset with a 96-step predictions.}  
    \label{fig:pems031}  
\end{figure}

\begin{figure}[htbp]  
    \centering  
    \includegraphics[width=1.0\textwidth]{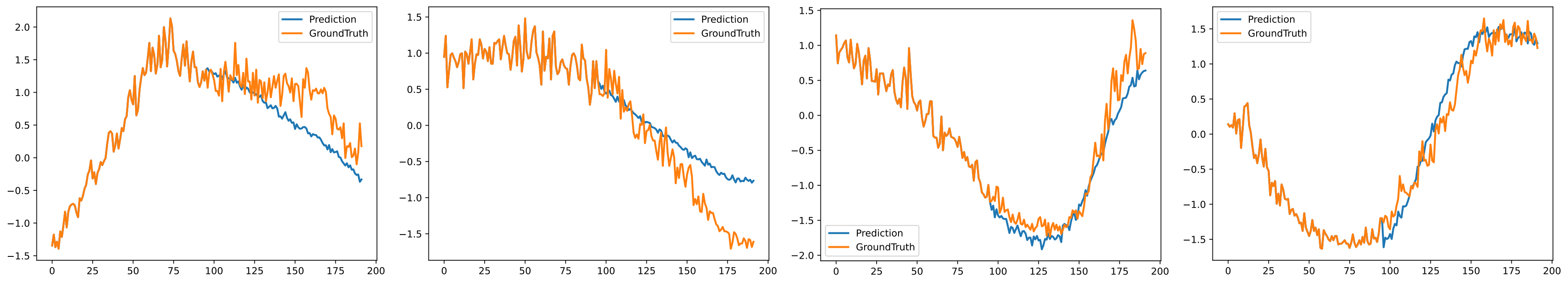}  
    \caption{Examples of forecasts for the PEMS04 dataset with a 96-step predictions.}  
    \label{fig:pems041}  
\end{figure}

\begin{figure}[htbp]  
    \centering  
    \includegraphics[width=1.0\textwidth]{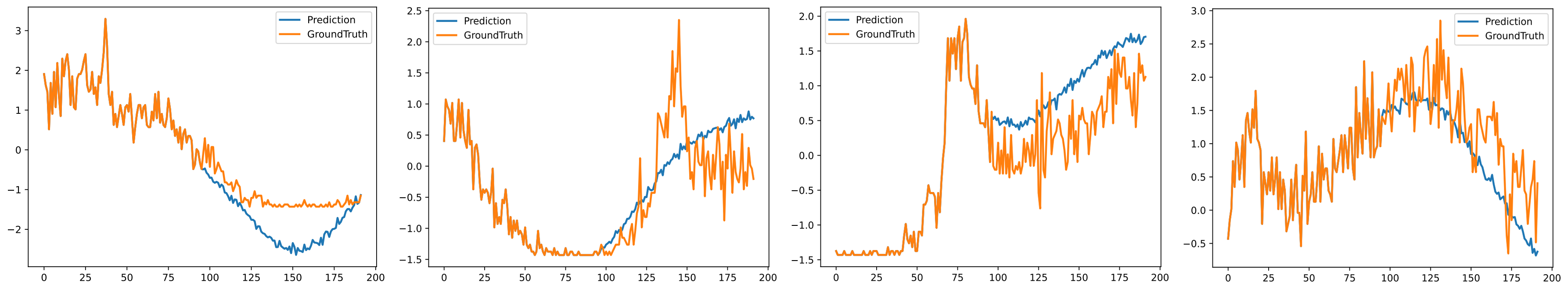}  
    \caption{Examples of forecasts for the PEMS08 dataset with a 96-step predictions.} 
    \label{fig:PEMS081}  
\end{figure}

We visualize HTMformer's ability to predict trends across various time series datasets, including ECL, Traffic, Solar-Energy, Weather, and PEMS (PEMS03, PEMS04 and PEMS08), as shown in~\Cref{fig:ECL,fig:traffic,fig:weather,fig:ETTh2,fig:solar,fig:pems031,fig:pems041,fig:PEMS081}. 
Each example uses a 96-step input to generate 96-step predictions. 
In the visualizations, the orange lines indicate the ground truth values, and the blue lines show the model's predictions.
It accurately captures the cyclical patterns and oscillatory behaviors, and successfully forecasts the overall directional trends.

\begin{figure}[htbp]  
    \centering  
    \includegraphics[width=1.0\textwidth]{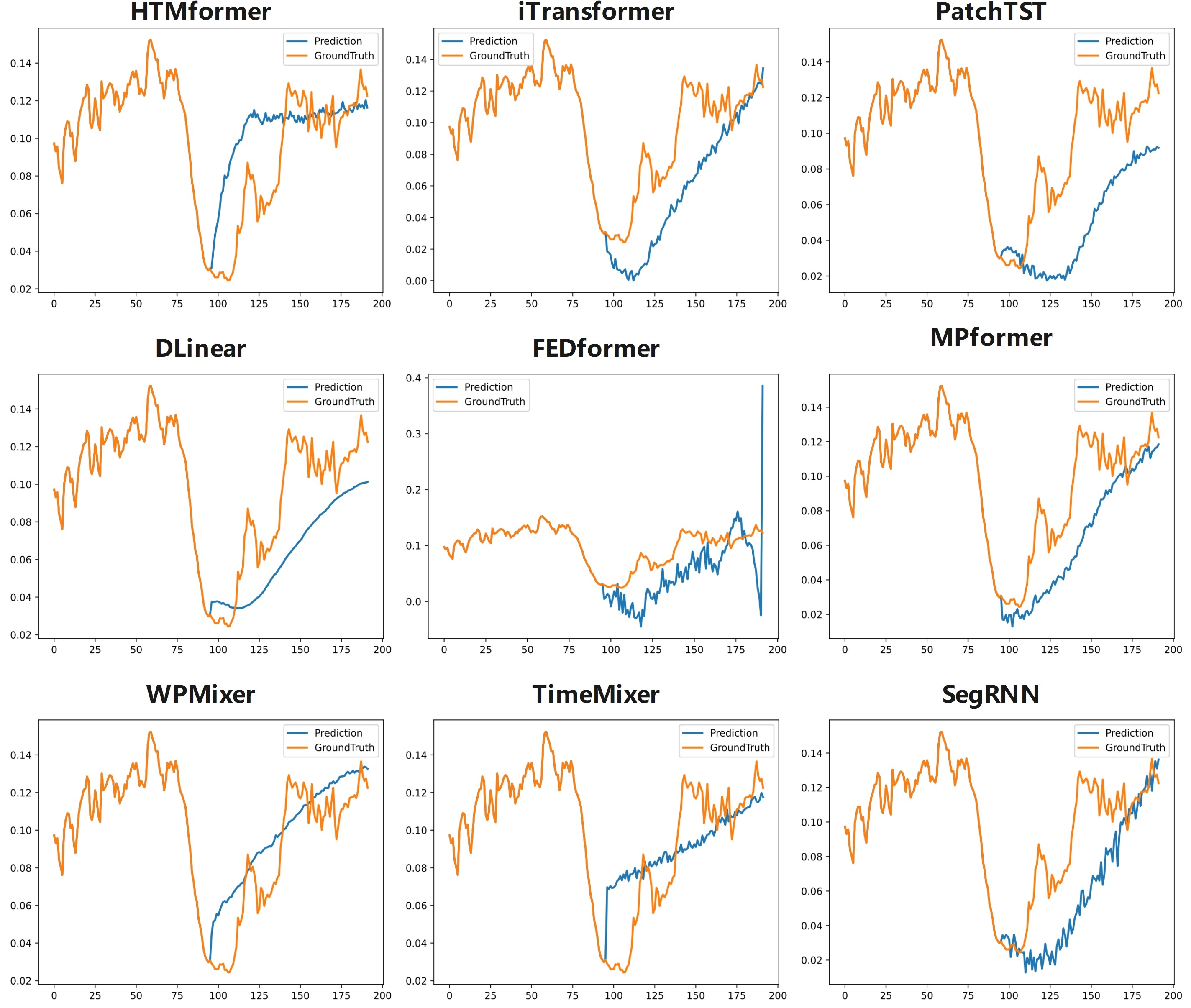}  
    \caption{Visualization of results on the Weather dataset across all selected models.}  
    \label{fig:showcase1}  
\end{figure}

\begin{figure}[htbp]  
    \centering  
    \includegraphics[width=1.0\textwidth]{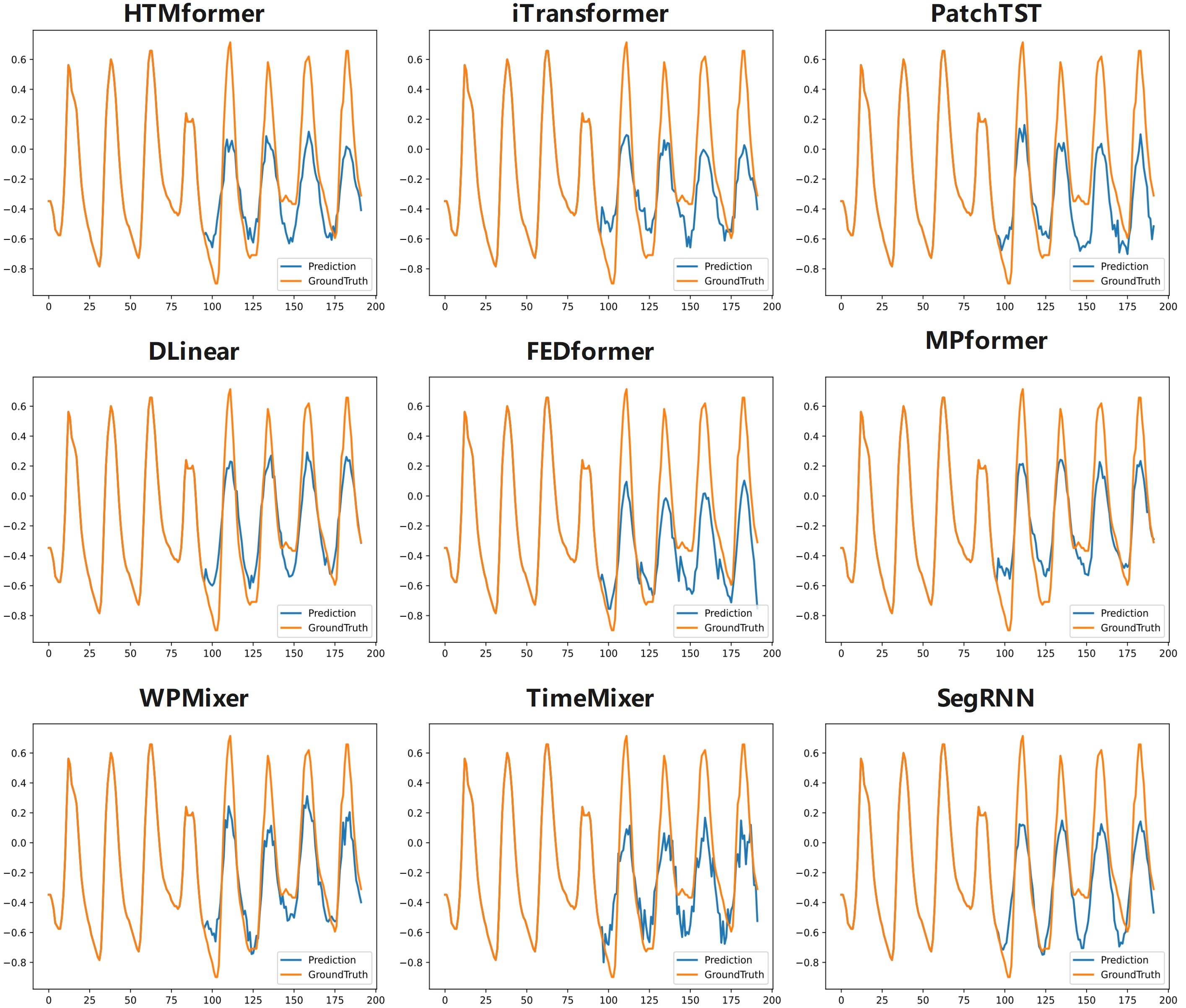}  
    \caption{Visualization of results on the ETTh2 dataset across all selected models.} 
    \label{fig:showcase2} 
\end{figure}

To assess the performance of various models, we perform a qualitative comparison by visualizing the final forecasting results derived from two representative datasets (Weather and ETTh2). 
Among the various models, HTMformer exhibits superior or comparable performance in predicting the most series variations.
Prediction showcases are listed in Figure~\ref{fig:showcase1} and Figure~\ref{fig:showcase2}.

\newpage

\end{document}